\documentclass[conference]{IEEEtran}

\usepackage{times}

\usepackage[numbers]{natbib}
\usepackage{multicol}
\usepackage[bookmarks=true]{hyperref}
\usepackage{amssymb}
\usepackage{amsmath}
\usepackage{booktabs}
\usepackage{amssymb}
\usepackage{array}
\usepackage{makecell}
\usepackage{tabularx}
\usepackage{graphicx}
\usepackage{multirow}
\usepackage{etoc}

\newif\ifanon

\newcommand{\anonhide}[1]{\ifanon\relax\else#1\fi}          
\newcommand{\anonrep}[2]{\ifanon#2\else#1\fi}              
\usepackage{xcolor}

\newif\ifalt

\IEEEoverridecommandlockouts

\anonfalse
\altfalse 

\begin{document}

\pagestyle{plain}  

\title{A Systematic Study of Data Modalities and Strategies for Co-training Large Behavior Models for Robot Manipulation}

\author{
  \anonrep{
      \IEEEauthorblockN{Fanqi Lin\IEEEauthorrefmark{2}\IEEEauthorrefmark{3},
                        Kushal Arora\IEEEauthorrefmark{2},
                        Jean Mercat\IEEEauthorrefmark{2},
                        Haruki Nishimura\IEEEauthorrefmark{2}, 
                        Paarth Shah\IEEEauthorrefmark{2}, 
                        Chen Xu\IEEEauthorrefmark{2}, 
                        Mengchao Zhang\IEEEauthorrefmark{2}, \\
                        Mark Zolotas\IEEEauthorrefmark{2}, 
                        Maya Angeles\IEEEauthorrefmark{2},
                        Owen Pfannenstiehl\IEEEauthorrefmark{2},
                        Andrew Beaulieu\IEEEauthorrefmark{2},
                        Jose Barreiros\IEEEauthorrefmark{2}\thanks{Correspondence to: \texttt{jose.barreiros@tri.global}}}
      \vspace{1ex}
      \IEEEauthorblockA{\IEEEauthorrefmark{2}Toyota Research Institute, Cambridge MA and Los Altos CA, USA}
      \IEEEauthorblockA{\IEEEauthorrefmark{3}Tsinghua University, Beijing, China}
  }
  {Author names omitted for anonymous review}
}

\maketitle

\begin{abstract}
Large behavior models (LBMs) have shown strong dexterous manipulation capabilities by extending imitation learning to large-scale training on extensive multi-task robot data, yet their generalization remains limited by the insufficient coverage of available robot data. To expand this coverage without costly additional data collection, recent work increasingly relies on \textit{co-training}: jointly learning from target robot data and heterogeneous data modalities. However, how different co-training data modalities and training strategies affect policy performance remains poorly understood.
We present a large-scale empirical study examining five co-training data modalities---standard vision-language data, dense language annotations for robot trajectories, cross-embodiment robot data, human videos, and discrete robot action tokens—across single- and multi-phase training strategies. Our study leverages 4,000 hours of robot and human manipulation data and 50M vision–language samples to train vision-language-action (VLA) policies. We evaluate 89 policies over 58,000 simulation rollouts and 2,835 real-world rollouts.
Our results show that co-training with various forms of vision-language and cross-embodiment robot data substantially improves generalization to distribution shifts, unseen tasks, and language following, while discrete action token variants yield no statistically significant benefits. Furthermore, combining effective modalities produces cumulative gains and enables rapid adaptation to unseen long-horizon dexterous tasks via fine-tuning. \ifalt \else We find that training exclusively on robot data degrades the visiolinguistic understanding of the vision-language model backbone, while co-training with the effective data modalities restores these capabilities, as measured by standard vision-language, spatial reasoning, and multimodal reasoning benchmarks. Finally, explicitly conditioning action generation on chain-of-thought traces learned from co-training data does not improve performance in our simulation benchmark. \fi Together, these results provide a systematic understanding of co-training and practical guidance for building scalable generalist robot policies.

\end{abstract}
\vspace{0.5em}
\noindent\anonhide{{\textbf{Project page:}\url{https://co-training-lbm.github.io/}}}

\IEEEpeerreviewmaketitle

\begin{figure*}[!t]
    \centering
    \includegraphics[width=0.98\textwidth]{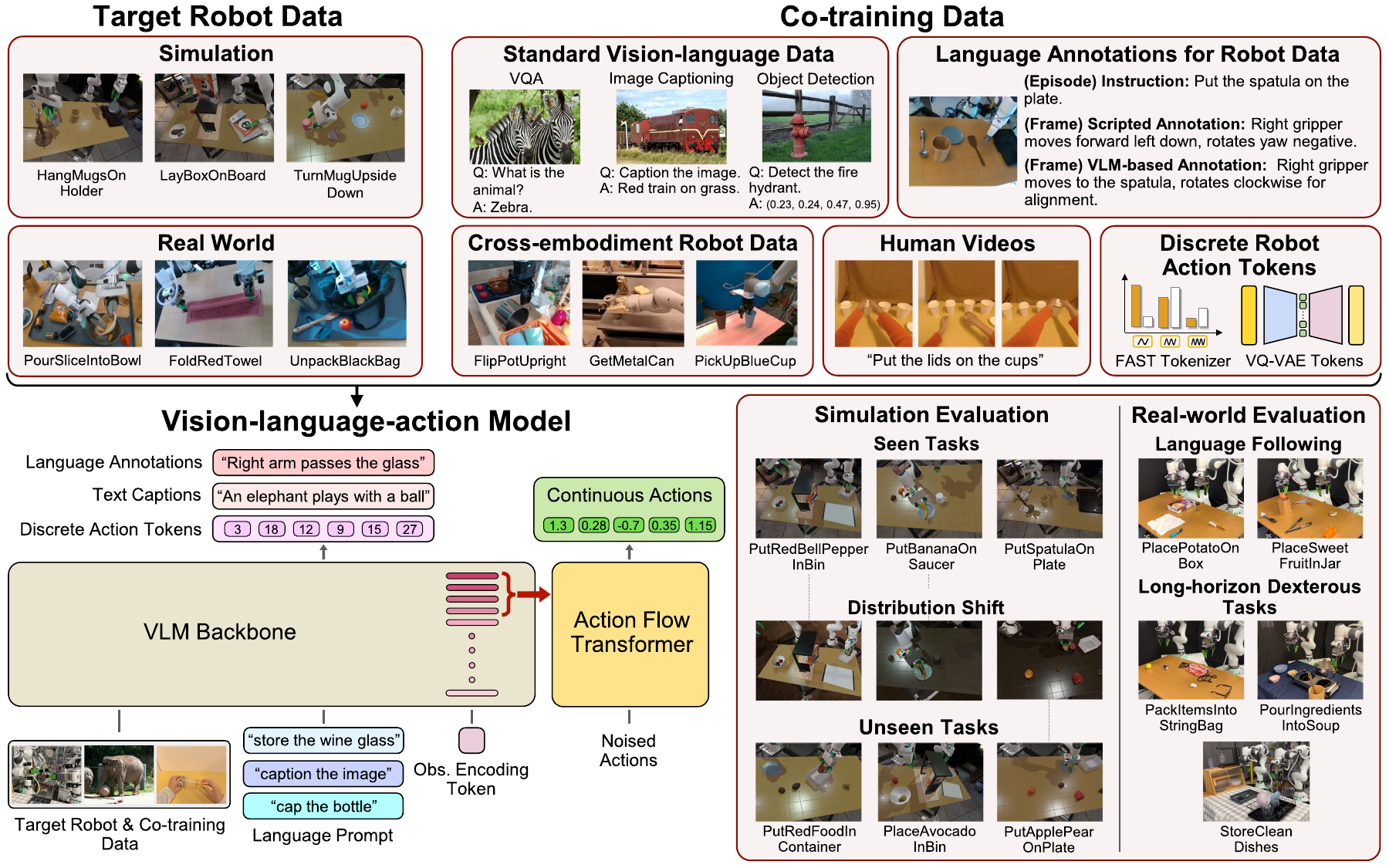}
    \caption{\textbf{Overview of the data, model architecture, and evaluation setup.} Our policy is built on a pretrained vision-language model backbone combined with an Action Flow Transformer. It is trained on target robot data alongside heterogeneous co-training modalities, including standard vision-language data, dense language annotations for robot data, cross-embodiment robot data, human videos, and discrete robot action tokens. We evaluate policies in simulation on seen and unseen tasks, under nominal conditions and distribution shifts, and in the real-world for language following, and long-horizon dexterous manipulation.
        }
    \label{fig:main_overview}
    \vspace{-4mm}
\end{figure*}

\section{Introduction}
Robot learning is increasingly moving towards generalist models that can perceive, understand, and act in the physical world. Recent efforts have focused on training large behavior models (LBMs)~\cite{barreiros2025careful}—embodied foundation models trained on large-scale multi-task robot datasets—to produce dexterous manipulation policies. Within this family, vision-language-action models (VLAs)~\cite{kim2024openvla, zitkovich2023rt, driess2023palm, black2024pi0visionlanguageactionflowmodel,intelligence2025pi05visionlanguageactionmodelopenworld, team2025gemini, bjorck2025gr00t} are a representative subclass of LBMs that integrate visual and linguistic inputs for action generation. Despite progress, LBMs still lag behind non-embodied foundation models, such as vision-language models (VLMs)~\cite{hurst2024gpt, comanici2025gemini, steiner2024paligemma, wang2024qwen2}, in semantic and spatial understanding and in open-world generalization. This limitation can be attributed to the significant disparity in data scale~\cite{goldberg2025good}: robot data is orders of magnitude smaller than the internet-scale text and image corpora used to train VLMs. 

To bridge this data gap, many recent works~\cite{intelligence2025pi05visionlanguageactionmodelopenworld, lin2025onetwovla, bjorck2025gr00t, lee2025molmoact, chen2025villa, qu2025eo, yang2025vlaser} use \textit{co-training}---jointly learning target robot (i.e., the deployment embodiment) data alongside heterogeneous data modalities, aiming to enhance the model’s understanding of the physical world and its generalization abilities. These co-training data modalities include: standard vision-language (VL) data~\cite{intelligence2025pi05visionlanguageactionmodelopenworld, zhou2025chatvla, zhou2025chatvla2, lee2025molmoact}, dense language annotations for robot trajectories~\cite{zawalski2024robotic, lin2025onetwovla, zhai2025igniting, yang2025vlaser, qu2025eo, chen2025training}, cross-embodiment robot data~\cite{barreiros2025careful, liu2024rdt, liu2025hybridvla, team2024octo, kim2024openvla, kim2502fine}, human videos~\cite{bu2025univla, chen2025villa, ye2024latent, luo2025being, yuan2025motiontrans, kareer2025egomimic}, and discrete robot action tokens~\cite{zhai2025igniting, driess2025knowledge, jiang2025galaxea, intelligence2025pi05visionlanguageactionmodelopenworld}. Despite the growing interest, current studies typically evaluate only a subset of these modalities using inconsistent experimental setups, leaving the empirical effectiveness of co-training largely unexplored. 

In this work, we systematically investigate how different data modalities and co-training strategies affect policy performance through large-scale experiments toward a generalist LBM. We adopt a VLA architecture consisting of a pretrained VLM backbone and an action head. Our model is trained with flow matching~\cite{lipman2022flow, liu2022rectified} to predict continuous robot actions and next-token objectives for discrete tokens. An overview of our study is illustrated in Fig.~\ref{fig:main_overview}. 

We evaluate five major co-training data modalities (Fig.~\ref{fig:data}): \textbf{(1) Standard vision-language data}, encompassing visual question answering, object localization, and spatial reasoning tasks, which provide rich commonsense knowledge about the physical world. \textbf{(2) Dense language annotations for robot trajectories}, generated through both heuristic scripting and VLM-based captioning, offering explicit semantic labels for actions, goals, and object relationships. \textbf{(3) Cross-embodiment robot data}, which encompasses manipulation demonstrations across varied robot morphologies and environments. \textbf{(4) Large-scale egocentric human videos} that expose models to diverse visual contexts, object interactions, and motion patterns beyond robot trajectories. We explore two approaches to leverage human videos: first, extracting discrete latent action tokens from a sequence of frames; second, generating language annotations by prompting a VLM with visual context and a task instruction to describe motions, goals, and objects on a per-frame basis. \textbf{(5) Discrete robot action tokens}, where continuous robot actions are compressed into low-dimensional discrete spaces through frequency-based methods (e.g., FAST~\cite{pertsch2025fast}) or vector quantization-based methods (e.g., VQ-VAE~\cite{van2017neural}), probing whether such abstraction improves generalization. 

We also study strategies for incorporating these modalities at different training phases (i.e., rounds of training with distinct data composition). Additionally, we examine whether combining effective co-training modalities yields cumulative performance gains. We further probe whether co-training improves representation quality by fine-tuning on unseen long-horizon, dexterous tasks. We then study how the VLM backbone is shaped by effective co-training modalities using a suite of standard vision-language benchmarks. Finally, we study explicit chain-of-thought (CoT)~\cite{wei2022chain} conditioning for action generation, where the policy first produces intermediate CoT traces learned from the co-training data and then uses them to generate continuous actions.

Our policies are evaluated in both simulation and real-world settings. In total, we train and compare 89 VLA policies using about 4,000 hours of robot and human manipulation data plus 50M vision-language samples. The policies are assessed over 58,000 simulation rollouts across seen and unseen tasks, in nominal and distribution shift (DS) settings, and over 2,835 real-world rollouts covering language following and long-horizon dexterous tasks.

Our results provide practical guidance for co-training LBMs. In summary: (1) Diverse vision-language data and cross-embodiment robot data consistently improve generalization to distribution shifts, unseen tasks, and language following, while discrete action token variants provide no benefit. (2) Combining effective modalities yields cumulative gains and enables more efficient adaptation to unseen long-horizon dexterous tasks via fine-tuning. (3) Training exclusively on robot data can erode the visiolinguistic capabilities of the VLM backbone, whereas effective co-training helps preserve this understanding, as reflected by improved performance on standard vision-language benchmarks. (4) Explicitly conditioning action generation on chain-of-thought traces learned from co-training data does not improve performance in our simulation benchmark. Together, these findings constitute a controlled, large-scale empirical map of which co-training signals and strategies are most useful for building scalable, generalist robot policies.

\section{Method}
In this section, we first introduce our co-training framework in Section~\ref{sec:framework}, including the problem formulation, model architecture, and our co-training and inference strategies. We then describe in Section~\ref{sec:data_curation} how the target robot data and diverse co-training datasets are curated for this study.

\subsection{Co-training Framework}
\label{sec:framework}

\subsubsection{Problem Formulation}
Our objective is to learn a policy $\pi_{\theta}$ that can leverage diverse co-training data modalities. The policy $\pi_{\theta}$ takes as input a sequence of $n$ images $I^{1:n}$ and a text prompt $\ell$. For continuous robot actions, the model is trained with flow matching (FM)~\cite{lipman2022flow, liu2022rectified} as the learning objective. Specifically, given an action chunk $A_{t}$, a FM timestep $\tau \in [0,1]$, and sampled noise $\epsilon \sim N(0,\mathbf{I})$, we construct a noised action chunk as $A^{\tau}_{t} = \tau A_{t} + (1-\tau)\epsilon$. The model is then trained by minimizing the loss:
\begin{equation}
\mathcal{L}_{\mathrm{FM}} = \left\lVert \pi_{\theta}^{a}\!\left(I_{t}^{1:n},\, \ell, A_{t}^{\tau},\, \tau \right) - \left(A_{t} - \epsilon\right)\right\rVert^{2},
\end{equation}
where $(A_{t} - \epsilon)$ and $\pi_{\theta}^{a}(\cdot)$ are the ground-truth and predicted flow vector, respectively. 
For text tokens or discrete action tokens, the model is optimized to minimize the cross-entropy (CE) loss between the ground-truth token sequence $x_{1:M}$ and the predicted logits $\pi_{\theta}^{\ell}(\cdot)$, such that:
\begin{equation}
\mathcal{L}_{\mathrm{CE}}
= \mathcal{H}\!\left(
x_{1:M},\;
\pi_{\theta}^{\ell}\!\left(I_{t}^{1:n},\, \ell\right)
\right)
\end{equation} 

When jointly optimizing for both continuous and discrete modalities, we combine the objectives into a weighted sum:
\begin{equation}
\mathcal{L} = M_{FM}\mathcal{L}_{FM} + w*M_{CE}\mathcal{L}_{CE},
\end{equation}
Here, $w$ is the weight applied to the CE loss, $M_{FM}$ is the FM loss mask indicating whether the model should predict continuous actions for a given sample, and $M_{CE}$ is the mask specifying token positions used to compute the CE loss.

\subsubsection{Model Architecture}
\label{sec:architecture}
We adopt a VLA architecture (Fig.~\ref{fig:main_overview}) composed of a pretrained VLM backbone and a flow transformer action head (ActionFT). The VLM is initialized from PaliGemma2-PT (google/paligemma2-3b-pt-224~\cite{steiner2024paligemma}). It encodes both the observed images and the language prompt describing the task, and can optionally be trained to generate text or discrete action tokens in addition to providing visiolinguistic representations. To obtain a compact representation for action generation, similarly to \cite{lin2025vote}, we introduce a special observation encoding token into the backbone’s vocabulary and append it to the end of the text prompt. We extract the hidden state vectors corresponding to this token from the last four layers of the VLM to form a single global conditioning embedding and feed this into the ActionFT. The ActionFT follows the diffusion transformer design introduced in~\cite{barreiros2025careful}. \ifalt \else     \ifalt Our ActionFT is a diffusion transformer consisting \else It consists \fi of 8 flow transformer layers, each conditioned on observation features and the flow-matching timestep via an adaptive layer norm (adaLN) MLP~\cite{peebles2023scalable}. The ActionFT receives the global conditioning embedding, a noised chunk of continuous actions, and the flow matching time variable, and is trained to predict the flow vector that guides iterative denoising of the actions toward the target trajectory. Unlike prevalent methods~\cite{black2024pi0visionlanguageactionflowmodel}, our approach uses only a single token as visiolinguistic representations rather than attention keys and values from all VLM layers. \fi Our ablation studies (see Appendix~\ref{sec:model_ablation}) indicate that this compact representation enhances the model's generalization ability to unseen tasks and distribution shift.

\subsubsection{Co-training and Inference Strategies}
\label{sec:cotraining_inference_strategies}
Co-training data can be used at different phases of model training. We explore three strategies (Table~\ref{tab:co_training_strategies}): (1) \textit{Single-phase co-training}: train jointly on target robot data and co-training data modalities in a single phase; (2) \textit{Two-phase 1st-phase-only co-training}: train only on co-training modalities during the 1st-phase, then on target robot continuous actions in the 2nd-phase; (3) \textit{Two-phase full co-training}: the same 1st-phase as strategy (2), but train on both co-training and target robot data in the 2nd-phase.

We fix the training loss weights and batch data ratios based on ablation experiment results (Appendix~\ref{sec:hyperparameter_ablation}). Full training details are provided in Appendix~\ref{sec:training_details}. \ifalt
    While our primary approach treats co-training data solely as auxiliary supervision, we explore chain-of-thought (CoT) conditioning in action generation to take advantage of co-training data, as shown in Appendix~\ref{sec:cot}.
\else

    Beyond investigating when to incorporate co-training data, we examine ways to leverage it. While our primary approach treats co-training data solely as auxiliary supervision, recent works~\cite{lin2025onetwovla, zawalski2024robotic, zhai2025igniting, team2025gemini, chen2025training, chen2025villa} suggest that explicitly conditioning action prediction on CoT traces can boost policy performance. We therefore also test this alternative paradigm for selected co-training modalities: language annotations for robot trajectories and latent actions for videos. 

When generating actions conditioned on CoT traces at inference time, the VLM backbone first produces a CoT trace (see Fig.~\ref{fig:model_cot}). The observation token is then appended to its end, and the resulting visiolinguistic embedding (encoding images, task prompts, and CoT traces) is extracted to condition the ActionFT for continuous action prediction. During training, we introduce probabilistic CoT conditioning: with probability $p$, the policy is trained to generate actions conditioned on CoT traces extracted from the co-training data; with probability $1-p$, the policy generates actions directly without CoT conditioning. Importantly, CoT conditioning differs from other co-training strategies only in that the predicted co-training tokens are used directly to form the visiolinguistic embedding, instead of just providing auxiliary supervision. \ifalt \else
    We evaluate the impact of this explicit inference-time CoT approach in Section~\ref{sec:cot}. 
\fi

\fi

\begin{table}[t]
\centering
\caption{Data composition for different co-training strategies.}
\label{tab:co_training_strategies}   
\scriptsize  
\begin{tabular}{p{1.9cm} >{\centering\arraybackslash}p{1.7cm} 
                        >{\centering\arraybackslash}p{1.7cm} 
                        >{\centering\arraybackslash}p{1.7cm}}
\toprule
\textbf{Co-training} \\ \textbf{strategy} & \textbf{Phase} 
& \multicolumn{2}{c}{\textbf{Data composition}} \\
\cmidrule(lr){3-4}
& & \textbf{Target-robot continuous-action data} 
& \textbf{Co-training data} \\
\midrule

Single-phase \newline co-training 
& 1st-phase 
& \checkmark 
& \checkmark \\

\midrule

\multirow{3}{=}{Two-phase 1st-phase-only co-training}
& 1st-phase & -- & \checkmark \\
& & & \\
& 2nd-phase & \checkmark & -- \\

\midrule

\multirow{2}{=}{Two-phase full co-training}
& 1st-phase & -- & \checkmark \\
& 2nd-phase & \checkmark & \checkmark \\

\bottomrule
\end{tabular}
\vspace{-4mm}
\end{table}

\ifalt

\else
    \subsection{Data Curation}
\label{sec:data_curation}

\begin{figure*}[t]
    \centering
    \includegraphics[width=\linewidth]{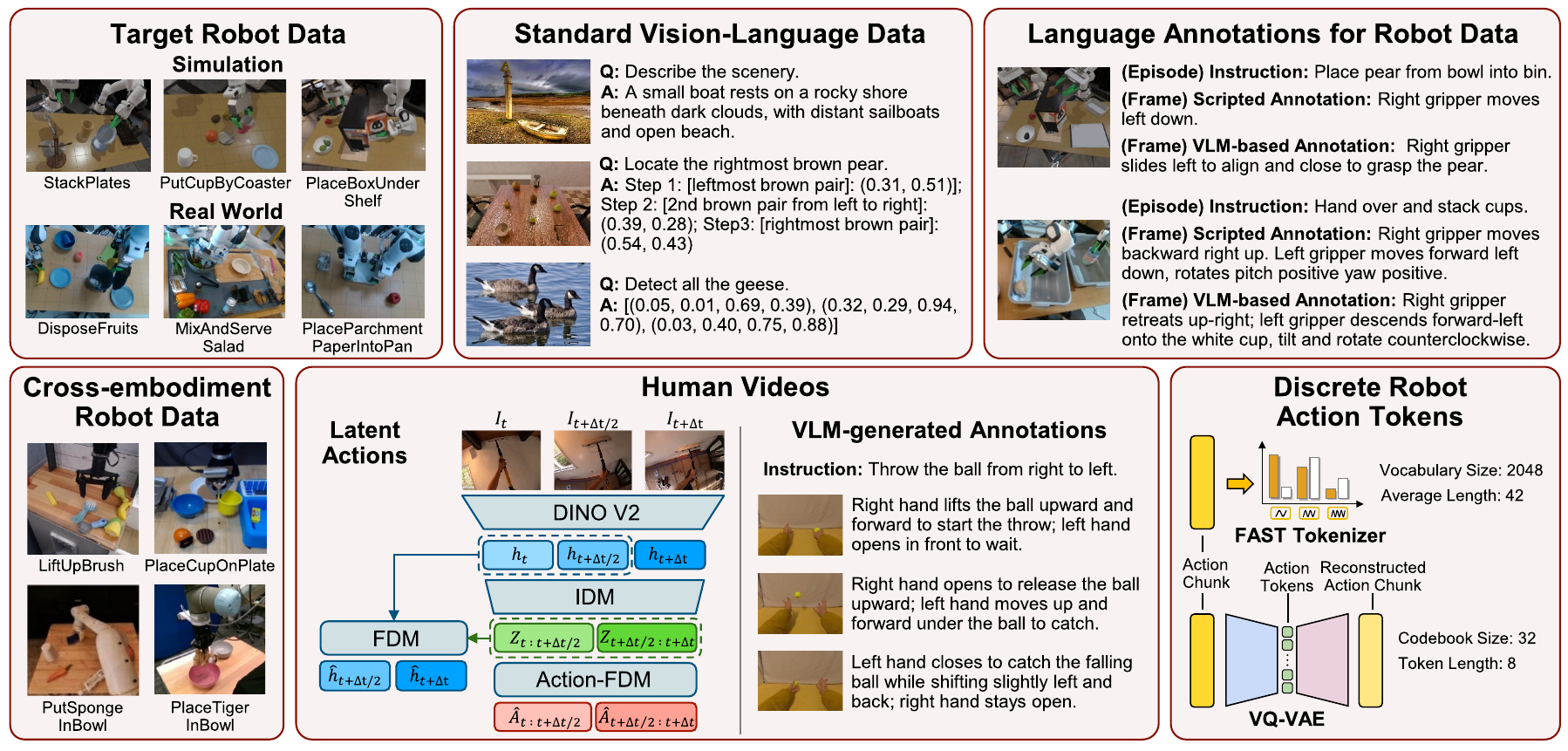}
    \caption{\textbf{Overview of the training data.} Our dataset comprises target robot data collected in both simulation and real-world, and five heterogeneous co-training data modalities: standard vision-language data for commonsense understanding, spatial reasoning, and object grounding; dense language annotations for robot data, generated via heuristic scripting and VLM-based captioning; cross-embodiment robot data capturing diverse robot morphologies and manipulation tasks; human videos, from which we derive either latent action tokens using a latent action model (LAM) or VLM-generated annotations; and discrete robot action tokens, including near-lossless FAST tokens and compact VQ-VAE tokens. These co-training modalities, together with the target robot data, constitute a unified dataset of $\sim$4,000 hours of manipulation data and 50M vision-language samples.}
    \label{fig:data}
\end{figure*}

We curate a comprehensive dataset consisting of target-robot expert demonstration data and five distinct co-training modalities. Our dataset comprises approximately 4,000 hours of manipulation data spanning both robot episodes and human videos, complemented by 50M vision-language samples that encompass standard vision-language data as well as dense annotations for both robot and human data. Fig.~\ref{fig:data} shows an overview of the training data. 

\subsubsection{Target Robot Data} We adopt high-quality robot data from our previous study~\cite{barreiros2025careful} as our primary in-distribution training corpus. This dataset, which is referred to as \anonrep{TRI}{Target}-Ramen, contains 523 hours of manipulation data spanning 403 tasks and 53,411 demonstrations. It consists of real-world  (478 hours, 362 tasks, 46,063 demonstrations; namely \anonrep{TRI}{Target}-Ramen-Real) and simulation data (45 hours, 41 tasks, 7,348 demonstrations; namely \anonrep{TRI}{Target}-Ramen-Sim).~\ifalt
    \subsection{Target Robot Data} 
\else\fi 
All demonstrations are collected via teleoperation on dual Franka Panda robotic arms \anonrep{as described in~\cite{barreiros2025careful}}{as in~\cite{barreiros2025careful}}. 

The observation space includes i) 4 RGB images (missing images are zero-padded), and ii) a natural-language instruction. The action space includes i) end-effector poses w.r.t. the station’s base frame, and ii) gripper widths. Actions are represented as relative trajectories as in~\cite{chi2024universal}. The policy is conditioned only on the current observation and predicts an action chunk with a horizon of 16. 

\subsubsection{Standard Vision-Language Data} To enhance the model's multimodal understanding (e.g., semantics, spatial, planning), we incorporate VL datasets specifically designed for robotic scenarios: RoboPoint~\cite{yuan2024robopoint} and RefSpatial~\cite{zhou2025roborefer}. RoboPoint comprises 1.3M data samples with 8.2M question answering pairs, while RefSpatial contains 2.5M samples with 20M question answering pairs. Both datasets provide annotations tailored for spatial referring tasks, spanning qualitative visual question answering, quantitative queries on object and spatial attributes/relations, point coordinate prediction, and multi-step spatial reasoning.

\subsubsection{Dense Language Annotations for Robot Trajectories} To augment \anonrep{TRI}{Target}-Ramen robot trajectories with action-grounded textual descriptions, we employ two complementary annotation strategies: \textbf{(1) Scripted Annotation.} Following~\cite{zawalski2024robotic}, we apply heuristic rules to generate per-step low-level action primitives by comparing future and current end-effector states over a 16-step horizon (corresponding to the horizon of the action chunk). These annotations capture the robot's end effector translational movement, rotational changes, and gripper state transitions. \textbf{(2) VLM-based Annotation.} While scripted annotations provide structured action descriptions, their linguistic diversity is limited and they lack contextual information about object and environment interactions. To address this, we prompt a VLM (GPT-5~\cite{gpt5ishere}) to generate rich, contextually grounded descriptions. Specifically, for each robot episode, we provide the VLM with: (i) frames downsampled at 2-second intervals (approximating the 1.6-second action chunk duration), (ii) the episode-level task instruction, (iii) action hints generated by the heuristic rules described above, and (iv) a reference image depicting a world-frame coordinate to calibrate spatial directions (forward/backward, left/right, up/down). The VLM is then prompted to produce diverse frame-level action descriptions that capture interactions with objects and the environment. To achieve a higher temporal density, we process the dataset twice with a one-second offset between passes, resulting in annotations at a per-second frequency. More details are available in Appendix~\ref{sec:dense_lang_details}.

\subsubsection{Cross-embodiment Robot Data} We adopt the same cross-embodiment dataset used in~\cite{barreiros2025careful}, referred to as OXE-Ramen, which is a curated subset of the Open X-Embodiment dataset~\cite{o2024open}. This collection encompasses diverse robot morphologies and manipulation scenarios, totaling 1,150 hours across 12 robot setups, 924 tasks, and 466,415 demonstrations. Observation and action spaces follow the same as in the target
robot data.

\subsubsection{Human Videos} To extract rich information about the motion and action (e.g, move forward and grasp) from human videos, we explore two distinct approaches: \textbf{(1) Latent Actions.} We utilize publicly available egocentric human video datasets (e.g., Ego4D~\cite{grauman2022ego4d}, EgoDex~\cite{hoque2025egodex}, Something-Something V2~\cite{goyal2017something}, Epic Kitchen~\cite{damen2020epic}, HoloAssist~\cite{wang2023holoassist}), totaling 2,271 hours after filtering (see detailed data composition in Table~\ref{tab:human_video_annotation_stats}). We train a latent action model (LAM) jointly on human videos, \anonrep{TRI}{Target}-Ramen, and OXE-Ramen to learn a unified discrete action representation. Given consecutive frames $I_{t}$, $I_{\,t + \frac{\Delta t}{2}}$, $I_{\,t + \Delta t}$, we encode them using a pretrained DINOv2~\cite{oquab2023dinov2} vision encoder to obtain visual features $h_{t}$, $h_{\,t + \frac{\Delta t}{2}}$, $h_{\,t + \Delta t}$. Following~\cite{chen2025villa, bu2025univla}, the LAM learns a quantized codebook of latent actions (codebook size $C$) using three modules: an inverse dynamics model (IDM), a visual forward dynamics model (FDM), and an action forward dynamics model (ActionFDM).

The IDM predicts two latent action segments:
\begin{equation}
Z_{t : t + \frac{\Delta t}{2}},\;
Z_{t + \frac{\Delta t}{2} : t + \Delta t}
=
\mathrm{IDM}\!\left(
h_{t},\;
h_{t + \frac{\Delta t}{2}},\;
h_{t + \Delta t}
\right),
\end{equation}
while the FDM reconstructs future visual features:
\begin{align}
\hat{h}_{t + \frac{\Delta t}{2}}
& = \mathrm{FDM}\!\left(
h_{t},\;
Z_{t : t + \frac{\Delta t}{2}}
\right), \\
\hat{h}_{t + \Delta t}
& = \mathrm{FDM}\!\left(
h_{t + \frac{\Delta t}{2}},\;
Z_{t + \frac{\Delta t}{2} : t + \Delta t}
\right).
\end{align}

To encourage latent tokens to capture physical dynamics alongside visual changes, we additionally supervise on robot data (\anonrep{TRI}{Target}-Ramen, OXE-Ramen) by reconstructing the ground-truth action chunks:
\begin{align}
\hat{A}_{t : t + \frac{\Delta t}{2}}
&= \mathrm{ActionFDM}\!\left(
Z_{t : t + \frac{\Delta t}{2}}
\right), \\
\hat{A}_{t + \frac{\Delta t}{2} : t + \Delta t}
&= \mathrm{ActionFDM}\!\left(
Z_{t + \frac{\Delta t}{2} : t + \Delta t}
\right).
\end{align}

For human videos, we omit action reconstruction due to missing ground-truth actions. After training, we run inference to obtain $Z_{t : t + \frac{\Delta t}{2}}$ and $Z_{t + \frac{\Delta t}{2} : t + \Delta t}$ at each time step, concatenate them to form $Z_t$, and quantize each $Z_t$ into 8 discrete tokens from a codebook of size $C=32$. For robot videos, $\Delta t=1.6$ seconds (matching the 16-step action chunk), while for human videos we use $\Delta t=1.0$ seconds to account for faster motion.Implementation details are provided in Appendix~\ref{sec:latent_action_details}.
\textbf{(2) VLM-generated Annotations.} As an alternative to latent actions, language can also serve as a unified representation across different embodiments. It captures rich semantic information about actions, goals, and objects while being naturally compatible with VLAs. Specifically, we provide GPT-5 with: (i) frames from human videos downsampled at 1-second intervals, (ii) episode-level instructions, and (iii) a reference image depicting the world-frame coordinate triad. We prompt the VLM to generate concise motion descriptions of both hands, including rich information about interactions with objects and the environment. We utilize Ego4D, EgoDex and Something-Something V2 datasets, yielding 9M annotated data samples. During training, we treat these samples as another form of VL data. More details are shown in Appendix~\ref{sec:vlm_generated_details}.

\subsubsection{Discrete Robot Action Tokens} Several works suggest that co-training models on both continuous and discrete robot action representations improves sample efficiency and generalization. Motivated by this, we explore two forms of discrete robot action tokens: \textbf{(1) FAST Tokens.} We employ FAST~\cite{pertsch2025fast} to convert continuous action chunks into a compressed, near-lossless sequence of discrete tokens. We use the off-the-shelf tokenizer without fine-tuning, as we observe no significant improvements in either average token length or reconstruction error after fine-tuning on our data (see Appendix~\ref{sec:appendix_discrete_action_tokens}). When applied to our \anonrep{TRI}{Target}-Ramen dataset, FAST produces sequences of average length 42.1 with a vocabulary of 2,048 tokens. \textbf{(2) VQ-VAE Discrete Action Tokens.} We compress action chunks into 8 discrete tokens with codebook size 32 using a VQ-VAE~\cite{van2017neural} on both \anonrep{TRI}{Target}-Ramen and OXE-Ramen. Notably, these dimensions are identical to those of the latent actions learned from videos. This results in a much more compact, lower-dimensional representation compared to FAST tokens.   
\fi

\section{Experiments}

\ifalt
\else
    \begin{figure*}[t]
    \centering
    \includegraphics[width=0.8\linewidth]{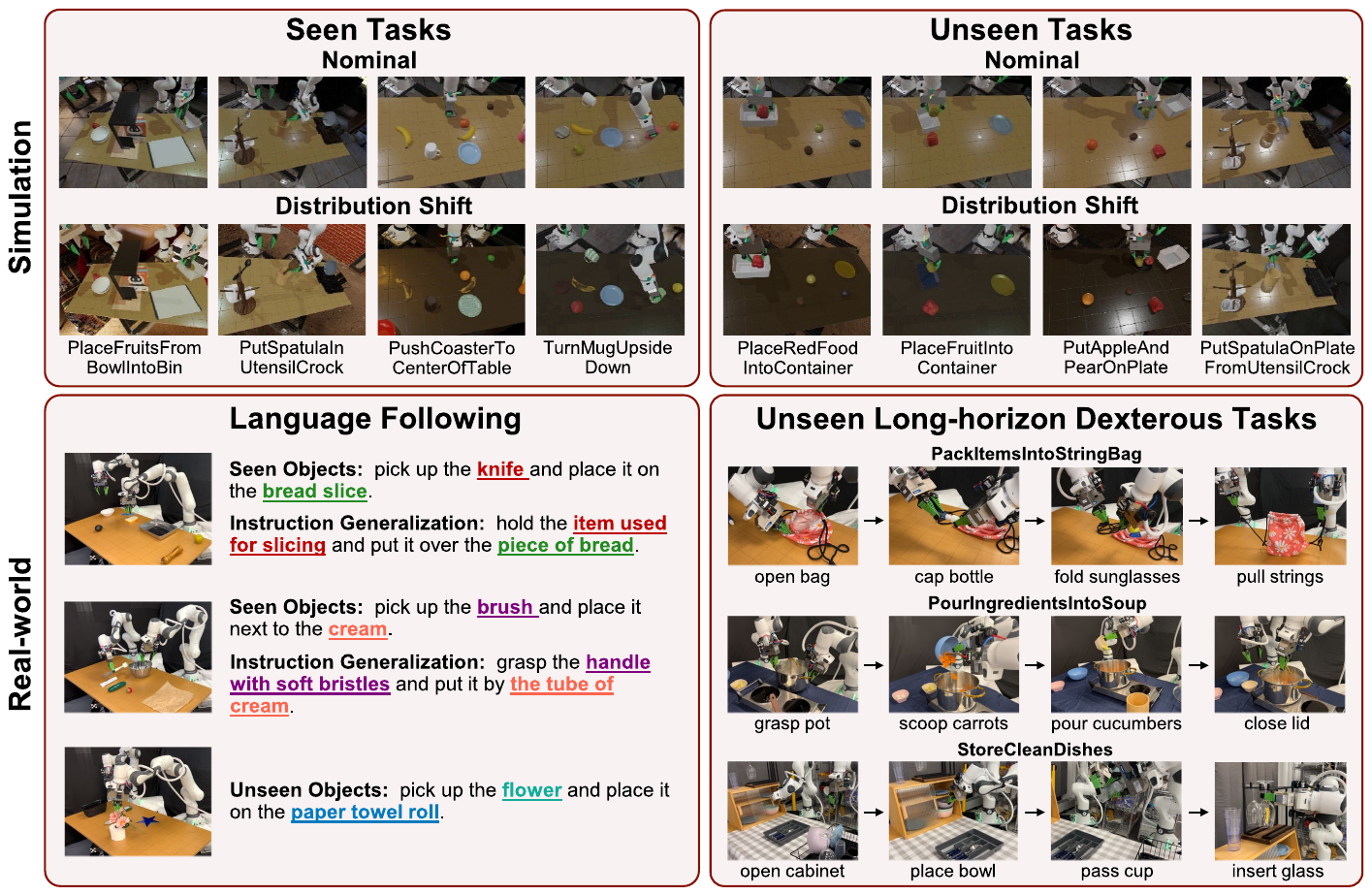}
    \caption{\textbf{Simulation and real-world evaluation.} 
    Policies are evaluated in simulation on 13 seen and 8 unseen tasks under nominal and distribution shift (DS) conditions, where DS introduces appearance changes (e.g., lighting, textures, distractors, camera parameters). Real-world evaluations include language-following experiments with seen objects, instruction generalization through paraphrasing, and unseen objects, as well as adaptation to unseen long-horizon dexterous tasks via fine-tuning.}
    \label{fig:sim_real_experiments}
\end{figure*}

\fi

To systematically investigate the effectiveness of co-training data modalities and strategies, we conduct large-scale experiments to address the following research questions:

\begin{enumerate}
    \item How do different co-training data modalities, incorporated at different training phases, influence policy performance on various dimensions (in-distribution, generalization to distribution shift, unseen tasks, and language following)?
    \item Does combining effective co-training modalities yield cumulative performance gains? 
    \item Can co-training enhance the quality of learned representations, thereby enabling rapid adaptation via fine-tuning to unseen long-horizon, dexterous tasks?

\ifalt
\else
    \item How do the effective co-training modalities shape the VLM backbone?
    \item How does explicitly conditioning action generation on CoT learned from co-training data affect policy performance?
\fi
    
\end{enumerate}

\subsection{Evaluation}
We evaluate policy performance across multiple important dimensions: in-distribution performance, robustness to DS, generalization to unseen tasks, and language following. We additionally assess adaptation to unseen long-horizon dexterous manipulation tasks via fine-tuning. To this end, we conduct large-scale simulation and real-world experiments \ifalt(Fig.~\ref{fig:main_overview},~\ref{fig:sim_real_experiments}).\else(Fig.~\ref{fig:sim_real_experiments}).\fi

\ifalt

\else
    \subsubsection{Simulation Benchmark}
We adopt the simulation benchmark presented in~\cite{barreiros2025careful} built on top of Drake~\cite{tedrake2019drake}. Our benchmark includes 13 seen tasks and 8 unseen tasks (3 unseen tasks defined in~\cite{barreiros2025careful} plus 5 newly introduced to further probe generalization; see Appendix~\ref{sec:sim_details}). Each policy is evaluated with 50 rollouts per task under both nominal and DS conditions, and performance is measured by success rate.

The 13 seen tasks fall within the training distribution, while the 8 unseen tasks are designed to probe generalization beyond it. These unseen tasks span several challenges: (i) semantic understanding (e.g., identifying ``red food" in \textit{PlaceRedFoodIntoContainer}, or distinguishing fruits from vegetables in \textit{PlaceFruitIntoContainer}); (ii) multi-step manipulation (e.g., sequentially placing an apple and a pear on a plate in \textit{PutAppleAndPearOnPlate}); and (iii) compositional generalization (e.g., generalizing from training demonstrations of placing object A on C and object B on D to placing object A on D). Here, “unseen tasks” refer to skills that do not appear in the training data, although the underlying objects and environments do.

To assess the model's robustness to appearance changes, we use DS conditions of the simulation benchmark, which introduce variations in lighting, environmental backgrounds, camera parameters, objects and table textures, and colors relative to the training distribution. Nominal conditions maintain consistency with the training data across these factors. 

\fi

\ifalt

\else
    \subsubsection{Real-world Evaluation}
We evaluate policies on a dual-arm Franka robot platform across the following three settings. Details of policy deployment are provided in Appendix~\ref{sec:policy_deployment}.

\textbf{Language Following.} To evaluate the model's ability to follow natural language instructions, we design a suite of language-guided pick-and-place experiments that assess three distinct scenarios:
\textbf{(i) Seen Objects:} Objects in this setting appear in the training data. Instructions follow a simple template: ``pick up [object A] and place it in/on/next to [object B]", where objects are explicitly referenced by name. 
\textbf{(ii) Instruction Generalization:} This setting tests the model's ability to comprehend the underlying meaning behind natural language by rephrasing the instruction. Specifically, the policy must (1) understand semantic object categories (e.g., ``writing tools" referring to pens), (2) recognize objects by their attributes (e.g., ``the handle with soft bristles" referring to a brush), and (3) demonstrate robustness to paraphrasing (e.g., altered syntax and verb choices).
\textbf{(iii) Unseen Objects:} This setting employs objects absent from the target robot training data. Instructions follow the same template as the seen objects setting.

Across all three settings, we evaluate 15 spatial layouts, each containing 6-8 tabletop objects. For each layout, we use three distinct language instructions targeting different objects, yielding a total of 45 rollouts per setting. For the instruction generalization setting, we employ identical spatial layouts and target manipulation outcomes as the seen objects setting, but vary the instruction phrasing. The full evaluation suite covers 49 seen objects and 52 unseen objects in total. We report average task completion percentage as the evaluation metric (see Appendix~\ref{sub:real_exp_details} for rubrics, experimental procedure, and rubric quality assurance (QA)).

\textbf{Long-horizon Dexterous Manipulation.} To investigate whether co-training facilitates rapid adaptation of the pretrained model to new challenging tasks unseen during pretraining, we design three long-horizon, dexterous manipulation tasks: \textit{PackItemsIntoStringBag}, \textit{PourIngredientsIntoSoup}, and \textit{StoreCleanDishes}. On average, each task consists of 13 steps and takes 93 seconds to execute. These tasks require fine-grained manipulation skills beyond simple pick-and-place operations, such as capping a bottle, scooping food from a bowl into a pot using a spatula, or inserting a wine glass upside-down into a dish rack. For each task, we collect 200 demonstrations for fine-tuning. Each checkpoint is evaluated for 30 rollouts per task, and we report average task completion percentage as the evaluation metric (see Appendix~\ref{sub:real_exp_details} for rubrics, experimental procedure, and rubric QA). 
\fi

\subsubsection{Statistical Analysis Framework}

We perform rigorous statistical analysis similar to recent work~\cite{barreiros2025careful}, with pairwise hypothesis tests~\cite{welch1947generalization, snyder2025step} and Compact Letter Display (CLD)~\cite{piepho2004algorithm} for comparison. Co-training strategies not sharing any CLD alphabet are significantly different in average performance at 5\% family-wise error rate (FWER). Bayesian uncertainty estimates are reported for individual strategies, with posterior uncertainty visualized as violin plots overlaid on bar charts. Dots and horizontal lines indicate empirical and posterior means, respectively. Raw empirical distributions are reported in Appendix~\ref{sec:statistical_analysis_details} for the task-completion results, together with more details of our statistical framework.

\subsection{How do different co-training modalities, incorporated at different training phases, influence policy performance?}
\begin{figure*}[t]
    \centering
    \includegraphics[width=0.9\linewidth]{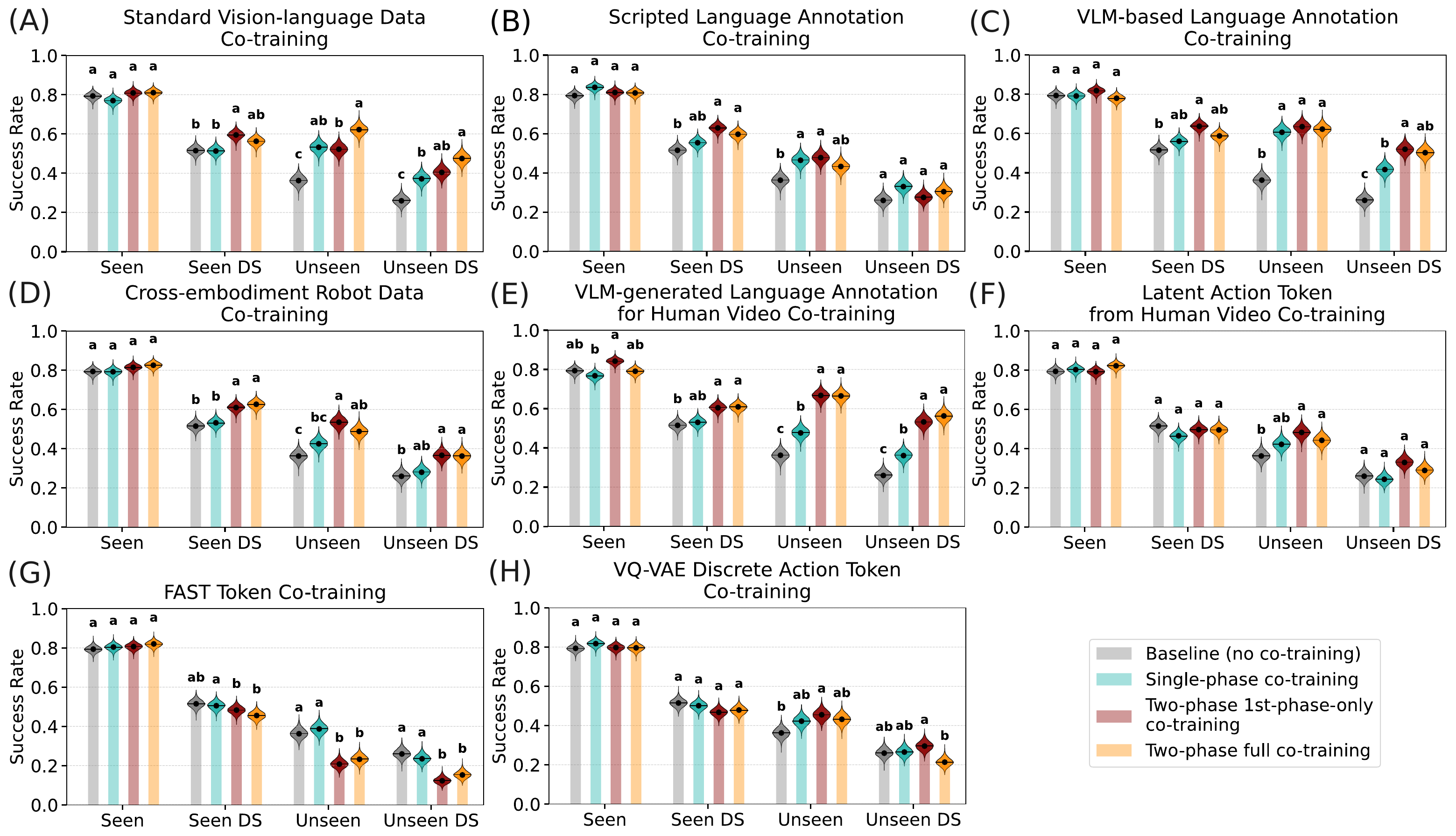}
    \caption{\textbf{Simulation ablation of co-training data and strategies.} Comparison of the no-co-training baseline with policies co-trained on a single data modality across sequential training phases. Policies are evaluated on seen and unseen tasks under nominal and distribution shift conditions (A--H denote data modalities).}
    \label{fig:single_modality_ablation}
    \vspace{-2mm}
\end{figure*}

\begin{figure*}[t]
    \centering
    \includegraphics[width=0.9\linewidth]{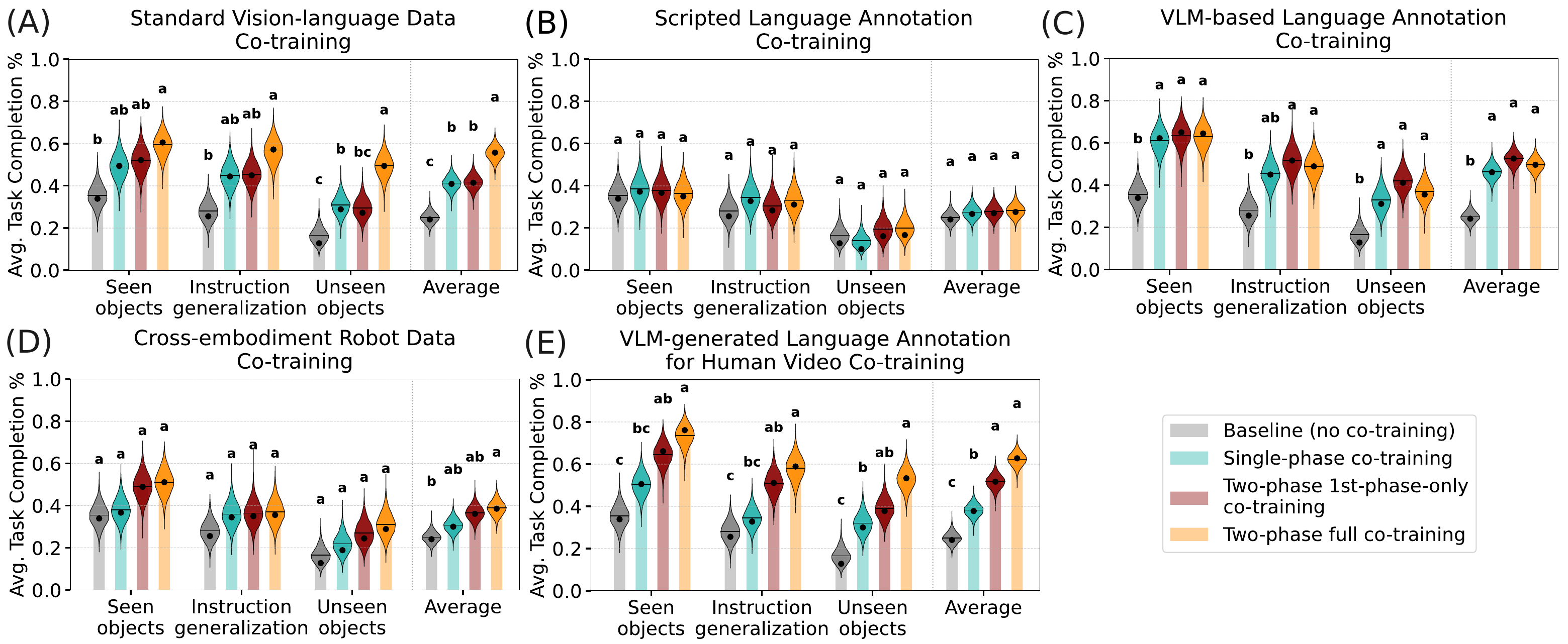}
    \caption{\textbf{Real-world ablation of co-training data and strategies.} 
    Performance of the no-co-training baseline and policies co-trained with a single data modality across training phases, evaluated at language-following with seen objects, instruction generalization, and unseen objects (A--E denote data modalities).}
    \label{fig:real_ablation}
    \vspace{-4mm}
\end{figure*}

We evaluate each co-training data modality using the three strategies described in Section~\ref{sec:cotraining_inference_strategies}: \textit{single-phase co-training}, \textit{two-phase 1st-phase-only co-training}, and \textit{two-phase full co-training}. We compare the policies obtained from these strategies against a baseline policy trained exclusively on continuous robot actions from \anonrep{TRI}{Target}-Ramen (namely, \textit{no-co-training baseline}) using the flow-matching objective ($M_{CE}$=0). All policies are first evaluated in simulation, with results summarized in Fig.~\ref{fig:single_modality_ablation}. Modalities found to be effective are then evaluated in real-world language-following experiments (Fig.~\ref{fig:real_ablation}).

\textbf{Standard Vision-language Data Co-training. }
(1) As shown in Fig.~\ref{fig:single_modality_ablation}A and \ref{fig:real_ablation}A, co-training with standard VL data substantially improves robustness to DS, generalization to unseen tasks, and language following, with no statistically significant change in in-distribution performance (seen tasks).
(2) Fine-tuning the pretrained VLM on our curated VL data enhances its representation for robot manipulation tasks, as evidenced by gains from \textit{two-phase 1st-phase-only co-training} over the baseline (Fig~\ref{fig:single_modality_ablation}A,~\ref{fig:real_ablation}A).
(3) Continuing to co-train with VL data during the 2nd-phase further enhances performance on unseen tasks and language following (particularly with unseen objects). We hypothesize that this continued exposure allows the model to retain the rich, generalizable knowledge from the VL corpus, which is absent in the robot data, thereby preventing catastrophic forgetting.

\textbf{Dense Language Annotations for Robot Data Co-training.}
(1) Co-training with scripted (Fig.~\ref{fig:single_modality_ablation}B,~\ref{fig:real_ablation}B) and VLM-based (Fig.~\ref{fig:single_modality_ablation}C,~\ref{fig:real_ablation}C) annotations for robot data improves the model's robustness to DS, generalization to unseen tasks, and language-following, with no statistically significant change in in-distribution performance.
(2) Owing to greater linguistic diversity and richer descriptions of object-environment interactions, VLM-based annotations yield more substantial improvements on unseen tasks and language following compared to scripted annotations (Fig.~\ref{fig:single_modality_ablation}B-C,~\ref{fig:real_ablation}B-C).
(3) In two-phase co-training, incorporating these annotations during the 2nd-phase yields no additional benefit, suggesting that they are most effective when used exclusively during the 1st-phase (Fig.~\ref{fig:single_modality_ablation}B-C,~\ref{fig:real_ablation}B-C). We posit that, because these annotations describe the same physical trajectories as the robot action data, their utility primarily lies in bootstrapping language-action alignment during the 1st-phase training, rather than introducing new information during the 2nd-phase.

\textbf{Cross-embodiment Robot Data Co-training.}
(1) As shown in Fig.~\ref{fig:single_modality_ablation}D and \ref{fig:real_ablation}D, co-training with cross-embodiment robot data improves robustness to DS, generalization to unseen tasks, and language following, with no statistically significant change in in-distribution performance.
(2) Cross-embodiment robot data is most effective, providing the largest gains, when confined to the first phase of two-phase co-training, particularly for unseen-task generalization and robustness under DS (Fig.~\ref{fig:single_modality_ablation}D). When included during the second phase of training in the \textit{two-phase full co-training}, it yields negligible additional benefit for language-following. We hypothesize that the diverse morphologies and manipulation strategies from cross-embodiment data are most valuable for learning generalizable visual and behavioral representations during the 1st-phase training. In the 2nd-phase, the model benefits from specializing to the target embodiment, at which point continued exposure to other embodiments provides limited value.

\textbf{Human Video Co-training.} \textit{A) Latent Actions:} In \textit{single-phase co-training}, the model jointly learns continuous actions for \anonrep{TRI}{Target}-Ramen and discrete latent action tokens extracted from all video data (\anonrep{TRI}{Target}-Ramen, OXE-Ramen, and human videos). For two-phase methods, the model learns latent actions from all video data in the 1st-phase. In \textit{two-phase full co-training}, it learns both \anonrep{TRI}{Target}-Ramen continuous actions and latent actions from all video data during the 2nd-phase.

\begin{figure}[t]
    \centering
    \includegraphics[width=0.9\linewidth]{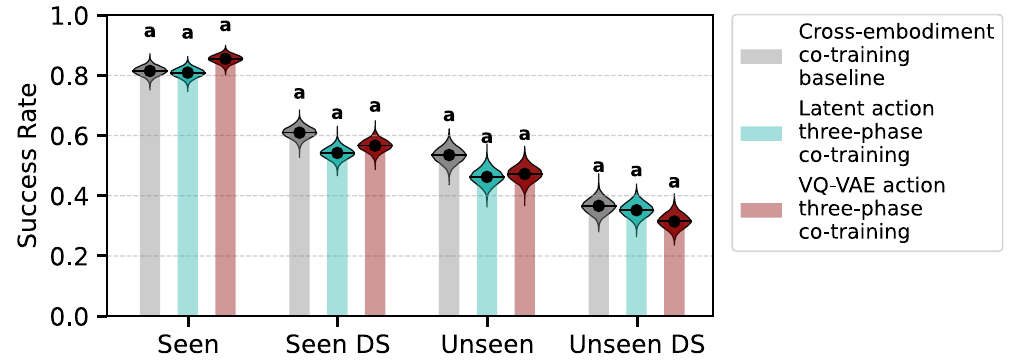}
    \caption{\textbf{Ablation of latent action and VQ-VAE action co-training under data and compute-rich settings.} Simulation results indicate that these modalities show no performance gains over the cross-embodiment baseline.}
    \label{fig:latent_and_vqvae_compute_rich}
    \vspace{-1mm}
\end{figure}

\textit{Single-phase co-training} with latent actions yields no improvement over the baseline (Fig.~\ref{fig:single_modality_ablation}F), whereas \textit{single-phase co-training} with other effective modalities (e.g., standard vision language data) consistently improves performance. On the other hand, for two-phase co-training: (1) latent action 1st-phase training improves performance on unseen tasks, and (2) incorporating latent actions during the 2nd-phase provides no additional benefit. These results suggest that the benefits of latent action 1st-phase training out of two phases may stem from increased computation rather than genuine knowledge transfer. To investigate this, we design a data- and compute-rich setting, comparing two policies: (a) Cross-embodiment co-training baseline: 1st-phase trained on all robot data (\anonrep{TRI}{Target}-Ramen and OXE-Ramen), and 2nd-phase on \anonrep{TRI}{Target}-Ramen; this baseline is equivalent to the \textit{two-phase 1st-phase-only co-training} with cross-embodiment data without latent action co-training. (b) Latent action three-phase co-training: (i) train on latent actions from all video data, (ii) train on all robot data for continuous actions, and (iii) train on \anonrep{TRI}{Target}-Ramen. Notably, the only difference between these two methods is that the latter includes an additional initial training phase on all video data. As shown in Fig.~\ref{fig:latent_and_vqvae_compute_rich}, the added initial latent action training phase provides no benefit in this data- and compute-rich setting.

\begin{figure}[t]
    \centering
    \anonrep{\includegraphics[width=0.9\linewidth]{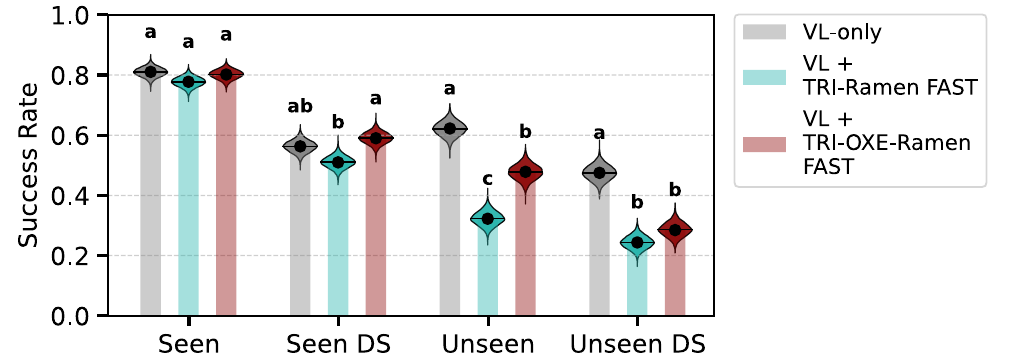}}{\includegraphics[width=0.9\linewidth]{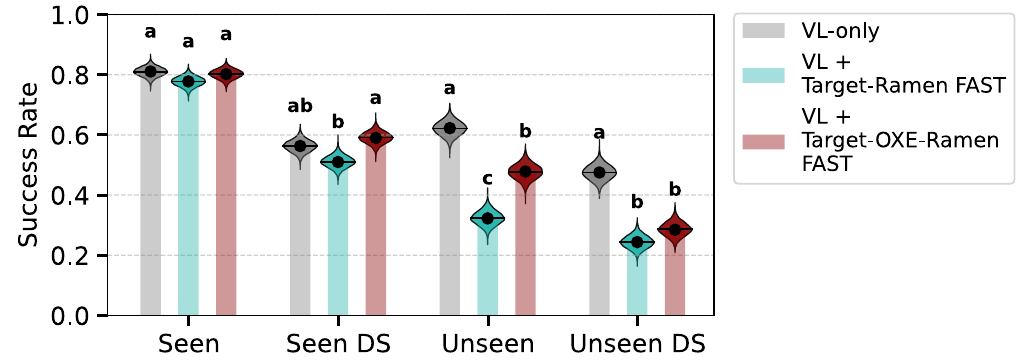}}
    \caption{\textbf{Ablation of FAST token co-training on a broad mix of robot and standard vision-language (VL) data.} FAST co-training along with VL data fails to improve overall performance and degrades generalization on unseen tasks compared to the VL-only co-training baseline.}
    \label{fig:fast_vl}
    \vspace{-4mm}
\end{figure}

\begin{figure*}[t]
    \centering
    \anonrep{\includegraphics[width=\linewidth]{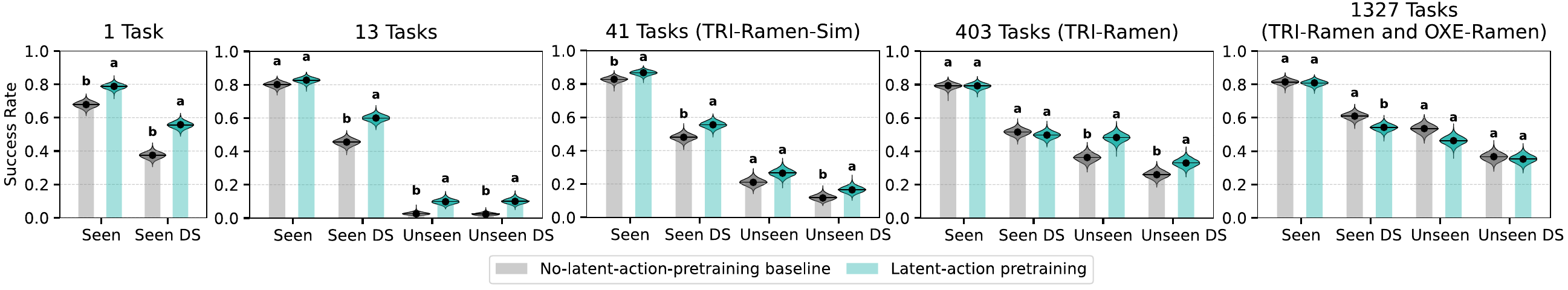}}{\includegraphics[width=\linewidth]{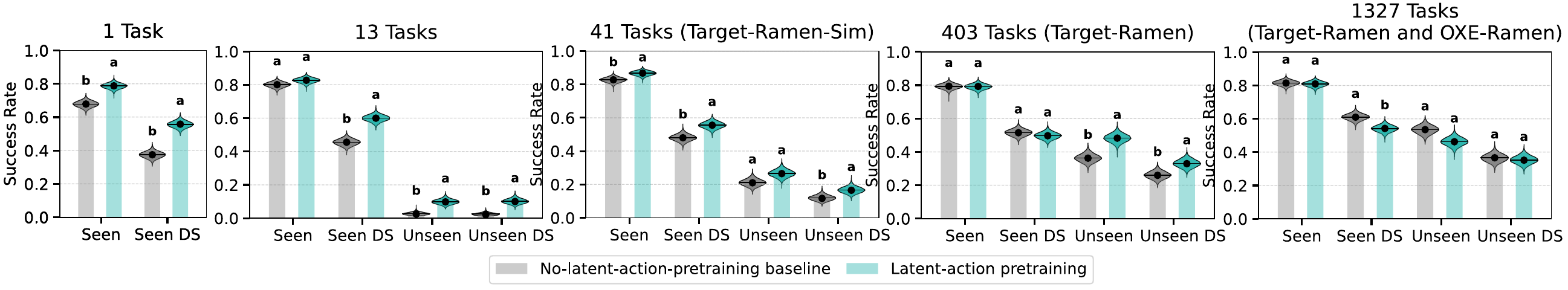}}
    \caption{\textbf{Ablation of latent action co-training on increasing data quantity.} Latent action 1st-phase in a multi-phase training yields performance gains in the low target-robot data regime, with diminishing returns as the quantity of fine-tuning robot data increases.}
    \label{fig:latent_increasing_data}
\vspace{-4mm}
\end{figure*}

Given that prior works~\cite{bu2025univla, bjorck2025gr00t, chen2025villa} highlight the utility of latent action pretraining in a low target-robot-data regime, we further explore its effectiveness across varying scales of robot data, ranging from single-task to the full robot dataset.  Fig.~\ref{fig:latent_increasing_data} shows that latent action 1st-phase training improves performance in the low target-robot-data regime, but these benefits diminish as the quantity of fine-tuning robot data increases.

\textit{B) VLM-generated Annotations:} (1) As shown in Fig.~\ref{fig:single_modality_ablation}E and ~\ref{fig:real_ablation}E, co-training with VLM-generated annotations for human videos improves robustness to DS, generalization to unseen tasks, and language following, with no statistically significant change in in-distribution performance.
(2) In two-phase co-training (Fig.~\ref{fig:real_ablation}E), continuing to incorporate these annotations during the 2nd-phase enhances language-following performance, particularly for unseen objects. We attribute this benefit to the rich diversity of motions, objects, and environments in human videos, which is absent from \anonrep{TRI}{Target}-Ramen. Joint training during the 2nd-phase allows the model to maintain this broader world knowledge, rather than narrowing to the distribution of target robot data.

\textbf{Discrete Robot Action Tokens Co-training.}
\textit{A) FAST Tokens:} Our results suggest that FAST token co-training fails to improve performance across all dimensions and degrades generalization on unseen tasks (Fig.~\ref{fig:single_modality_ablation}G). Prior works~\cite{intelligence2025pi05visionlanguageactionmodelopenworld, driess2025knowledge} show that FAST token co-training could improve performance when pretrained on a broad mix of robot and standard VL data. To examine this claim, we compare three approaches: (a) VL + \anonrep{TRI}{Ramen}-OXE-Ramen FAST: 1st-phase training on all robot data using FAST tokens, alongside VL data. (b) VL + \anonrep{TRI}{Target}-Ramen FAST: 1st-phase training only on \anonrep{TRI}{Target}-Ramen using FAST tokens, alongside VL data. (c) VL-only: 1st-phase training with VL data only. All methods employ an identical 2nd-phase: continuous action learning on \anonrep{TRI}{Target}-Ramen through flow matching, with standard VL data co-training. 

As illustrated in Fig.~\ref{fig:fast_vl}, FAST token co-training fails to improve overall performance and degrades generalization on unseen tasks. However, including OXE-Ramen during the 1st-phase significantly outperforms training on \anonrep{TRI}{Target}-Ramen alone, indicating that FAST co-training might prove beneficial when scaled to substantially larger robot datasets, though it remains ineffective at our current data scale. We attribute this to the nature of FAST tokens as near-lossless action representations: co-training with FAST tokens may bias the VLM backbone toward learning precise action mappings rather than generalizable features.

\textit{B) VQ-VAE Discrete Action Tokens:} (1) VQ-VAE discrete action token co-training yields marginal improvements on unseen tasks but slightly degrades performance under DS (Fig.~\ref{fig:single_modality_ablation}H). (2) Incorporating VQ-VAE tokens during the 2nd-phase of two-phase co-training provides no additional benefit.

Similar to latent action co-training, we examine its effectiveness in a data- and compute-rich setting by comparing a cross-embodiment co-training baseline against VQ-VAE discrete action three-phase co-training policy that begins with learning VQ-VAE discrete action tokens on all robot data during the 1st-phase. As illustrated in Fig.~\ref{fig:latent_and_vqvae_compute_rich}, VQ-VAE co-training yields no improvement in this regime.

\textbf{Summary.}
(1) Co-training with diverse VL data and cross-embodiment robot data substantially enhances the model's generalization to DS, unseen tasks, and language-following capabilities. Notably, owing to their information richness, co-training with standard VL data and language annotations for human videos benefits both 1st-phase and 2nd-phase co-training, whereas language annotations for robot trajectories and cross-embodiment data are primarily effective during 1st-phase in two-phase co-training.
(2) Across all the effective co-training data modalities, standard VL data, VLM-based language annotations for robot data, and language annotations for human videos are the most beneficial. These three specific modalities are all in the form of diverse VL data, suggesting that strengthening VL understanding of the VLM backbone translates into better robot policies.
(3) Discrete action tokens (including latent actions extracted from videos, FAST tokens, and action tokens learned from VQ-VAE) co-training yield no statistically significant performance improvements in our experiments. Specifically, co-training with FAST tokens decreases generalization, while latent actions from videos only provide benefits in the low target-robot-data regime, with benefits diminishing as the proportion of robot data increases.
(4) Across all co-training modalities examined, we observe no statistically significant impact on in-distribution performance.

\begin{figure}[t]
    \centering
    \includegraphics[width=\linewidth]{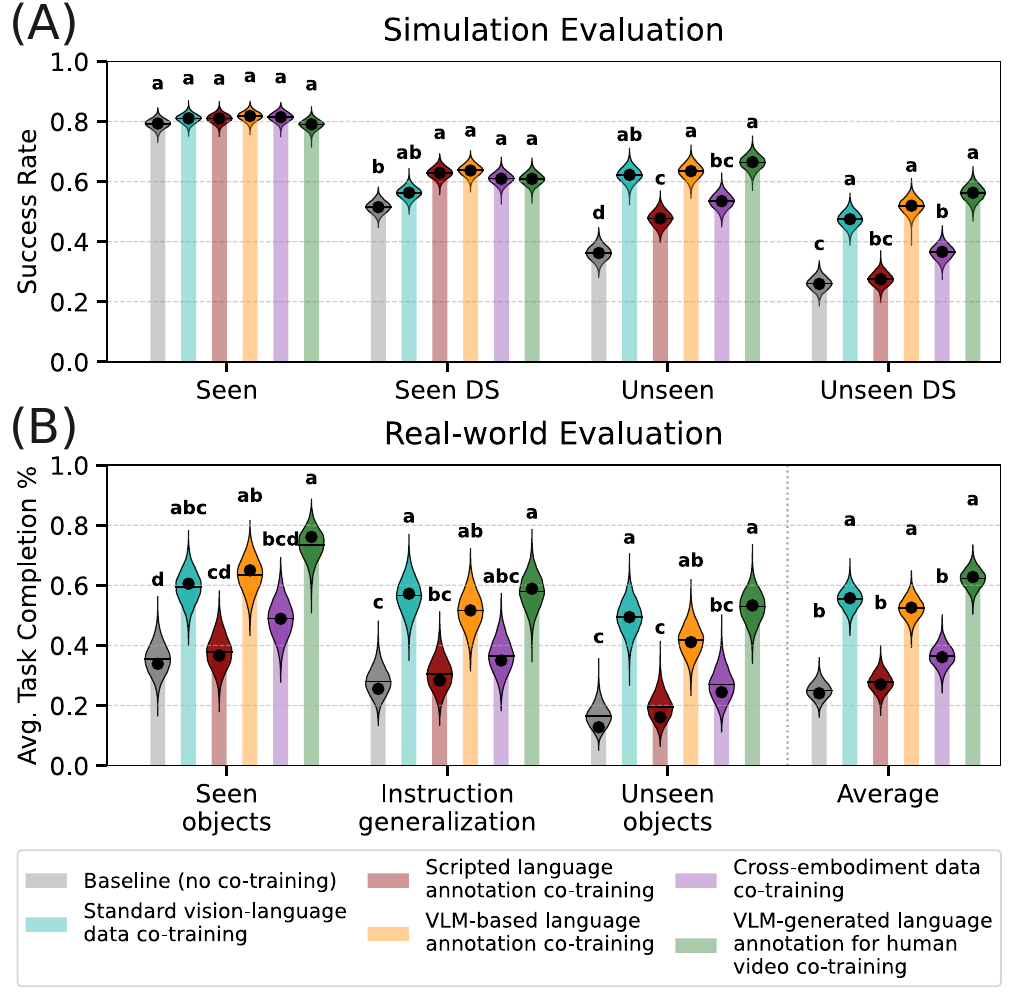}
    \caption{\textbf{Performance of policies trained with the best co-training strategies for effective co-training data modalities.} A) Simulation results on seen and unseen tasks under nominal and distribution-shift conditions. B) Real-world language-following performance on seen objects, instruction generalization, and unseen objects. Diverse vision-language and cross-embodiment robot data substantially enhance the model's generalization to distribution shifts, unseen tasks, and language-following without affecting in-distribution performance.}
    \label{fig:effective_modalities}
    \vspace{-4mm}
\end{figure}

Fig.~\ref{fig:effective_modalities} compares the best training strategies for each useful co-training modality.
Specifically, for standard VL and VLM-generated language annotations for human videos, the best models correspond to \textit{two-phase full co-training}. For scripted and VLM-based language annotation and cross-embodiment robot data, the best models correspond to \textit{two-phase 1st-phase-only co-training}. Co-training with standard VL data, VLM-based annotations for robot data, and annotations for human videos most substantially improves performance on unseen tasks and language following. Furthermore, benefiting from the rich information absent in robot demonstrations, co-training with standard VL data and annotations for human videos more effectively enables the model to recognize unseen objects compared to VLM-based annotations for robot data.

\subsection{Does combining effective co-training modalities yield cumulative performance gains?}
\label{sec:combined_modalities}
Having identified effective co-training data modalities along with their optimal training strategies, we further investigate whether combining these modalities yields cumulative benefits. We conduct an ablation study where we incrementally add each effective data source (training details are provided in Table~\ref{tab:data_batch_ratios_special_policies}): (1) Baseline without co-training, (2) +Vision-Language-Data: Standard VL data co-training only, (3) +Robot-Annotation-Data: Adding dense language annotations (both scripted and VLM-based) for robot data, (4) +Human-Video-Annotation-Data: Adding VLM-generated annotations for human videos, (5) +Cross-Embodiment-Robot-Data (namely, Final Model): Adding cross-embodiment robot data.

\begin{figure}[t]
    \centering
    \includegraphics[width=\linewidth]{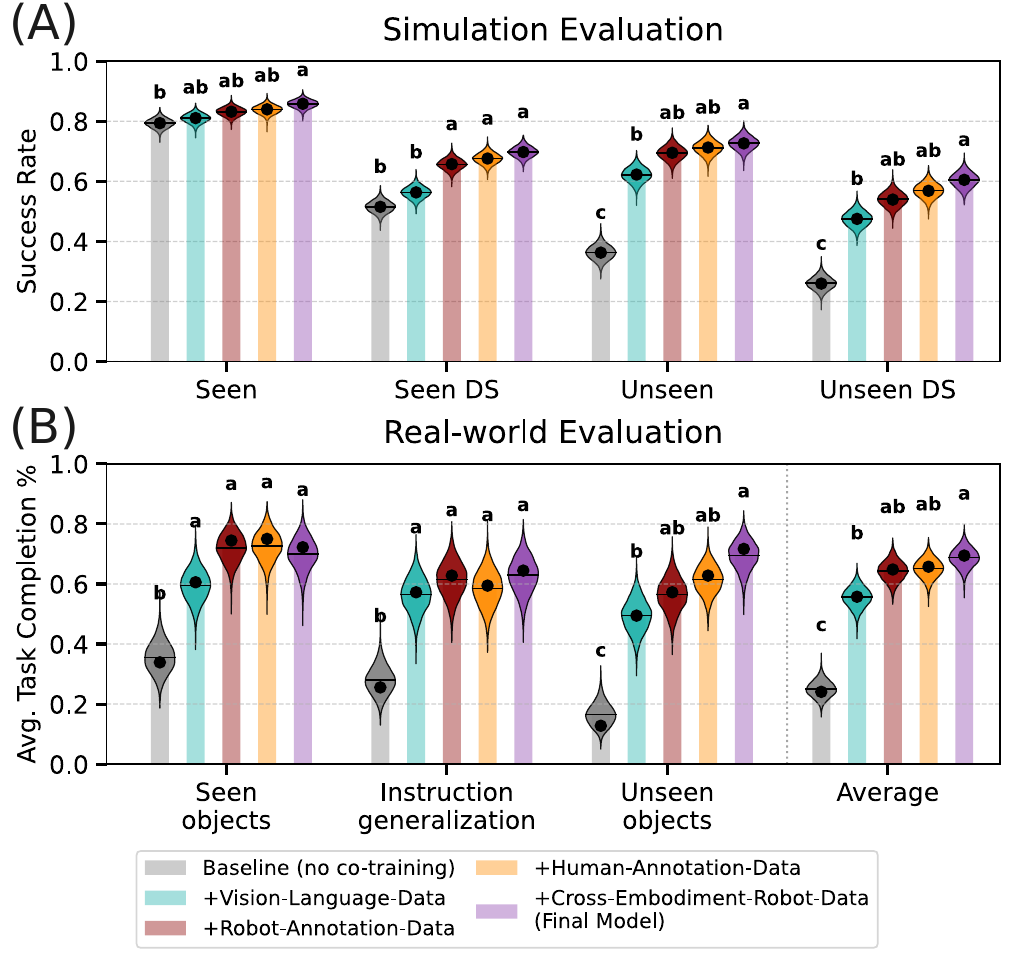}
    \caption{\textbf{Performance of policies co-trained with the effective data modalities additively combined.} A) Simulation results on seen and unseen tasks under nominal and distribution shift conditions. B) Real-world evaluations for language-following performance across seen objects, instruction generalization, and unseen objects settings. Combining the effective co-training data modalities yields cumulative gains in policy performance.}
    \label{fig:combined_modalities}
\vspace{-4mm}
\end{figure}

As shown in Fig.~\ref{fig:combined_modalities}, combining effective co-training modalities yields consistent cumulative performance gains across all evaluation dimensions. Our Final Model achieves strong performance across all settings, attaining a 72.6\% empirical success rate on simulation unseen tasks (36.4\% improvement over baseline) and 69.4\% empirical average task completion on real-world language following (45.3\% improvement over baseline\footnote{Improvement rates are calculated using the empirical means.}). \ifalt In Appendix~\ref{sec:vlm_benchmarking}, we benchmark how the combined effective modalities shape the VLM backbone of our policies. 
\else
\fi

\subsection{Can co-training enhance the quality of learned representations, thereby enabling rapid adaptation to unseen long-horizon, dexterous tasks?}

Prior works~\cite{barreiros2025careful, black2024pi0visionlanguageactionflowmodel, intelligence2025pi05visionlanguageactionmodelopenworld} have demonstrated that high-quality pretraining enables policies to rapidly adapt to downstream unseen tasks through fine-tuning. We investigate whether co-training enhances learned representation quality by fine-tuning (FT) on our suite of unseen long-horizon, dexterous manipulation tasks. We compare three approaches: (1) FT Final Model: fine-tune our Final Model from Section~\ref{sec:combined_modalities} (trained with all effective co-training modalities), (2) FT Baseline: fine-tune the no-co-training baseline model (trained only on \anonrep{TRI}{Target}-Ramen), (3) Single Task: a single-task policy without pretraining.

\begin{figure}[t]
    \centering
    \includegraphics[width=0.9\linewidth]{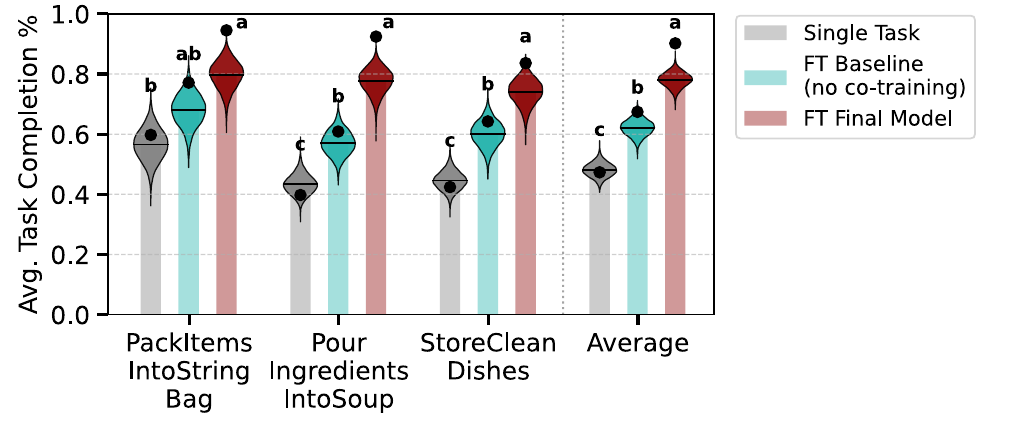}
    \caption{\textbf{Adaptation to unseen long-horizon, dexterous tasks via fine-tuning.} Real-world evaluation of a Single Task policy trained from scratch, a fine-tuned (FT) Baseline pretrained only on robot data, and our FT Final Model pretrained with all effective co-training modalities. The FT Final Model achieves higher average task completion, highlighting the role of co-training in learning transferable, high-quality representations.}
    \label{fig:dexterous}
\vspace{-4mm}
\end{figure}

As shown in Fig.~\ref{fig:dexterous}, benefiting from our curated co-training data and co-training strategies, our Final Model rapidly acquires new skills through fine-tuning, achieving 90.2\% average task completion with only 200 demonstrations---a 22.8\% improvement over the FT Baseline and a 42.9\% improvement over the Single Task policy. This demonstrates that effective co-training substantially enhances representation quality, thereby enabling more fine-grained action learning in downstream tasks. We observe that failures in the FT Baseline and the Single Task policy predominantly stem from insufficient precision in manipulation: they frequently fail to align and secure the cap onto the bottle in \textit{PackItemsIntoStringBag}, misalign the spatula for grasping in \textit{PourIngredientsIntoSoup}, and struggle to grasp the transparent cup in \textit{StoreCleanDishes}. In contrast, the FT Final Model consistently executes these fine-grained manipulations with high precision.

\ifalt
\else
    \ifalt
    \section{How do the effective co-training modalities shape the VLM backbone?}
\else
    \subsection{How do the effective co-training modalities shape the VLM backbone?}
\fi

\label{sec:vlm_benchmarking}

To investigate how the VLM backbone is shaped during co-training, we benchmark the VLMs extracted from our trained VLA policies on a suite of standard vision-language benchmarks using VLMEvalKit~\cite{duan2024vlmevalkit}. We evaluate policies trained with individual effective co-training modalities as well as the policy trained with all effective modalities additively combined across three complementary axes: semantic understanding and reasoning (MMBench, MME, SeedBench), spatial reasoning (RealWorldQA, GQA, SpatialEval), and planning and long-term reasoning (LEGO). We additionally report results for PaliGemma2-PT, the pretrained VLM used to initialize our policies, and PaliGemma2-Mix, a version of the backbone further fine-tuned for instruction following. These baselines contextualize how VLA training and co-training modify the backbone relative to its pretrained and instruction-tuned counterparts.
Fig.~\ref{fig:vlm_benchmarking} shows normalized performance across benchmarks, and we present unnormalized scores in Table~\ref{tab:vlm_benchmark}. Several trends emerge:
(1) The no-co-training baseline performs poorly across nearly all benchmarks. Compared to PaliGemma2-PT and PaliGemma2-Mix, it exhibits substantial degradation, losing all ability to generate language, indicating that training exclusively on robot data erodes the VLM backbone visiolinguistic understanding inherited from pretraining.
(2) Co-training with standard vision-language data leads to strong improvements over the PaliGemma2-PT baseline across most benchmarks, particularly in spatial reasoning and real-world question answering. In contrast, other individual modalities yield no gains when used in isolation. (3) When effective co-training modalities are combined additively, the VLM backbone exhibits consistent, across-the-board improvements, outperforming both the no-co-training baseline and the pretrained PaliGemma2-PT model. The combined model achieves balanced gains across spatial, reasoning, and perception benchmarks, approaching or matching the performance of PaliGemma2-Mix, indicating a more robust and well-rounded multimodal representation. (4) The no-co-training baseline shows both degraded VLM benchmark performance and the weakest generalization in robot tasks, whereas policies co-trained with effective co-training modalities combined improve VLM benchmark scores and also generalize better under DS and to unseen tasks. 

\ifalt
    \begin{table*}[t]
\centering
\caption{VLM backbone evaluation results across vision-language benchmarks.}
\label{tab:vlm_benchmark}
\scriptsize
\setlength{\tabcolsep}{4pt}
\begin{tabular}{p{8cm}cccccccc}
\toprule
\textbf{Model} & \textbf{RealWorldQA} & \textbf{GQA} & \textbf{SpatialEval} & \textbf{MMBench} & \textbf{SEED} & \textbf{MME-P} & \textbf{MME-R} & \textbf{LEGO} \\
\midrule
PaliGemma2-PT (VLA backbone) & 0.24 & 31.68 & 0.26 & 0.01 & 0.24 & 701.56 & 242.50 & 0.03 \\
PaliGemma2-Mix & 0.55 & 59.81 & 0.34 & 0.68 & 0.70 & 1449.43 & 299.29 & 0.27 \\
Baseline (no co-training) & 0.02 & 0.06 & 0.02 & 0.00 & 0.03 & 193.75 & 44.29 & 0.01 \\
Standard Vision Language Data Co-training (+Vision-Language-Data) & 0.39 & 57.69 & 0.28 & 0.26 & 0.49 & 88.24 & 54.64 & 0.27 \\
Scripted Language Annotation Co-training & 0.05 & 0.00 & 0.07 & 0.00 & 0.06 & 80.06 & 32.14 & 0.02 \\
VLM-based Language Annotation Co-training & 0.02 & 0.06 & 0.03 & 0.00 & 0.04 & 92.00 & 31.07 & 0.01 \\
Cross-embodiment Robot Data Co-training & 0.01 & 0.01 & 0.01 & 0.00 & 0.01 & 114.02 & 33.21 & 0.00 \\
VLM-generated Language Annotation for Human Video Co-training & 0.03 & 0.02 & 0.01 & 0.07 & 0.09 & 105.43 & 65.36 & 0.03 \\
+Robot-Annotation-Data & 0.49 & 57.58 & 0.32 & 0.60 & 0.65 & 977.79 & 177.14 & 0.27 \\
+Human-Annotation-Data & 0.48 & 55.78 & 0.39 & 0.33 & 0.51 & 1397.43 & 253.57 & 0.27 \\
+Cross-Embodiment-Robot-Data (Final Model) & 0.47 & 58.02 & 0.33 & 0.56 & 0.63 & 1319.75 & 267.86 & 0.27 \\
\bottomrule
\end{tabular}
\end{table*}

\else
\fi

\begin{figure}[t]
    \centering
    \includegraphics[width=\linewidth]{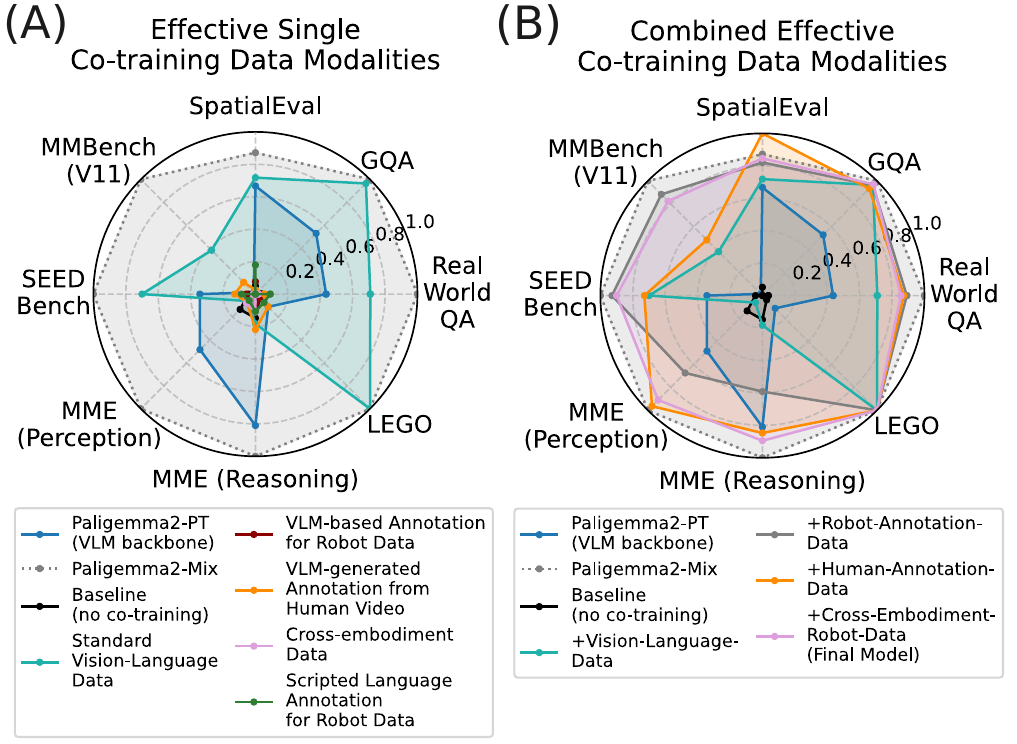}
    \caption{\textbf{VLM backbone benchmarking for policies trained with the effective co-training data modalities.} A) Normalized performance of VLM backbones co-trained with individual effective data modalities compared to the no-co-training baseline, the pretrained PaliGemma2-PT used as backbone of all our policies, and the instruction-tuned PaliGemma2-Mix model, evaluated on standard vision-language, spatial, reasoning, and perception benchmarks. B) Performance of the VLM backbone when effective co-training modalities are combined additively. Combining modalities yields consistent, cumulative improvements across all benchmarks, mirroring gains in downstream robot policy generalization.}
    \label{fig:vlm_benchmarking}
\end{figure}

\fi

\begin{figure}[t]
    \centering
    \includegraphics[width=\linewidth]{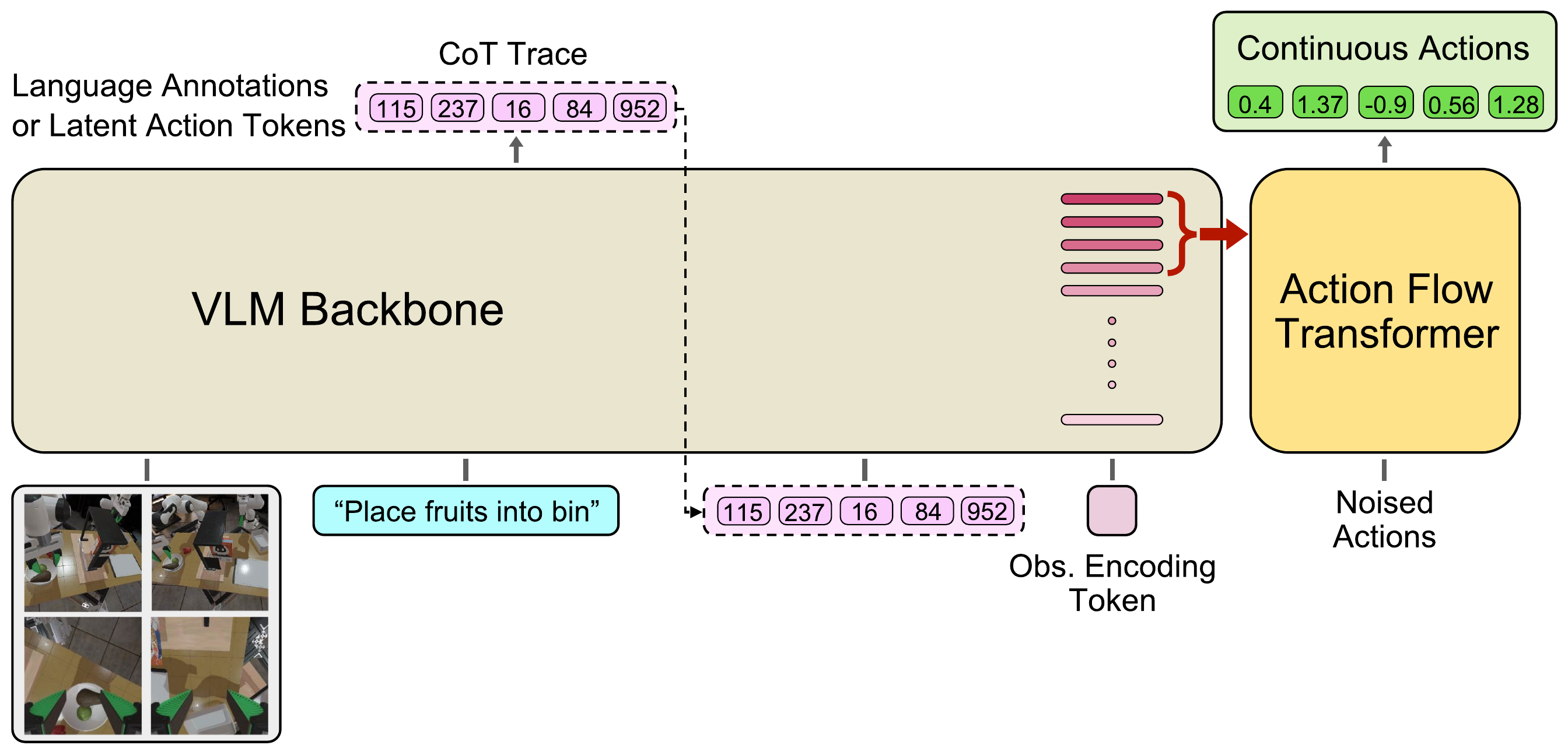}
    \caption{Depiction of the training and inference process for the CoT-conditioned policy.}
    \label{fig:model_cot}
\end{figure}

\ifalt
\else
    \ifalt
    \section{How does explicitly conditioning action generation on Chain-of-Thought (CoT) learned from co-training data affect policy performance?}
\else
    \subsection{How does explicitly conditioning action generation on Chain-of-Thought (CoT) learned from co-training data affect policy performance?}
\fi

\label{sec:cot}
    
\ifalt
    Beyond investigating when to incorporate co-training data, we examine ways to leverage it. While our primary approach treats co-training data solely as auxiliary supervision, recent works~\cite{lin2025onetwovla, zawalski2024robotic, zhai2025igniting, team2025gemini, chen2025training, chen2025villa} suggest that explicitly conditioning action prediction on CoT traces can boost policy performance. We therefore also test this alternative paradigm for selected co-training modalities: language annotations for robot trajectories and latent actions for videos. 

When generating actions conditioned on CoT traces at inference time, the VLM backbone first produces a CoT trace (see Fig.~\ref{fig:model_cot}). The observation token is then appended to its end, and the resulting visiolinguistic embedding (encoding images, task prompts, and CoT traces) is extracted to condition the ActionFT for continuous action prediction. During training, we introduce probabilistic CoT conditioning: with probability $p$, the policy is trained to generate actions conditioned on CoT traces extracted from the co-training data; with probability $1-p$, the policy generates actions directly without CoT conditioning. Importantly, CoT conditioning differs from other co-training strategies only in that the predicted co-training tokens are used directly to form the visiolinguistic embedding, instead of just providing auxiliary supervision. \ifalt \else
    We evaluate the impact of this explicit inference-time CoT approach in Section~\ref{sec:cot}. 
\fi

\else
\fi

We examine whether generating intermediate CoT traces to condition action generation provides any advantage over standard training. Specifically, we evaluate three co-training data types that naturally yield CoT-like intermediate contents: (1) scripted annotations for robot data, (2) VLM-based annotations for robot data, and (3) latent actions from videos.

For each co-training data type, we train three additional policies that differ in their CoT conditioning strategies. All three methods follow the same 1st-phase training procedure on the corresponding co-training data as described in Section~\ref{sec:cotraining_inference_strategies}. During the 2nd-phase, all methods jointly learn from both continuous robot actions and co-training data. Specifically: (1) 50\%-CoT Training-Only: During the 1st-phase, the model conditions action generation on CoT with 50\% probability. At inference time, actions are generated directly without CoT. (2) 50\%-CoT with Inference: Training procedure is identical to (1). At inference time, the model first generates CoT, then conditions subsequent action generation on it. (3) 100\%-CoT: In the 2nd-phase, the model always conditions action generation on CoT (100\% probability). At inference time, the model generates CoT and conditions action generation on it.

We compare these explicit CoT strategies against the implicit two-phase co-training approaches discussed in Section~\ref{sec:cotraining_inference_strategies}, and the baseline policy trained without co-training. As shown in Fig.~\ref{fig:cot_ablation}, while explicit CoT conditioning demonstrates improvements over the baseline in certain settings (e.g., co-training with VLM-based annotations for robot data on unseen tasks), it consistently fails to improve upon the implicit two-phase co-training approaches across all settings, with discernible performance degradation when using VLM-based annotations and latent actions as CoT sources. This lack of improvement can be attributed to the nature of our evaluation tasks. Prior works~\cite{lin2025onetwovla, zhai2025igniting, team2025gemini} demonstrating benefits from explicit CoT generation typically evaluate on tasks requiring multi-step planning or complex semantic reasoning. In contrast, our simulation benchmark focuses on manipulation tasks with clear objectives and immediate visual feedback, where the mapping from observation to action is relatively direct. In such settings, the implicit reasoning learned during co-training appears sufficient, making explicit CoT generation redundant.

The observed performance degradation, particularly pronounced with VLM-based annotations and latent actions, likely stems from two factors. First, these co-training data sources may contain inherent inaccuracies—VLM-generated annotations can misinterpret visual scenes or actions, while latent action models may capture irrelevant visual changes such as background variations rather than true action semantics. Second, these two sources produce richer and more complex CoT content compared to scripted annotations. When explicitly conditioning action generation on such CoT, any errors or imprecision in the generated CoT directly propagate to the subsequent action prediction, compounding the initial inaccuracies and leading to less precise manipulation behavior.

\begin{figure}[t]
    \centering
    \includegraphics[width=\linewidth]{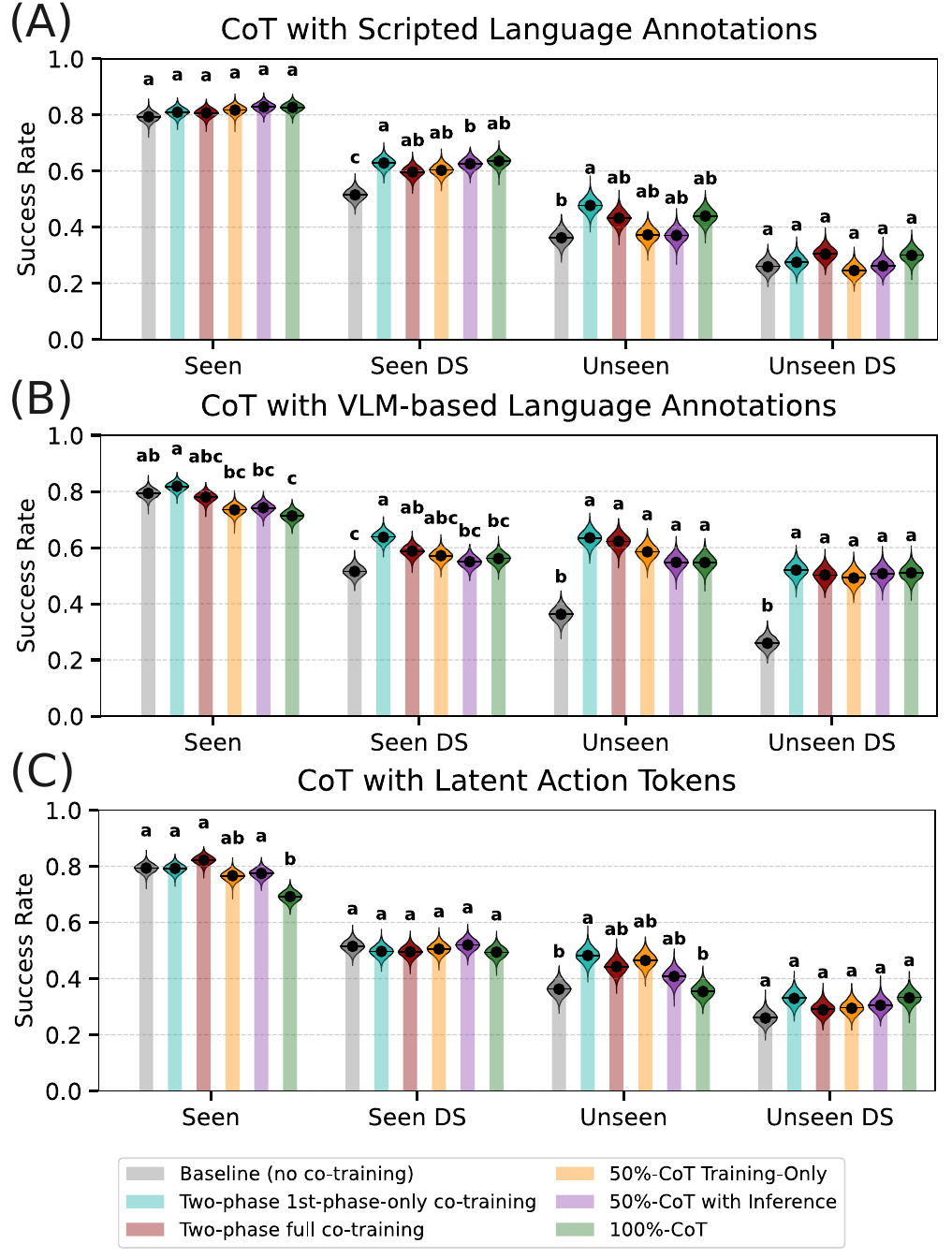}
    \caption{\textbf{Simulation ablation of CoT-conditioned action generation.} Evaluation of policies trained to explicitly condition action generation on chain-of-thought (CoT) traces derived from A) scripted language annotations for robot data, B) VLM-based language annotations for robot data, and C) latent action tokens from videos. Action generation conditioned on CoT traces from co-training data yields no improvement in our simulation benchmark.}
    \label{fig:cot_ablation}
\end{figure}    
\fi

\section{Related Work}

\subsection{Large Behavior Models}
Developing a general-purpose policy capable of perceiving, understanding, and acting in the physical world is a central objective in robotics. Early robot learning systems rely on task-specific policies trained on limited robot demonstrations~\cite{levine2016end, finn2017one, xu2023roboninja}, which constrain their ability to generalize beyond narrow training distributions. In contrast, LBMs scale imitation learning, both in model capacity and dataset size, and have shown impressive performance on dexterous tasks~\cite{barreiros2025careful}. A prominent class within LBMs is VLAs~\cite{kim2024openvla, zitkovich2023rt, driess2023palm, black2024pi0visionlanguageactionflowmodel, intelligence2025pi05visionlanguageactionmodelopenworld, driess2025knowledge, team2503gemini, team2025gemini, bjorck2025gr00t, wen2025tinyvla, wen2025dexvla}, which integrate pretrained VLM~\cite{beyer2024paligemma, steiner2024paligemma, liu2024improved, wang2024qwen2} backbones. Representative examples include VLMs paired with an action head as in~\cite{black2024pi0visionlanguageactionflowmodel, intelligence2025pi05visionlanguageactionmodelopenworld, shukor2025smolvla} and fully autoregressive models~\cite{lee2025molmoact, pertsch2025fast, goyal2025vla}. Despite data scaling efforts, VLAs exhibit a limited generalization~\cite{barreiros2025careful, zhang2025vlabench, zhou2025libero} to new objects, environments, and instructions compared to non-embodied foundation models like VLMs. This generalization gap primarily stems from the vast difference in training data availability~\cite{goldberg2025good, o2024open}: robot datasets remain orders of magnitude smaller than the internet-scale text and image corpora used for VLMs.

\subsection{Co-training for Robot Learning}
To bridge the gap between limited robot data and internet-scale multimodal resources, numerous studies have employed co-training with diverse data modalities. 

Public VL datasets~\cite{chen2015microsoft, tong2024cambrian, yu2024capsfusion, deitke2024molmo, yuan2024robopoint, zhou2025roborefer}, which are rich in commonsense knowledge and naturally compatible with VLA architectures, are widely adopted for co-training robot policies~\cite{intelligence2025pi05visionlanguageactionmodelopenworld, zhou2025chatvla, zhou2025chatvla2, lee2025molmoact}. Beyond public datasets, recent works~\cite{lin2025onetwovla, yang2025vlaser, qu2025eo, zhai2025igniting} construct embodied reasoning VL datasets directly on robot trajectories, incorporating rich planning and spatial information. These efforts show evidence that co-training with diverse vision-language data enhances generalization~\cite{intelligence2025pi05visionlanguageactionmodelopenworld, lin2025onetwovla} and improves VLM’s learned representations~\cite{chen2025training, yang2025vlaser} for manipulation tasks.

Several works have explored using cross-embodiment robot data for co-training. Some efforts~\cite{bauer2025latent, doshi2024scaling, yang2023polybot, yang2024pushing, bohlinger2024one} train a single policy directly on multiple robot embodiment data, enabling a unified model to operate with diverse morphologies. \cite{black2024pi0visionlanguageactionflowmodel, intelligence2025pi05visionlanguageactionmodelopenworld, driess2025knowledge, liu2024rdt, liu2025hybridvla} incorporate cross-embodiment data during pretraining to learn generalizable, embodiment-agnostic representations that are subsequently adapted to a specific target robot. \cite{team2024octo, barreiros2025careful, kim2024openvla, kim2502fine, shi2025hi} leverage the Open X-Embodiment dataset~\cite{o2024open}, a large-scale aggregation of demonstrations from many robot platforms, aiming to improve robustness and generalization.

Compared to robot data that typically requires teleoperation for collection, human videos offer a more scalable data source. A number of works~\cite{lepert2025phantom, luo2025being, kareer2025egomimic, lepert2025masquerade, liu2025immimic} explicitly extract action labels (e.g., hand poses) from videos for policy co-training; however, obtaining accurate labels often requires additional sensing modalities, such as VR devices~\cite{qiu2025humanoid, yuan2025motiontrans} or wearable exoskeletons~\cite{xu2025dexumi, tao2025dexwild}. Another line of research~\cite{bu2025learning, ye2024latent, chen2025villa, bu2025agibot, chen2024igor, bjorck2025gr00t} explores latent action representations by extracting discrete action tokens from video frames with methods such as VQ-VAE~\cite{van2017neural}, which can serve as a unified representation across different embodiments that encode motion information. However, these approaches have been validated only in low target robot data regimes.

Discretizing continuous robot actions into tokens allows policies to treat action generation as a sequence modeling problem. Beyond the naive per-dimension binning approach~\cite{kim2024openvla, zitkovich2023rt}, advanced methods~\cite{pertsch2025fast, wang2025vq} employ frequency-based techniques (e.g., FAST tokenizer) or vector quantization (e.g., VQ-VAE) to compress the action space. Notably, relying on discrete tokens for low-level control often results in limited precision~\cite{driess2025knowledge} and slow inference~\cite{pertsch2025fast, lee2025molmoact}. To mitigate this, recent approaches~\cite{intelligence2025pi05visionlanguageactionmodelopenworld, driess2025knowledge, jiang2025galaxea, zhai2025igniting} utilize these tokens solely for pre-training or co-training objectives, while retaining an action head for continuous action generation. This strategy has been shown to enhance sample efficiency and generalization.

\subsection{Chain-of-thought for Robot Control}
Inspired by the substantial benefits that CoT has brought to language models when dealing with complex tasks~\cite{guo2025deepseek, jaech2024openai, wei2022chain, zhang2022automatic}, recent works~\cite{zawalski2024robotic, intelligence2025pi05visionlanguageactionmodelopenworld, chen2025training, li2025hamster} have explored adapting CoT for robot control. Specifically, these approaches first generate intermediate content before conditioning action generation on it. This intermediate content can be linguistic, such as subtask decomposition or visual grounding information (e.g., object locations)~\cite{lin2025onetwovla, zhai2025igniting, team2025gemini, hancock2025actions, yang2025instructvla}, or control-centric, such as end-effector movements~\cite{lee2025molmoact, belkhale2024rt} and latent actions~\cite{bu2025univla, chen2025villa}. While CoT has been shown to be beneficial for long-horizon tasks or those requiring complex reasoning~\cite{lin2025onetwovla, zhai2025igniting, team2025gemini}, there is limited empirical evidence comparing (i) explicitly conditioning the policy on predicted CoT traces, versus (ii) using the same CoT content solely as an auxiliary co-training objective. This evidence is especially insufficient for manipulation tasks with clear and well-defined goals.

\section{Discussion, Limitations, And Future Work}
We present a large-scale empirical study that systematically dissects the impact of diverse co-training data and strategies on the performance of LBMs. Our findings reveal that co-training with vision-language data and cross-embodiment robot data substantially enhances generalization to DS, unseen tasks, and language following capabilities, while discrete action token variants yield no statistically significant benefits. Furthermore, we show that combining effective modalities produces cumulative performance gains and enables rapid adaptation to dexterous, long-horizon tasks via fine-tuning. 

Notably, among all useful co-training modalities, diverse vision-language data—including standard datasets and rich annotations for robot and human videos—demonstrate the most substantial improvements. This observation resonates with the Good Regulator Theorem~\cite{conant1970every}, which states that a system must incorporate an internal model (implicit or explicit) of its operating world to effectively regulate it. In our setting, strong foundation models (such as VLMs) provide precisely such internal models that have rich semantic and spatial understanding of the physical world. Our results suggest that progress towards truly generalist robot policies is intrinsically linked to advances in these foundation models. \ifalt
\else Specifically, the VLM benchmarking results corroborate this interpretation: co-training with effective data modalities not only improves downstream robot performance but also preserves the visiolinguistic reasoning, spatial understanding, and perception capabilities of the VLM backbone itself. While this pattern is consistent across the models we evaluate, further investigation is needed to more rigorously characterize the relationship between backbone visiolinguistic understanding and policy generalization. Interestingly, we find that explicitly conditioning action generation on CoT learned from co-training data provides no benefit for our manipulation tasks, which have clear immediate objectives, suggesting that implicit reasoning learned during co-training suffices for such settings.
\fi 

While our study provides promising insights, several limitations should be acknowledged. First, while we examine various sources of vision-language data, we do not systematically analyze their impact by task taxonomy (e.g., visual question answering, image captioning, object detection, spatial reasoning). Understanding how different vision-language task categories affect specific policy capabilities would enable more targeted and sample-efficient data curation. Second, we explore only coarse-grained representations for human videos through latent actions and language annotations. As hand pose estimation techniques advance and dexterous robotic hands continue to evolve, explicitly extracting fine-grained dexterous motions from human video may become a valuable co-training signal. \ifalt
\else
    Third, our exploration of CoT is limited to forms naturally arising from our co-training data—primarily low-level action abstractions lacking high-level planning or complex reasoning. Future work could investigate richer CoT formulations, such as history and reflection traces or hierarchical plans for tasks requiring complex decision-making. 
\fi Finally, our study focuses exclusively on imitation learning; exploring co-training within alternative learning paradigms, such as world modeling or reinforcement learning, remains an open frontier for developing scalable generalist policies.

\section*{Acknowledgments}
\anonrep{
We thank Muhammad Zubair Irshad, Vitor Campagnolo Guizilini, Sedrick Keh, and Swati Gupta for the manuscript reviews; Kosei Tanada, Takuto Ito, Phoebe Horgan, Gordon Richardson, and Aykut Onol for evaluation quality assurance; Sam Shereda, Mariah Smith-Jones, and Matthew Tran for evaluation and data collection; Alejandro Castro and Joseph Masterjohn for simulation support, and to the TRI's PROPS team for assistance with experimental setup.
}{[anonymized]}

\bibliographystyle{plainnat}
\bibliography{references}

\input{appendix_preamble_arxiv}

\section{Model Architecture and Hyperparameter Ablation}

\subsection{Model Architecture Ablation.} 
\label{sec:model_ablation}

\ifalt 
    \ifalt Our ActionFT is a diffusion transformer consisting \else It consists \fi of 8 flow transformer layers, each conditioned on observation features and the flow-matching timestep via an adaptive layer norm (adaLN) MLP~\cite{peebles2023scalable}. The ActionFT receives the global conditioning embedding, a noised chunk of continuous actions, and the flow matching time variable, and is trained to predict the flow vector that guides iterative denoising of the actions toward the target trajectory. Unlike prevalent methods~\cite{black2024pi0visionlanguageactionflowmodel}, our approach uses only a single token as visiolinguistic representations rather than attention keys and values from all VLM layers.
\else
\fi

\begin{figure}[t]
    \centering
    \includegraphics[width=\linewidth]{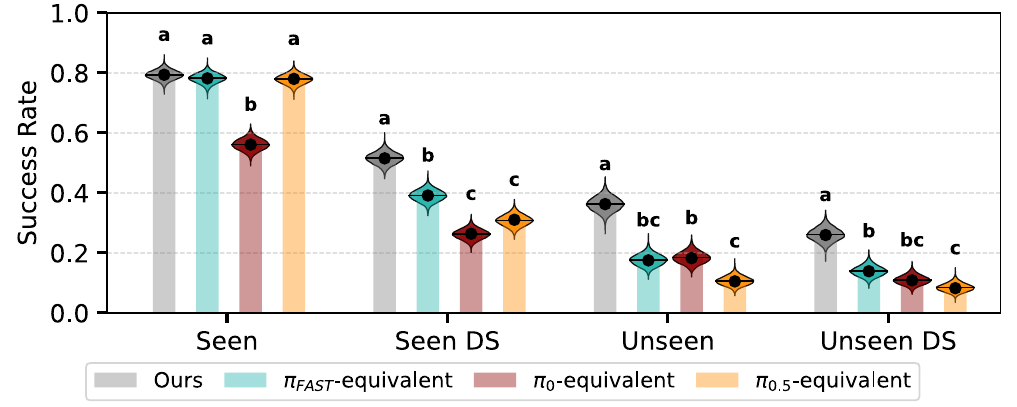}
    \caption{Architecture ablation in simulation comparing our model to $\pi_{0}$-, $\pi_{0.5}$-, and $\pi_{FAST}$-equivalent variants across seen and unseen tasks, with and without distribution shift.}
    \label{fig:arch_ablation}
\end{figure}

We present ablation results comparing four model architectures: (1) Ours: Comprises a VLM and a flow transformer action head (ActionFT) (2) $\pi_{FAST}$-equivalent~\cite{pertsch2025fast}: Consists solely of a VLM that auto-regressively generates FAST tokens, which are further decoded into continuous robot actions. (3) $\pi_0$-equivalent: Comprises a VLM and an action expert as formulated in~\cite{black2024pi0visionlanguageactionflowmodel}. All attention keys and values from all tokens across VLM layers are used as visual-linguistic representations for the action expert. (4) $\pi_{0.5}$-equivalent~\cite{intelligence2025pi05visionlanguageactionmodelopenworld}: This is similar to $\pi_0$-equivalent with VLM and action expert, but the flow-matching timestep embedding is injected at each layer of the action expert, whereas in $\pi_0$-equivalent, the timestep embedding is injected only at the first layer together with the noised action.
All four architectures use the same pretrained VLM (PaliGemma2-PT) and are trained on \anonrep{TRI}{Target}-Ramen data with identical hyperparameters, except that $\pi_0$-equivalent and $\pi_{0.5}$-equivalent use batch size of 112 for 230k steps due to higher memory consumption, while Ours and $\pi_{FAST}$-equivalent use batch size of 128 for 200k steps. As shown in Fig.~\ref{fig:arch_ablation}, while $\pi_{FAST}$-equivalent and $\pi_{0.5}$-equivalent achieve comparable in-distribution performance to Ours, they exhibit significantly degraded performance on distribution shifts and unseen tasks, demonstrating that our more compact representation enhances generalization. Additionally, $\pi_0$-equivalent substantially underperforms on in-distribution tasks, indicating the importance of injecting flow-matching timestep embeddings at each layer of the action network.

\subsection{Hyperparameter Ablation.} 
\label{sec:hyperparameter_ablation}

We study two hyperparameters that are empirically critical for effective training: the weighting factor used to balance the flow-matching loss and the cross-entropy loss, and the ratio of target robot data to co-training data within each batch.

\begin{figure}[t]
    \centering
    \includegraphics[width=\linewidth]{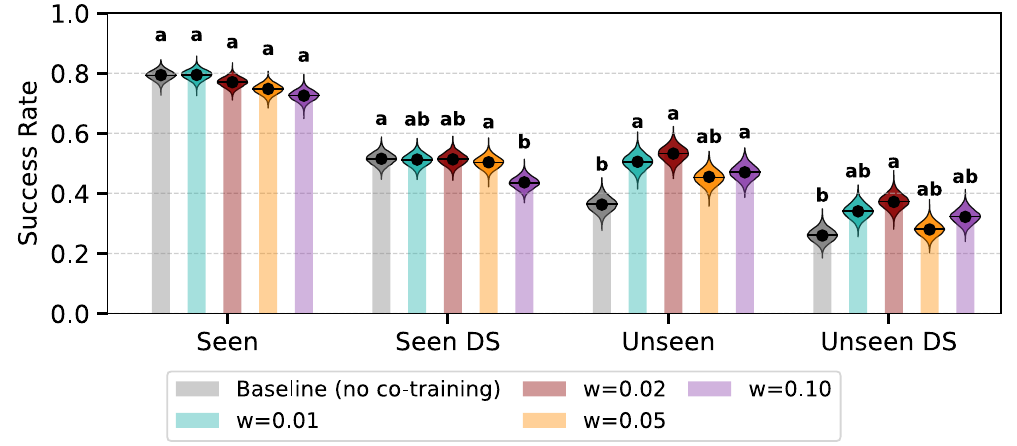}
    \caption{Ablation of the cross-entropy loss weight $w$ under \textit{single-phase co-training} with standard vision-language data.}
    \label{fig:ce_ablation}
\end{figure}

\begin{figure}[t]
    \centering
    \includegraphics[width=\linewidth]{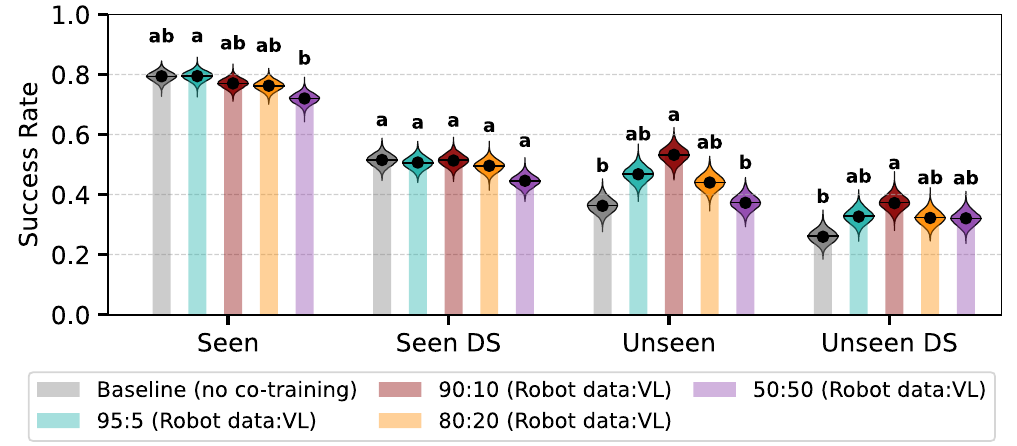}
    \caption{Ablation of the robot-to-co-training data ratio per batch under \textit{single-phase co-training} with standard vision-language data. }
    \label{fig:batch_balance}
\end{figure}

We analyze these hyperparameters under the vision-language data \textit{single-phase co-training} setting. As shown in Figs.~\ref{fig:ce_ablation} and~\ref{fig:batch_balance}, increasing either the weighting factor or the co-training data ratio degrades in-distribution performance, while setting these parameters too low diminishes the generalization benefits from co-training. We adopt $w=0.02$ and a 9:1 ratio of robot data to co-training data for co-training with modalities whose samples are not paired one-to-one with target robot data (i.e., standard vision-language data, cross-embodiment robot data, and human videos). This configuration preserves in-distribution performance while maximizing generalization improvements. 

For modalities paired with target robot data, such as language annotations or discrete action tokens for robot data, the model learns from both the continuous actions and the corresponding co-training sample. Cross-embodiment robot data constitutes a special case, as the learning objective is still flow matching over robot actions; in this setting, target robot and cross-embodiment data are mixed during the 1st-phase training using a 6:4 batch ratio. 

\section{Training Details}
\label{sec:training_details}
Each phase of multi-task VLA policy training runs for 200k steps with a default batch size of 128, using a cosine learning rate schedule that decays from 2e-5 to 2e-6 over the last 60k steps. Each training phase requires approximately 64 hours on 16 H100 GPUs in this default setting. The ratios of robot data to co-training data follow Appendix~\ref{sec:hyperparameter_ablation}. Exceptions include: (1) the latent action 1st-phase training on all video data uses a batch size of 256 due to the large video volume, (2) for multi-modality co-training in Section~\ref{sec:combined_modalities}, to ensure the effective sample count per modality remains consistent with single-modality co-training policies, we increase batch size in certain training phases. The data modalities, data ratios, and batch sizes used at each training phase for these special VLA policies are listed in Table~\ref{tab:data_batch_ratios_special_policies}. 
For VLA policies trained or fine-tuned on a single task, we train for 30k steps and decay the learning rate from 2e-5 to 2e-6 with a cosine schedule over the last 24k steps. All other shared training hyperparameters are listed in Table~\ref{tab:shared_hparams}.

\begin{table*}[t]
\centering
\caption{Data modalities, batch data ratios, and batch sizes (BS) used at each training phase for the special policies.}
\label{tab:data_batch_ratios_special_policies}
\scriptsize
\setlength{\tabcolsep}{4pt}  
\begin{tabular}{p{3.4cm} p{4.3cm} p{4.3cm} p{4.3cm}}
\toprule
\textbf{Policy} & \textbf{1st-phase} & \textbf{2nd-phase} & \textbf{3rd-phase} \\
\midrule

\textbf{Three-phase latent action co-training} &
\anonrep{TRI}{Target}-Ramen : OXE-Ramen : Human videos = 3 : 3 : 4; BS = 256 &
\anonrep{TRI}{Target}-Ramen : OXE-Ramen = 6 : 4; BS = 128 &
\anonrep{TRI}{Target}-Ramen : OXE-Ramen = 9 : 1; BS = 128 \\

\midrule

\textbf{VL + \anonrep{TRI}{Ramen}-OXE-Ramen FAST} &
\anonrep{TRI}{Target}-Ramen (FAST) : Standard vision-language data : OXE-Ramen (FAST) = 5 : 2 : 3; BS = 128 &
\anonrep{TRI}{Target}-Ramen : Standard vision-language data = 9 : 1; BS = 128 &
--- \\

\midrule

\textbf{+Robot-Annotation-Data} &
Standard vision-language data : Language annotations for \anonrep{TRI}{Target}-Ramen = 1 : 1; BS = 256 &
\anonrep{TRI}{Target}-Ramen : Standard vision-language data = 9 : 1; BS = 128 &
--- \\

\midrule

\textbf{+Human-Annotation-Data} &
Standard vision-language data : Language annotations for \anonrep{TRI}{Target}-Ramen : Language annotations for human videos = 1 : 1 : 1; BS = 384 &
\anonrep{TRI}{Target}-Ramen : Standard vision-language data : Language annotations for human videos = 9 : 0.5 : 0.5; BS = 128 &
--- \\

\midrule

\textbf{+Cross-Embodiment-Robot-Data (Final Model)} &
Standard vision-language data : Language annotations for \anonrep{TRI}{Target}-Ramen : Language annotations for human videos = 1 : 1 : 1; BS = 384 &
\anonrep{TRI}{Target}-Ramen : Standard vision-language data : OXE-Ramen : Language annotations for human videos = 4 : 1 : 4 : 1; BS = 256 &
\anonrep{TRI}{Target}-Ramen : Standard vision-language data : Language annotations for human videos = 9 : 0.5 : 0.5; BS = 128 \\

\bottomrule
\end{tabular}
\normalsize
\end{table*}

\begin{table}[t]
\centering
\caption{Training hyperparameters shared across all policies.}
\label{tab:shared_hparams}
\begin{tabular}{p{4cm} p{4cm}}
\toprule
\textbf{Hyperparameter} & \textbf{Value} \\
\midrule
Image observation horizon & 1 \\
Camera number (\anonrep{TRI}{Target}-Ramen) & 4 (2 scene cameras, 1 wrist camera per robotic arm) \\
Image augmentation &
Random crop (for \anonrep{TRI}{Target}-Ramen, $256 \times 342 \rightarrow 224 \times 224$);
color jitter (brightness $=0.3$, contrast $=0.4$, saturation $=0.5$, hue $=0.05$) \\
Learning rate warm-up steps & 1000 \\
Learning rate warm-up scheduler & linear \\
Learning rate decay scheduler & cosine decay \\
\bottomrule
\end{tabular}
\end{table}

\begin{table}[t]
\centering
\caption{Summary of the hours and number of VLM-generated annotation data samples for human video data.}
\label{tab:human_video_annotation_stats}
\begin{tabular}{l c c}
\toprule
\textbf{Dataset} & \textbf{Video Hours} & \textbf{Annotation Data Samples} \\
\midrule
Ego4D & 774.5 hours & 5.2M \\
EgoDex & 744.4 hours & 3.0M \\
EgoDex (reversed) & 455.7 hours & -- \\
Something-Something V2 & 155.8 hours & 0.8M \\
Epic Kitchen & 60.4 hours & -- \\
HoloAssist & 80.8 hours & -- \\
\midrule
\textbf{Total} & \textbf{2271.6 hours} & \textbf{9.0M} \\
\bottomrule
\end{tabular}
\end{table}

\section{Data Processing and Curation Details}

\ifalt
    An overview of the training data is illustrated in Fig.~\ref{fig:data}.
    \begin{figure*}[t]
    \centering
    \includegraphics[width=\linewidth]{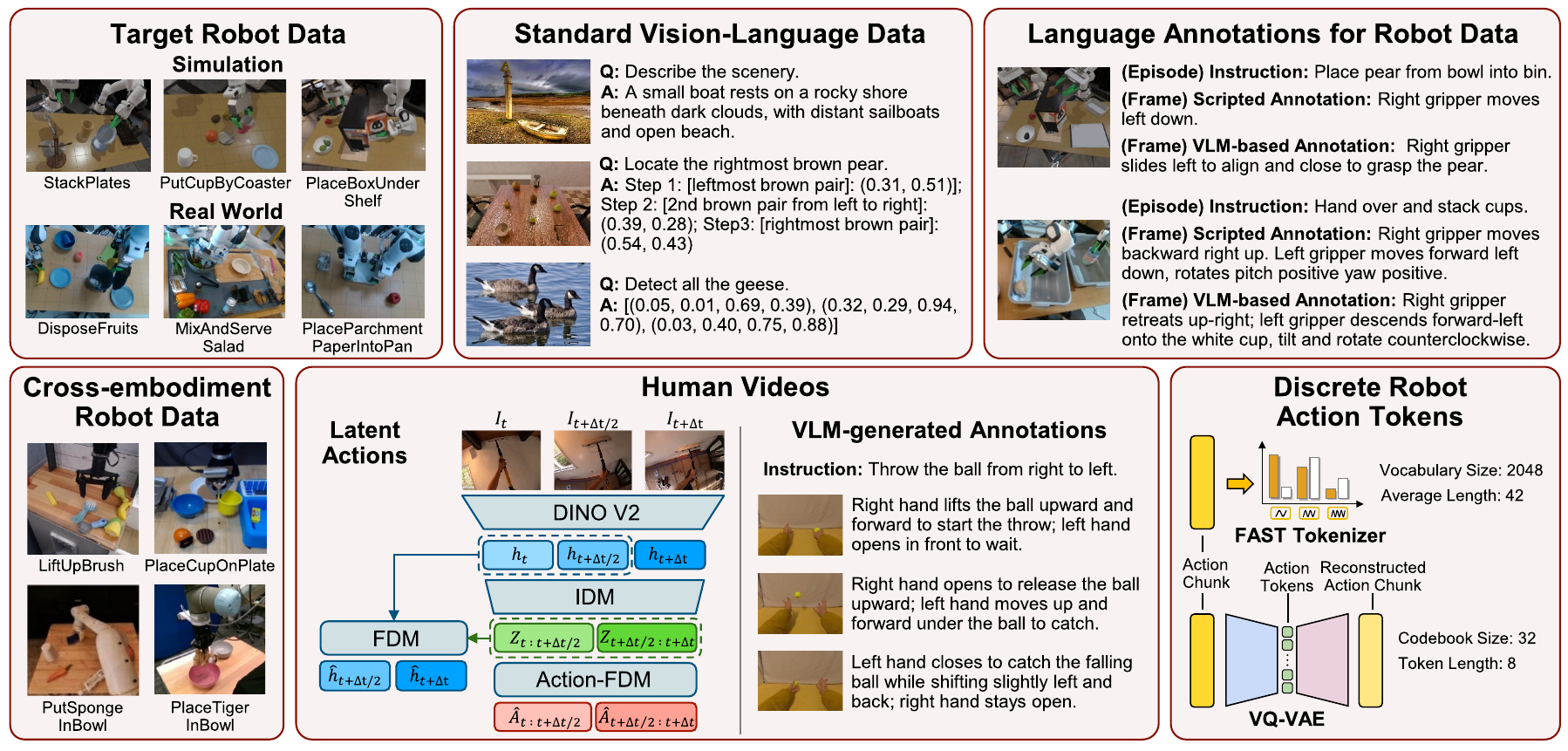}
    \caption{\textbf{Overview of the training data.} Our dataset comprises target robot data collected in both simulation and real-world, and five heterogeneous co-training data modalities: standard vision-language data for commonsense understanding, spatial reasoning, and object grounding; dense language annotations for robot data, generated via heuristic scripting and VLM-based captioning; cross-embodiment robot data capturing diverse robot morphologies and manipulation tasks; human videos, from which we derive either latent action tokens using a latent action model (LAM) or VLM-generated annotations; and discrete robot action tokens, including near-lossless FAST tokens and compact VQ-VAE tokens. These co-training modalities, together with the target robot data, constitute a unified dataset of $\sim$4,000 hours of manipulation data and 50M vision-language samples.}
    \label{fig:data}
\end{figure*}
\else
\fi

\ifalt
    \ifalt
    \subsection{Target Robot Data} 
\else\fi 
All demonstrations are collected via teleoperation on dual Franka Panda robotic arms \anonrep{as described in~\cite{barreiros2025careful}}{as in~\cite{barreiros2025careful}}. 

The observation space includes i) 4 RGB images (missing images are zero-padded), and ii) a natural-language instruction. The action space includes i) end-effector poses w.r.t. the station’s base frame, and ii) gripper widths. Actions are represented as relative trajectories as in~\cite{chi2024universal}. The policy is conditioned only on the current observation and predicts an action chunk with a horizon of 16. 
\else
\fi

\subsection{Dense Language Annotations for Robot Trajectories}
\label{sec:dense_lang_details}
\begin{figure*}[t]
    \centering
    \includegraphics[width=\linewidth]{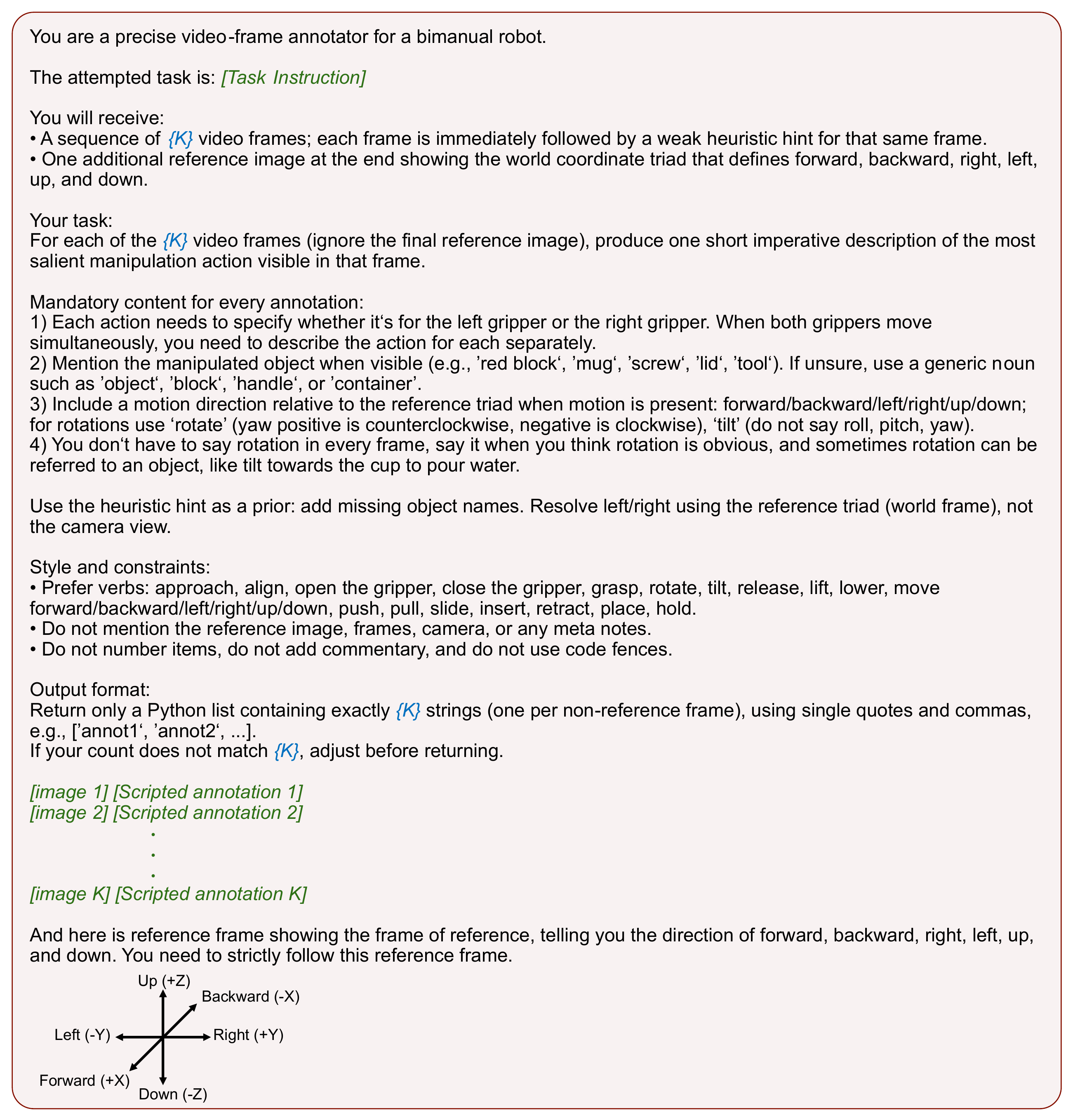}
    \caption{The prompt used to generate VLM-based language annotations for robot data.}
    \label{fig:robot_prompt}
\end{figure*}

\begin{figure*}[t]
    \centering
    \includegraphics[width=0.95\linewidth]{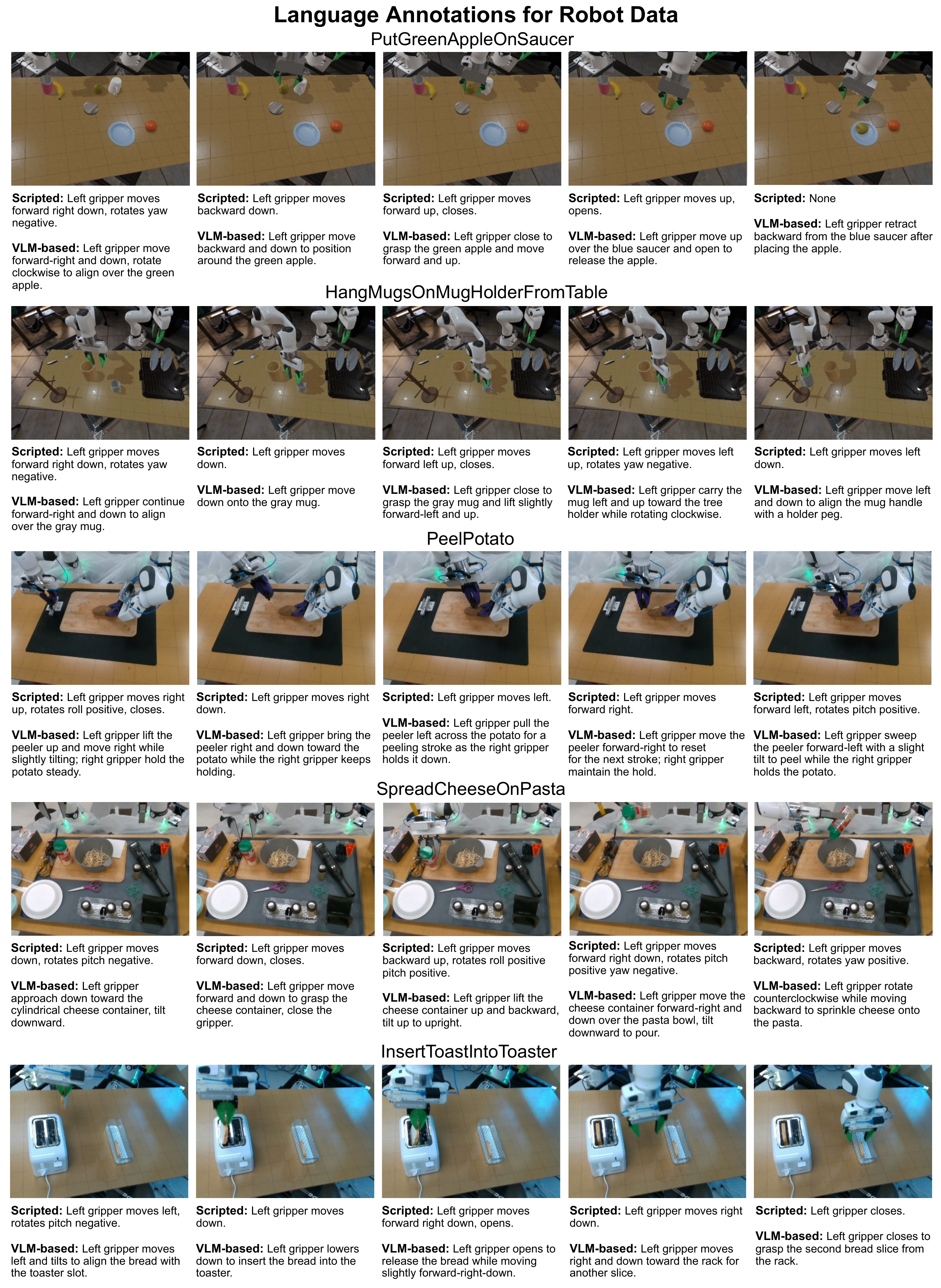}
    \caption{Examples of scripted and VLM-based language annotations for robot data. Each row corresponds to a segment of an episode.}
    \label{fig:examples_scripted_vlm_robot}
\end{figure*}

We derive scripted annotations by comparing the robot’s end effector states at the current step and 16 steps into the future (matching the action chunk horizon). This comparison yields, for each gripper, the movement along the xyz axes, rotational changes in roll, pitch, and yaw, and variations in gripper width. We design the following template for both left and right grippers:
[right/left] gripper moves [forward/backward] [right/left] [up/down], rotates [roll positive/negative] [pitch positive/negative] [yaw positive/negative], [opens/closes]
We set thresholds of 2.5 cm for xyz movement, 20 degrees for rotation, and 1.3 cm for gripper width. When a change does not exceed its corresponding threshold, we omit the associated descriptor block from the annotation.
For VLM-based annotations, we present the complete prompt provided to GPT-5~\cite{gpt5ishere} in Fig.~\ref{fig:robot_prompt}. Furthermore, Fig.~\ref{fig:examples_scripted_vlm_robot} illustrates examples of scripted annotations and VLM-based annotations.

\subsection{Human Videos}
\label{sec:human_videos}
\subsubsection{Latent actions}
\label{sec:latent_action_details}
\textbf{Human Video Data Curation.}
We utilize five publicly available egocentric human video datasets: Ego4D~\cite{grauman2022ego4d}, EgoDex~\cite{hoque2025egodex}, Something-Something V2~\cite{goyal2017something}, Epic Kitchen~\cite{damen2020epic}, and HoloAssist~\cite{wang2023holoassist}. For Ego4D, Epic Kitchen, and HoloAssist, we filter videos using the provided language annotations with heuristic rules (e.g., maximum episode duration and removal of manipulation-irrelevant episodes by action verbs such as \textit{look}, \textit{walk}, \textit{laugh}). For EgoDex and Something-Something V2, which are already structured as high-quality short clips with precise instructions, we use them directly without additional filtering. We also utilize reverse-playback annotations available for a subset of EgoDex videos. The total hours for each dataset are shown in Table~\ref{tab:human_video_annotation_stats}.

\textbf{Latent Action Model (LAM) Implementation.}

\ifalt
    
\else
\fi 

The IDM is a 12-layer spatial-temporal transformer~\cite{xu2020spatial}. A Vector Quantization module (codebook size 32) maps continuous IDM outputs to 8 discrete latent tokens per time step. The visual FDM is a 12-layer spatial transformer~\cite{xu2020spatial}. ActionFDM is a lightweight convolutional decoder from tokens to robot action chunks. We train the LAM with batch size 1024 on 16 H100 GPUs for 300k steps (approximately 68 hours), using learning rate 1e-4 with 1k-step linear warmup and 20k-step cosine decay.

\subsubsection{VLM-generated annotations}
\label{sec:vlm_generated_details}
\begin{figure*}[t]
    \centering
    \includegraphics[width=0.95\linewidth]{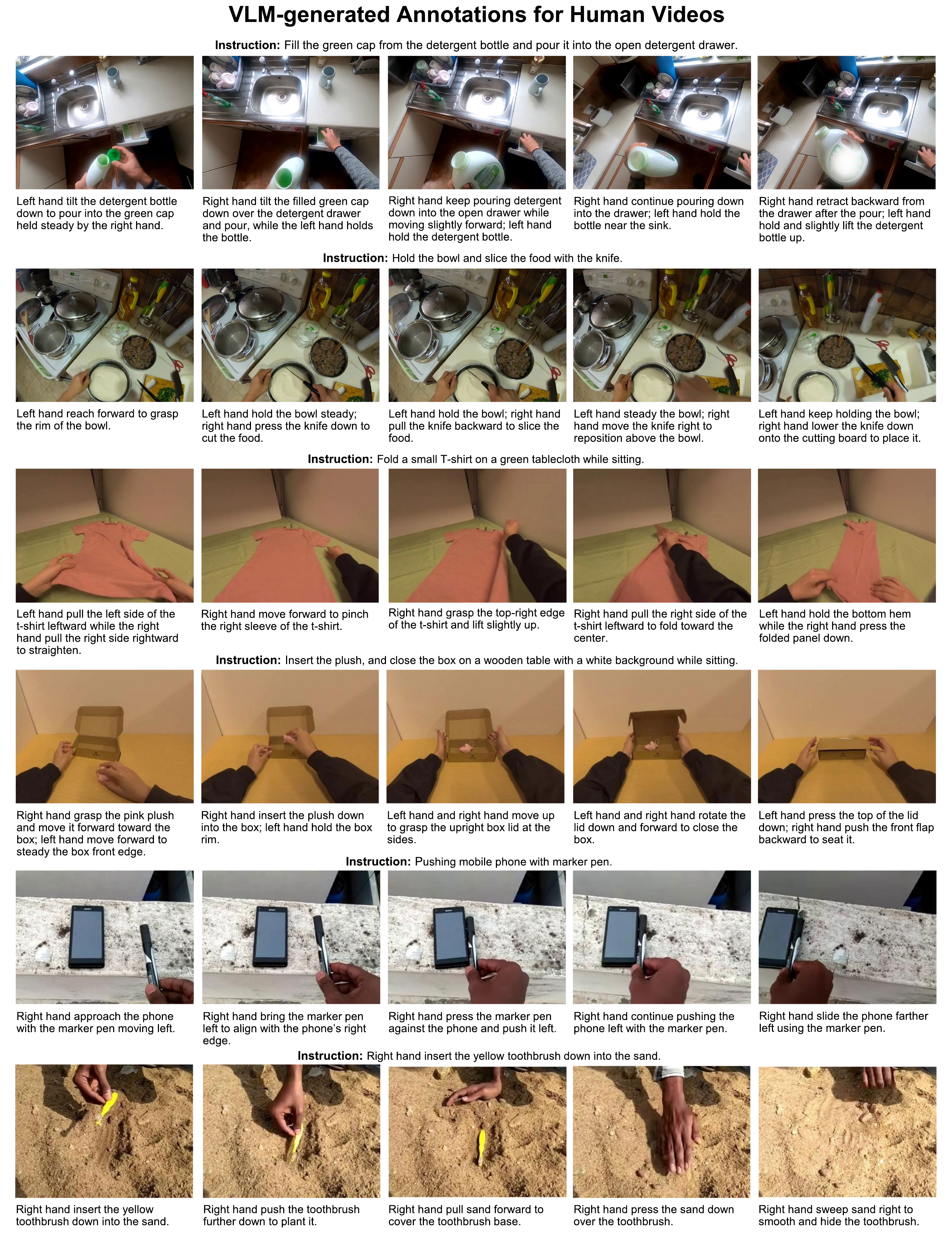}
    \caption{Examples of VLM-generated annotations for human videos. Each row corresponds to a segment of a clip.}
    \label{fig:examples_vlm_egovideo}
\end{figure*}

\begin{figure*}[t]
    \centering
    \includegraphics[width=\linewidth]{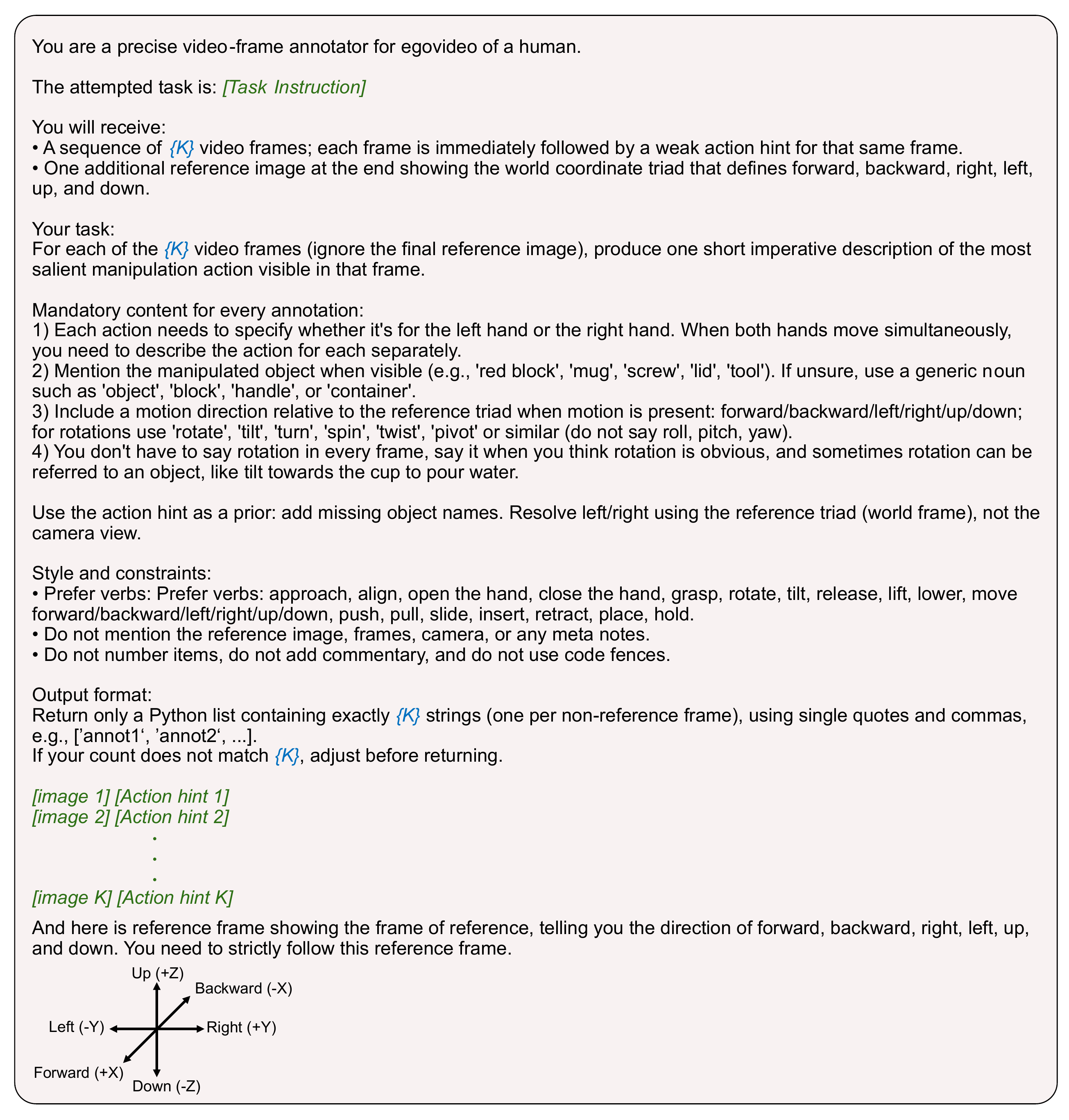}
    \caption{The prompt used for VLM-generated annotations for the Ego4D dataset.}
    \label{fig:ego4d_annotation_prompt}
\end{figure*}

\begin{figure*}[t]
    \centering
    \includegraphics[width=\linewidth]{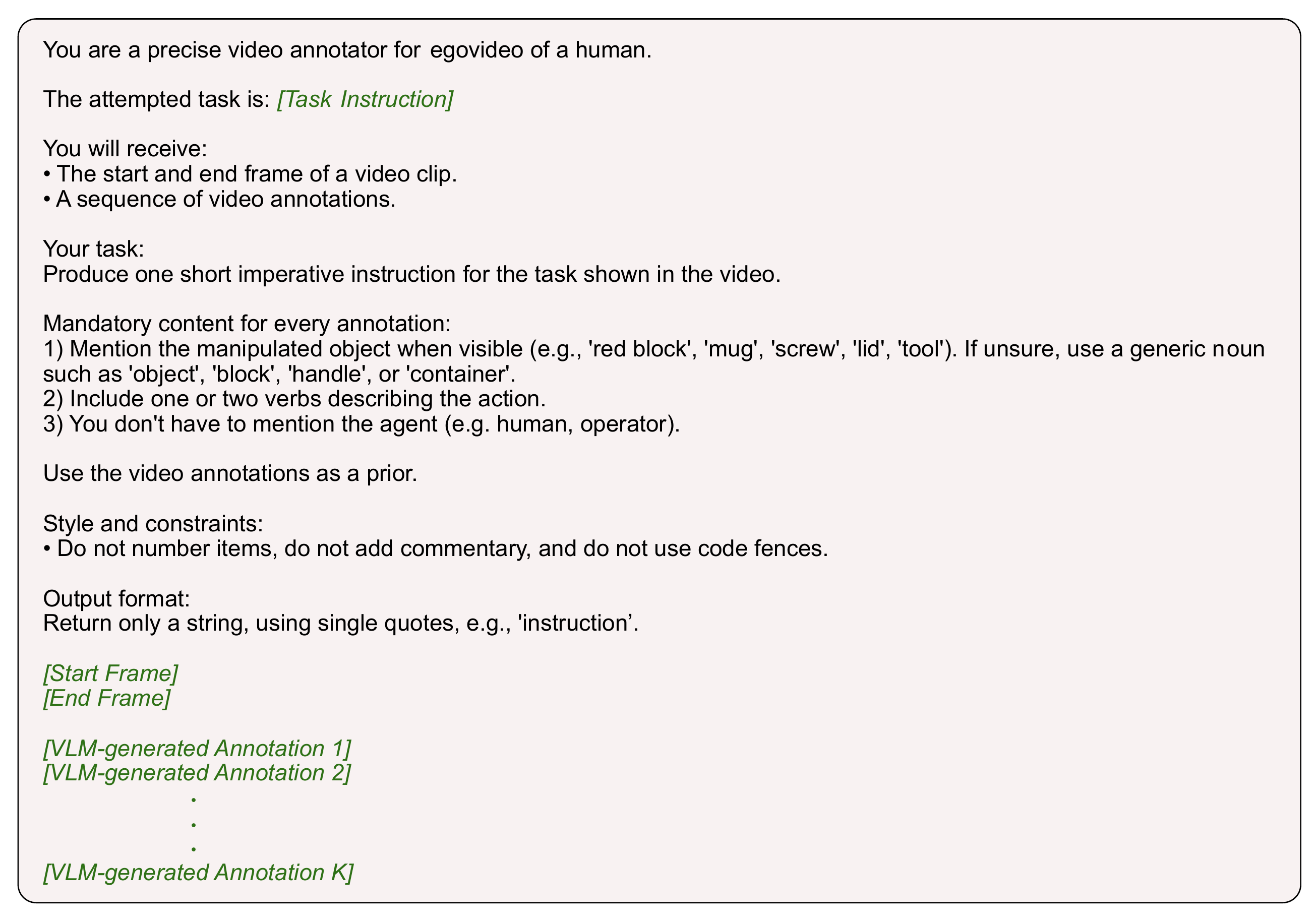}
    \caption{The prompt used to generate an episode instruction for a video clip from the Ego4D dataset.}
    \label{fig:ego4d_instruction_prompt}
\end{figure*}

We utilize three human video datasets for annotation: Ego4D, EgoDex, and Something-Something V2. Representative annotation examples are shown in Fig.~\ref{fig:examples_vlm_egovideo}, with the number of annotated data samples summarized in Table~\ref{tab:human_video_annotation_stats}.
For Ego4D, we first filter the dataset by computing the intersection of video unique identifiers (UIDs) from two filtered subsets: EgoVid~\cite{wang2024egovid} and EgoHOD~\cite{pei2025modeling}, ensuring that only videos retained by both filtering pipelines are included. We then segment the filtered videos into 4-second clips following EgoVid's splits, and downsample each clip at 1-second intervals. Moreover, Ego4D provides fine-grained action descriptions, which we pair with each frame as action priors. Consequently, our prompt to GPT-5 includes, for each clip: (i) the frames downsampled at 1-second intervals, (ii) action priors for each frame, (iii) instructions, and (iv) a reference image depicting the world-frame coordinate triad. The complete prompt is provided in Fig.~\ref{fig:ego4d_annotation_prompt}. This yields a set of frame-level pairs, with one language annotation for each frame. To obtain accurate clip-level instructions, we perform a second prompting step where we prompt GPT-5 with the first and last frames of each clip, along with the complete list of generated annotations, to produce a clear and precise summary of the clip (prompt shown in Fig.~\ref{fig:ego4d_instruction_prompt}).

For EgoDex and Something-Something V2, since neither dataset provides fine-grained action descriptions, we omit the frame-level action priors from the first prompting step. Additionally, both datasets already include high-quality clip-level instructions for each video, so we skip the second prompting step.

\subsection{Discrete Robot Action Tokens}
\label{sec:appendix_discrete_action_tokens}
\textbf{FAST Tokens.} We directly apply the off-the-shelf FAST tokenizer~\cite{pertsch2025fast} to \anonrep{TRI}{Target}-Ramen without fine-tuning, which yields an average token length of 42.1 and a reconstruction error of 2.26e-4. We also experimented with fine-tuning the tokenizer on \anonrep{TRI}{Target}-Ramen, which results in an average token length of 36.8 and a reconstruction error of 2.26e-4. Fine-tuning does not show substantial improvements. Hence, given that the off-the-shelf tokenizer is trained on a larger and more diverse robot dataset, which provides a more generalizable representation, we opt to use it without fine-tuning.

\textbf{VQ-VAE Discrete Action Tokens.} We employ a lightweight VQ-VAE architecture to compress action chunks into discrete tokens. The encoder and decoder are both compact networks: the encoder uses a sequence of 1D convolutional layers with residual blocks to map the action chunk (horizon 16, dimension 20) into a lower-dimensional latent space, while the decoder mirrors the encoder’s architecture to reconstruct the original action chunk from quantized embeddings. The vector quantizer maps each latent vector to its nearest entry in a learned codebook of size 32, discretizing the continuous action space into 8 tokens per action chunk. The model is trained using a combination of reconstruction loss, quantization loss, and commitment loss as described in~\cite{van2017neural}.

\section{Policy Deployment}
\label{sec:policy_deployment}
We employ the same dual-arm Franka robot platform as in~\cite{barreiros2025careful}. For simulation rollouts, the policy predicts a 16-step action chunk at each inference step and executes the first 8 steps in an open-loop manner before recomputing actions. During real-world deployment, we observe that discontinuities between predicted action chunks lead to jerky motions. To address this issue, we adopt a temporal ensemble technique. Specifically, the policy performs continuous inference (with an average latency of 0.146 seconds) to generate temporally overlapping action chunks. At each timestep, the actual action to be executed is computed as a uniform average of the corresponding actions from the four most recently predicted action chunks.

\section{Statistical Analysis}
\label{sec:statistical_analysis_details}

To mitigate the risk of over-interpreting small empirical differences, we adopt the statistical analysis framework of~\cite{barreiros2025careful} to (1) rigorously compare multiple co-training strategies via A/B testing and (2) quantify epistemic uncertainty using Bayesian analysis. For comparisons based on average task completion percentage, we use Welch’s t-test~\cite{welch1947generalization}. For success rate comparisons, we use the STEP test~\cite{snyder2025step}, a state-of-the-art method for A/B testing of binary outcomes.

When comparing $k$ strategies, we perform all $k(k-1)/2$ pairwise tests and control the FWER at 5\% using Bonferroni correction. Final results are summarized using CLD~\cite{piepho2004algorithm}, a label-based representation of A/B tests. Two CLD representations that do not share an English alphabet are statistically separated with at least 95\% confidence. For example, if some strategy A, B, and C respectively have \textit{a}, \textit{ab}, and \textit{b} as the corresponding CLDs, only the pair (A, C) is distinguished with a significant difference in performance. Lexicographical ordering of CLDs often aligns with empirical performance (e.g., A $>$ B $>$ C in this example), but this is not guaranteed when some pairwise differences are not statistically significant. See Fig. \ref{fig:single_modality_ablation}G (``Seen DS" column) for an illustrative case. 

\begin{figure*}[t]
    \centering
    \includegraphics[width=\linewidth]{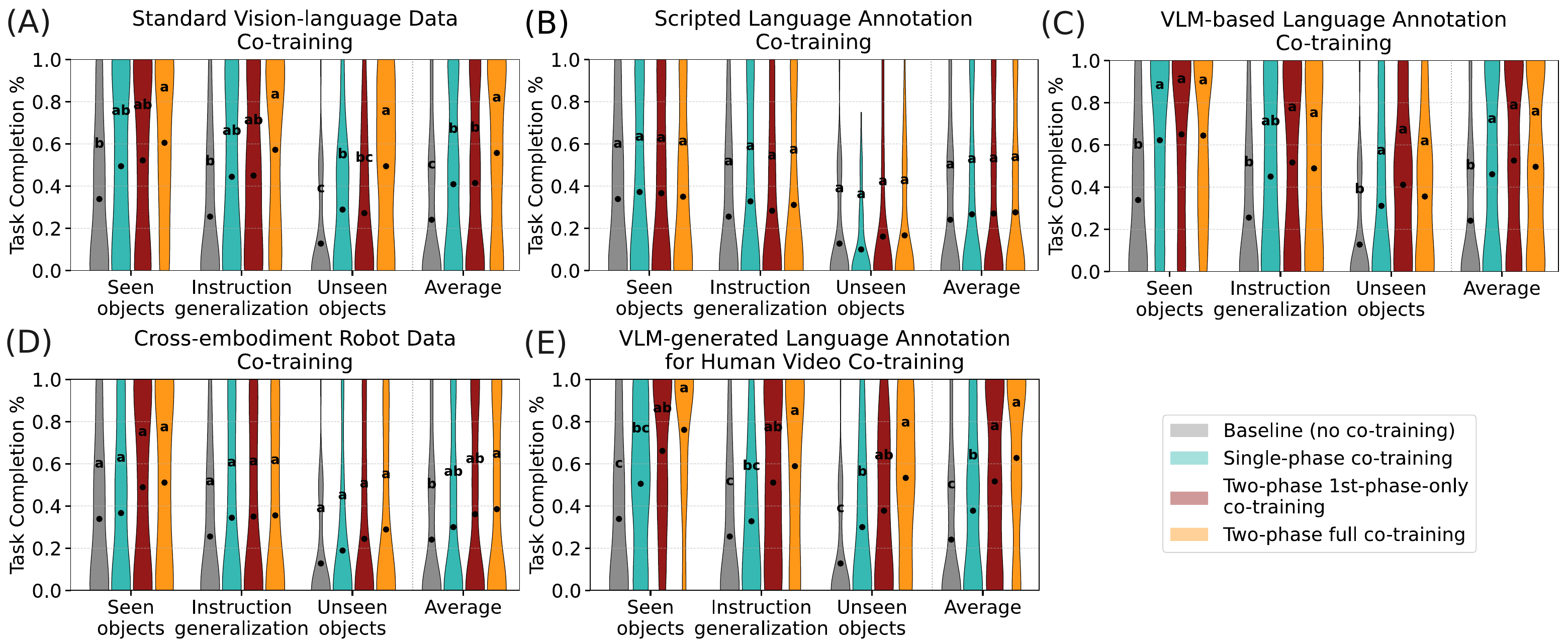}
    \caption{Full distribution figures for the ablation of co-training data sources and strategies in real-world (Fig.~\ref{fig:real_ablation}).}
    \label{fig:real_ablation_full}
\end{figure*}

\begin{figure}[t]
    \centering
    \includegraphics[width=\linewidth]{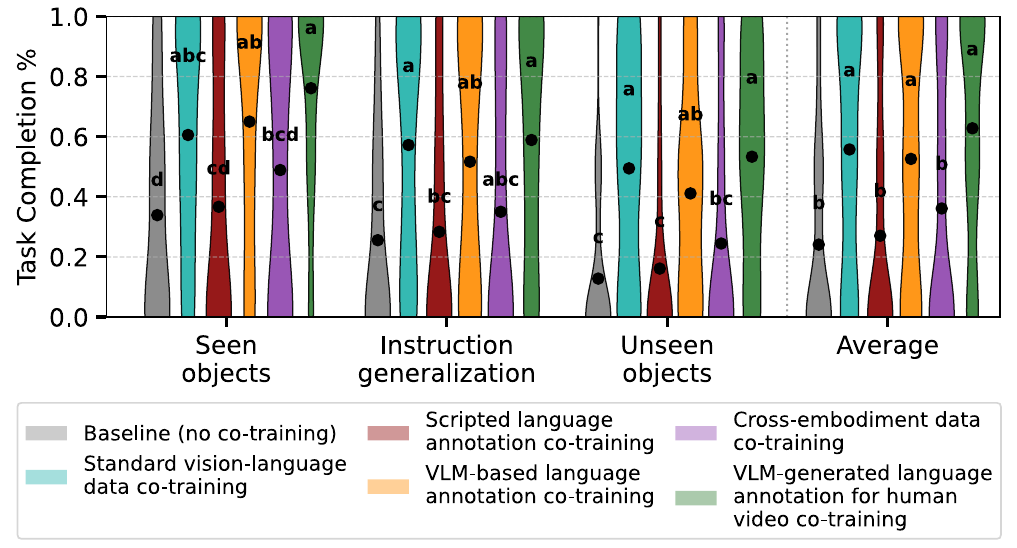}
    \caption{Full distribution figure for the real-world performance of the policies trained with the best co-training strategy for all the effective co-training data modalities (Fig.~\ref{fig:effective_modalities}B).}
    \label{fig:effective_modalities_full}
\end{figure}

\begin{figure}[t]
    \centering
    \includegraphics[width=\linewidth]{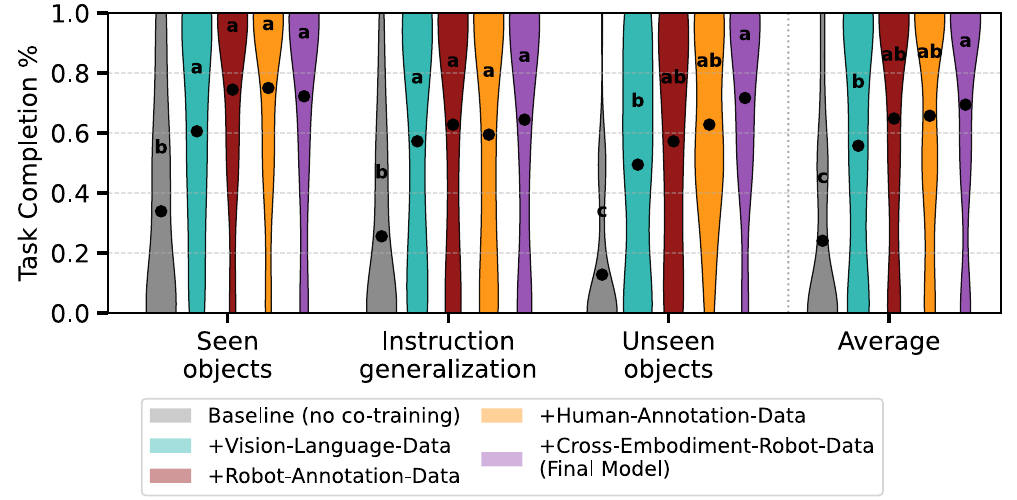}
    \caption{Full distribution figure for the real-world the performance of the policy co-trained with all the effective data modalities combined (Fig.~\ref{fig:combined_modalities}B).}
    \label{fig:combined_modalities_full}
\end{figure}

\begin{figure}[t]
    \centering
    \includegraphics[width=\linewidth]{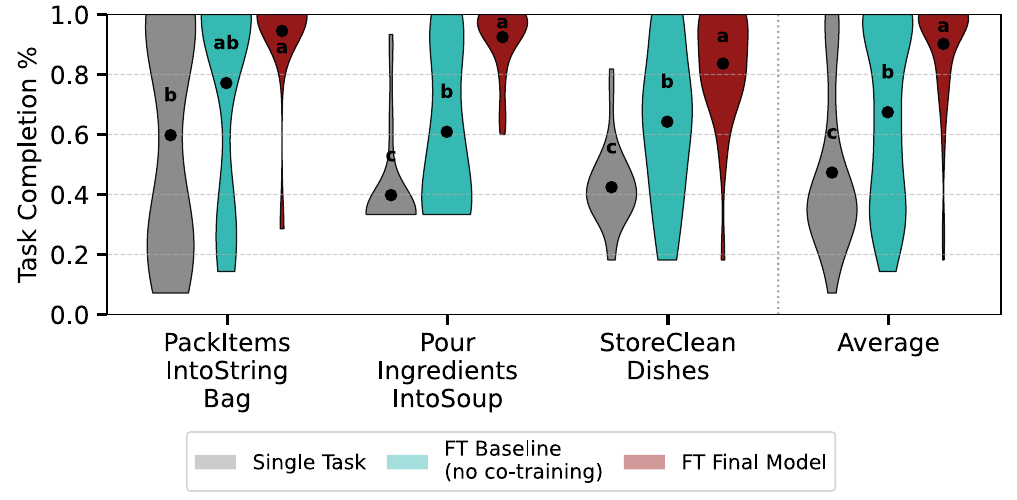}
    \caption{Full distribution figure for the adaptation to unseen long-horizon, dexterous manipulation tasks via fine-tuning (Fig.~\ref{fig:dexterous}).}
    \label{fig:dexterous_full}
\end{figure}

In addition to hypothesis testing, we report Bayesian uncertainty estimates for individual strategies. For success rate, we compute posterior distributions using a uniform Beta prior and observed success counts. For average task completion percentage, we use a uniform Dirichlet prior over task progress values. Posterior uncertainty is visualized as violin plots overlaid on bar charts in the main text (with dots and horizontal lines indicating empirical and posterior means, respectively), while raw empirical distributions are reported in Fig.~\ref{fig:real_ablation_full},~\ref{fig:effective_modalities_full},~\ref{fig:combined_modalities_full},~\ref{fig:dexterous_full} for completeness (with dots only for empirical means).

\section{Simulation Experiment Details}
\label{sec:sim_details}

\begin{figure*}[t]
    \centering
    \includegraphics[width=0.8\linewidth]{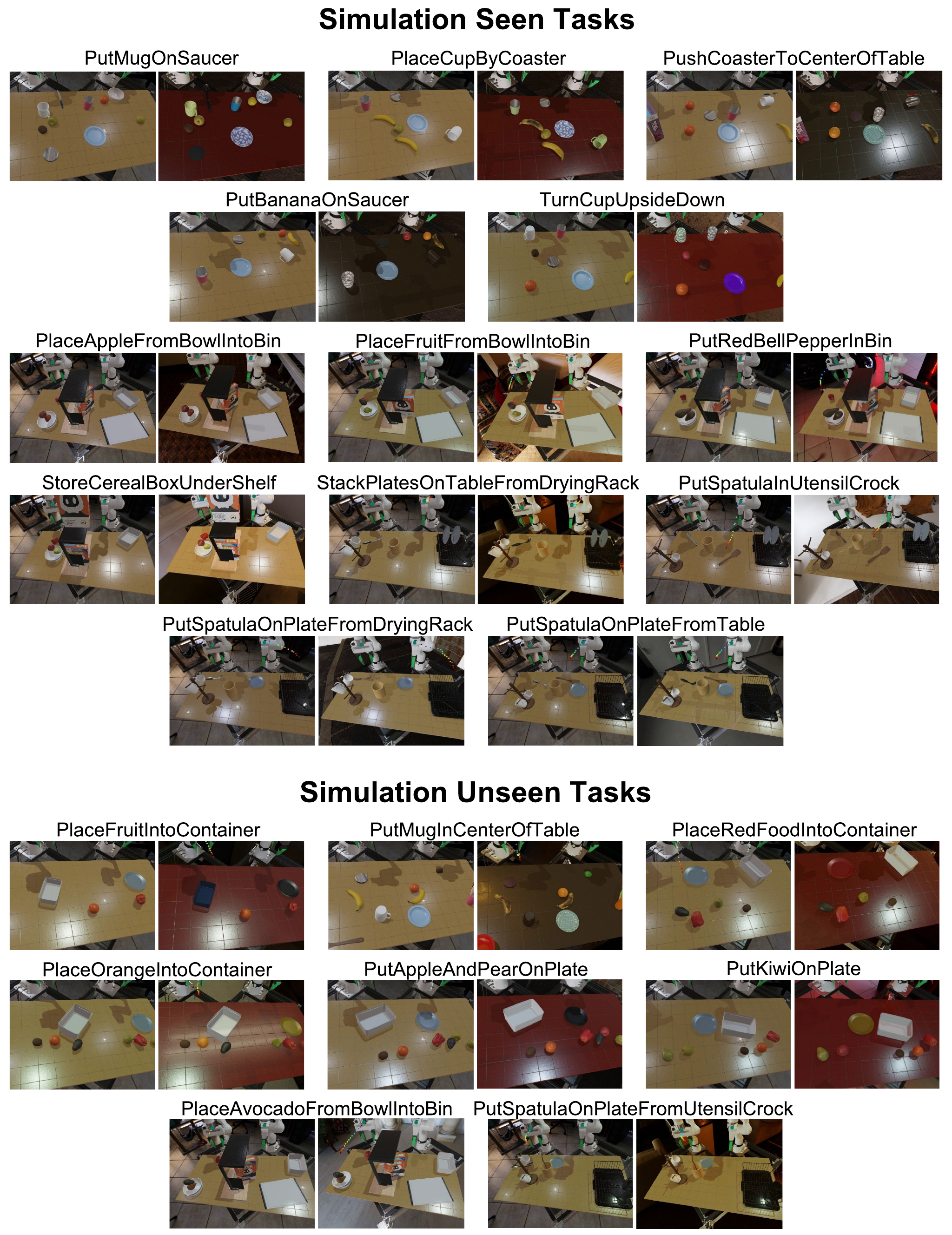}
    \caption{Illustration of the 13 seen tasks and 8 unseen tasks used for evaluation in our simulation benchmark under both nominal and distribution shift conditions.}
    \label{fig:sim_ics}
\end{figure*}

\ifalt
    An overview of the evaluation experiments is shown in Fig.~\ref{fig:sim_real_experiments}. 
    \begin{figure*}[t]
    \centering
    \includegraphics[width=0.8\linewidth]{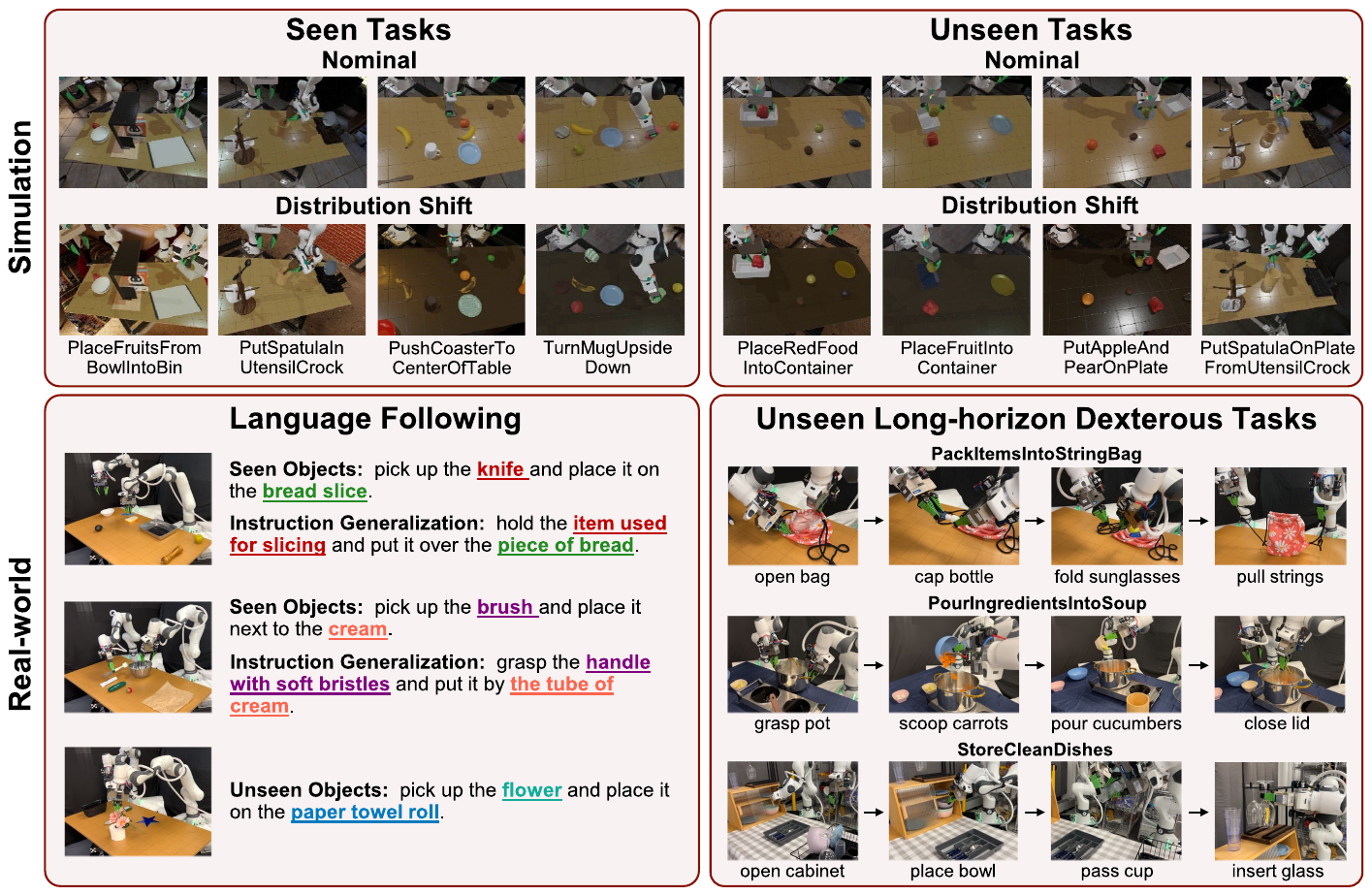}
    \caption{\textbf{Simulation and real-world evaluation.} 
    Policies are evaluated in simulation on 13 seen and 8 unseen tasks under nominal and distribution shift (DS) conditions, where DS introduces appearance changes (e.g., lighting, textures, distractors, camera parameters). Real-world evaluations include language-following experiments with seen objects, instruction generalization through paraphrasing, and unseen objects, as well as adaptation to unseen long-horizon dexterous tasks via fine-tuning.}
    \label{fig:sim_real_experiments}
\end{figure*}

\else
\fi We visualize the 13 seen tasks and 8 unseen tasks in our simulation benchmark under both nominal and distribution shift conditions in Fig.~\ref{fig:sim_ics}, with each task represented by one snapshot showing the initial configuration. For both initial conditions and success criteria, we adopt the same specifications as in~\cite{barreiros2025careful}. Specifically, for initial conditions, each rollout samples from a predefined distribution deterministically based on the simulation seed. We evaluate all the policies using identical seed sets to ensure fair comparison. As evident from the snapshots (Fig.~\ref{fig:sim_ics}), the scenarios are visually ambiguous; hence, policies cannot determine the task based purely on visual appearance and need language to disambiguate. For success criteria, each task is assigned a fixed time budget, and during execution, we monitor a set of task-specific predicates over the simulator state to determine whether the policy has successfully completed the task within the allocated time budget.

\section{Real-world Experiment Details}
\label{sub:real_exp_details}

\subsection{Language Following.} 
\begin{figure*}[t]
    \centering
    \includegraphics[width=\linewidth]{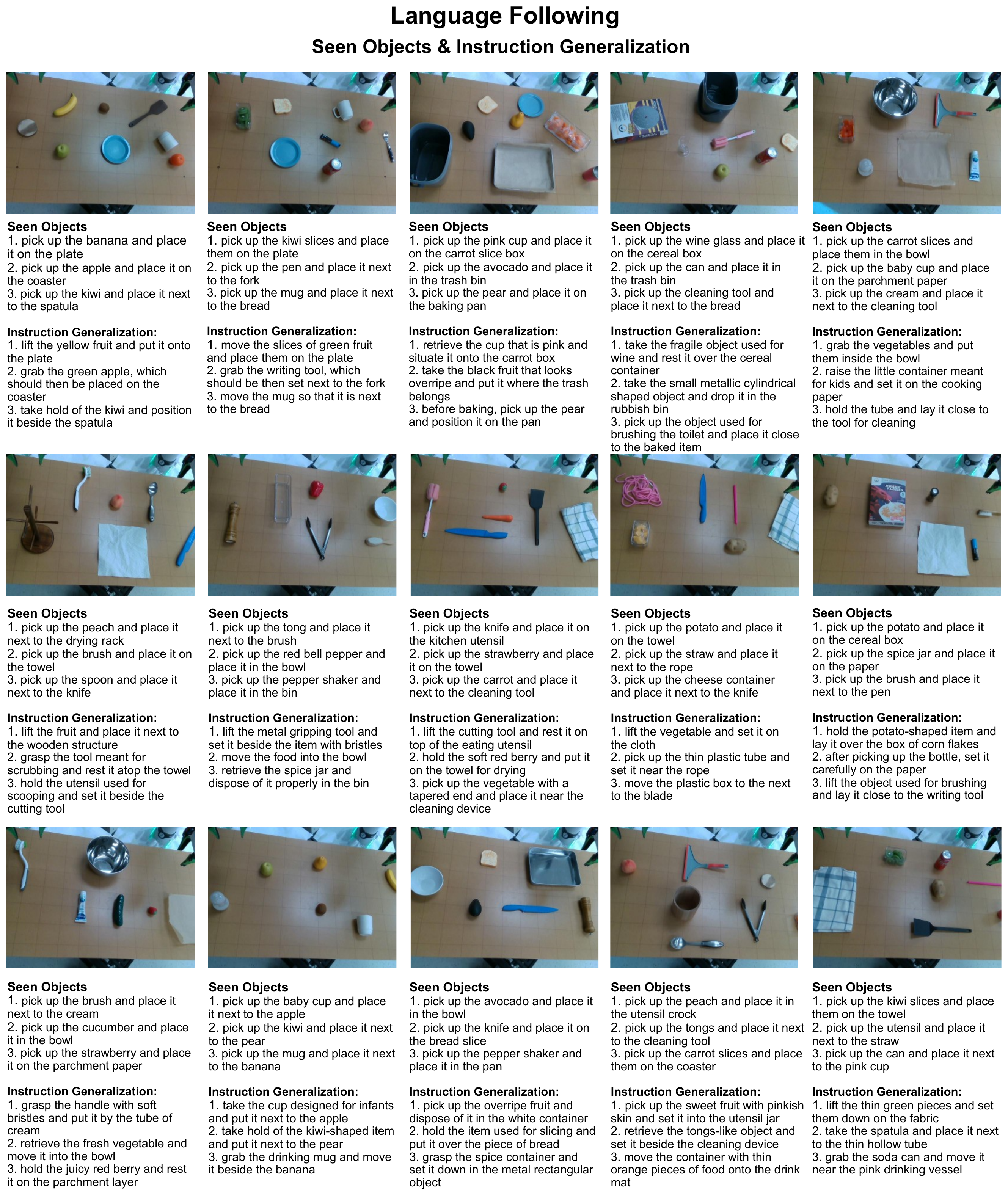}
    \caption{All the initial conditions and instructions used for real-world language following experiments for seen objects and instruction generalization settings. For each initial condition, the numbered labels (e.g., 1.) indicate the corresponding instruction used in the seen objects evaluation and its alternative phrasing used in the instruction generalization evaluation.}
    \label{fig:lang_seen_and_generalization_ics}
\end{figure*}

\begin{figure*}[t]
    \centering
    \includegraphics[width=\linewidth]{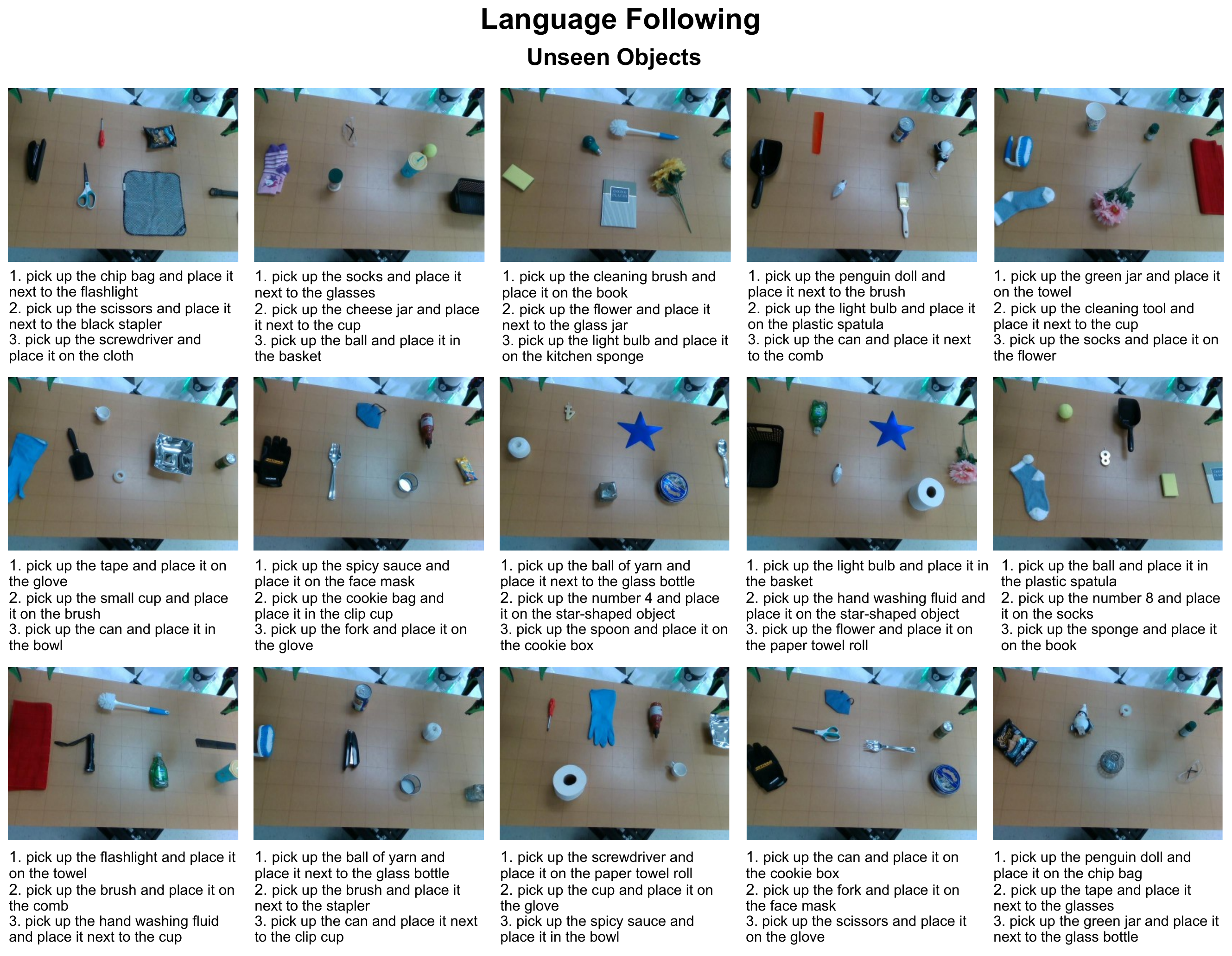}
    \caption{All the initial conditions and instructions used for real-world language following experiments for the unseen objects setting.}
    \label{fig:lang_unseen_ics}
\end{figure*}

We visualize all the initial conditions and corresponding instructions for the Seen Objects, Instruction Generalization, and Unseen Objects settings in Fig.~\ref{fig:lang_seen_and_generalization_ics} and Fig.~\ref{fig:lang_unseen_ics}. Task completion is evaluated based on four milestones:

\begin{enumerate}
    \item The robot moves towards the target object for picking and reaches it within 10 cm.
    \item The robot successfully grasps the target object for picking.
    \item The robot moves the picked object to within 10 cm of the target destination.
    \item The picked object is correctly positioned at the destination according to the instruction:
    \begin{itemize}
        \item \textbf{In/On:} The picked object rests inside or on top of the destination.
        \item \textbf{Next to:} The picked object is positioned within 10 cm of the destination.
    \end{itemize}
\end{enumerate}

\subsection{Long-horizon Dexterous Manipulation.} 
\begin{figure*}[t]
    \centering
    \includegraphics[width=\linewidth]{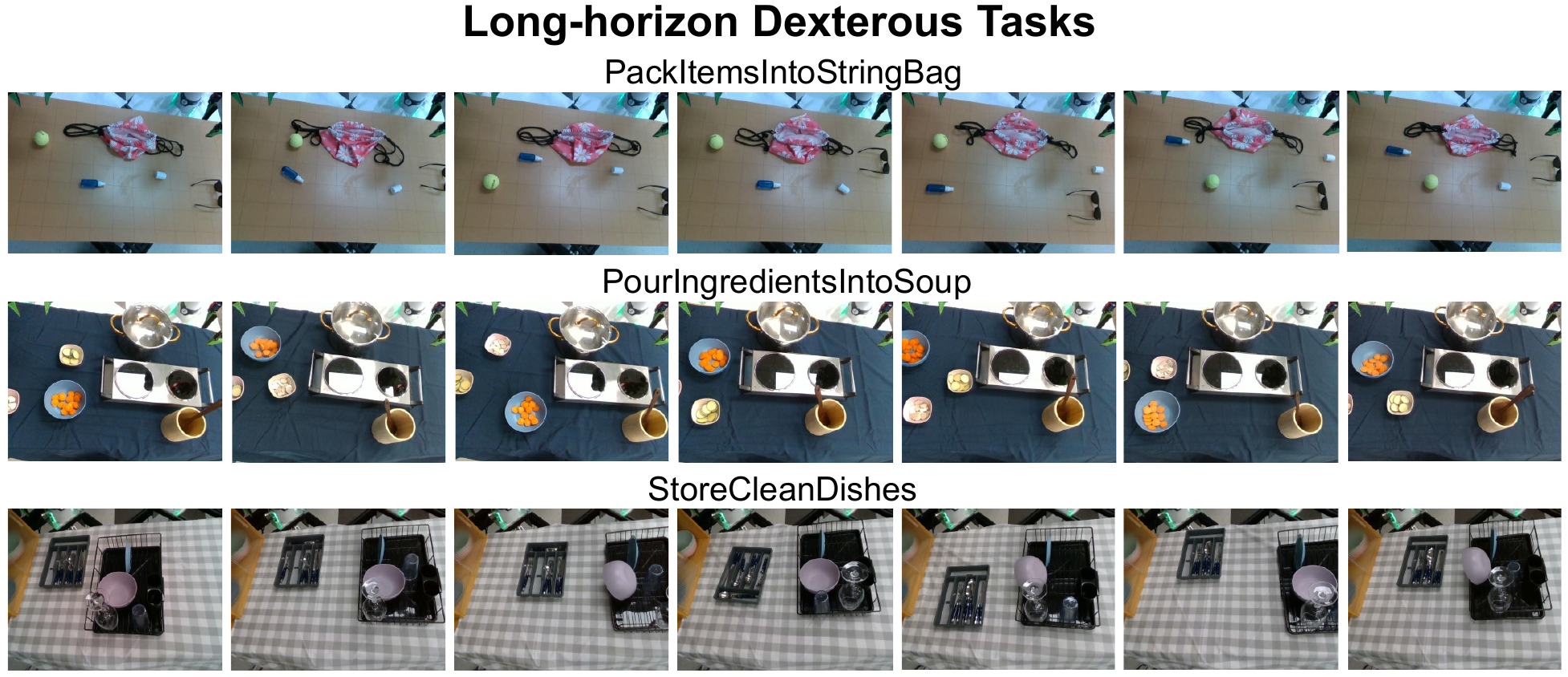}
    \caption{Sample initial conditions for the three real-world unseen, long-horizon, and dexterous tasks.}
    \label{fig:dexterous_ics}
\end{figure*}

\begin{figure*}[t]
    \centering
    \includegraphics[width=\linewidth]{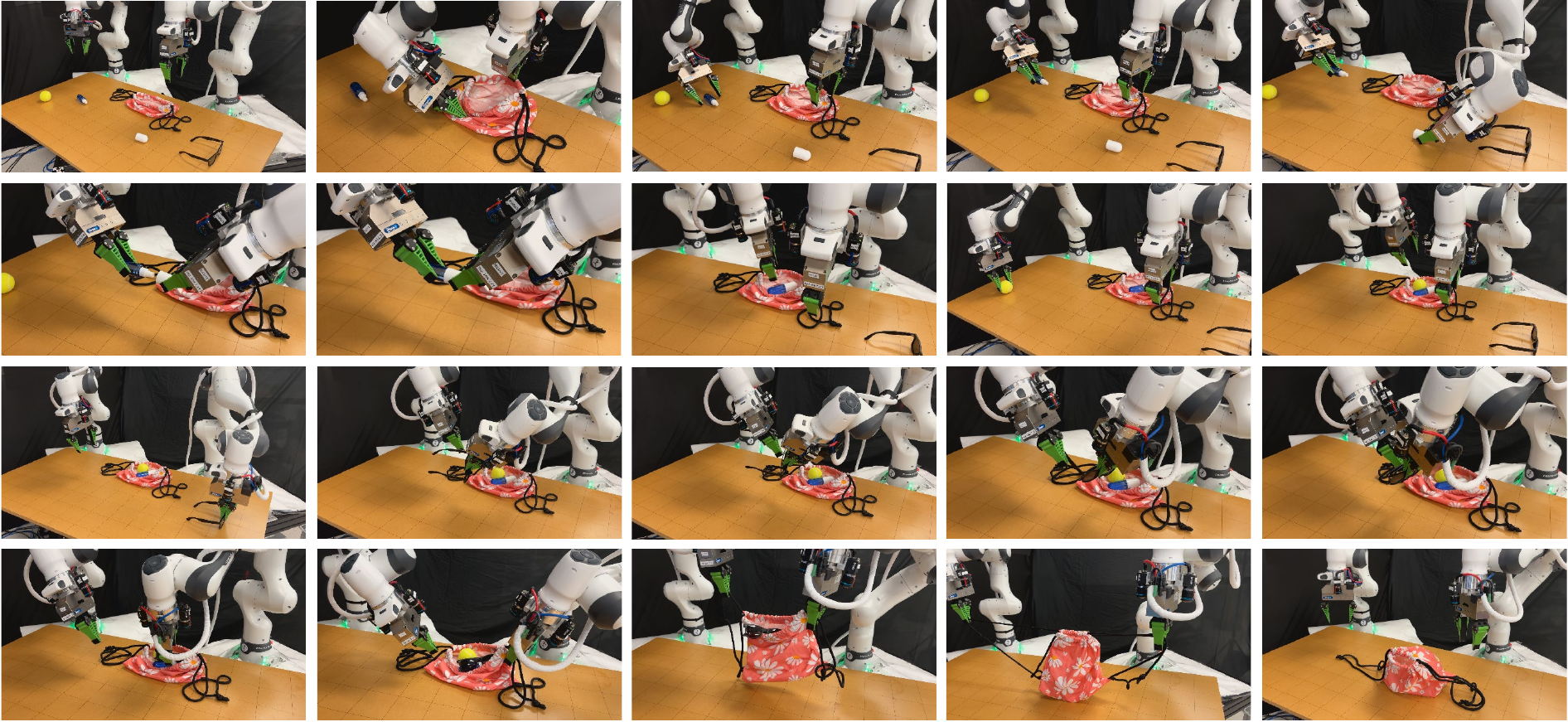}
    \caption{Video snapshots showing the robot executing the \textit{PackItemsIntoStringBag} task. The goal is to open the string bag, pack all the items (bottle, tennis ball, and sunglasses) into it and pull its strings to close it. The robot needs to cap the bottle and fold the sunglasses before packing them.}
    \label{fig:video_pack_items_bag}
\end{figure*}

\begin{figure*}[t]
    \centering
    \anonrep{\includegraphics[width=\linewidth]{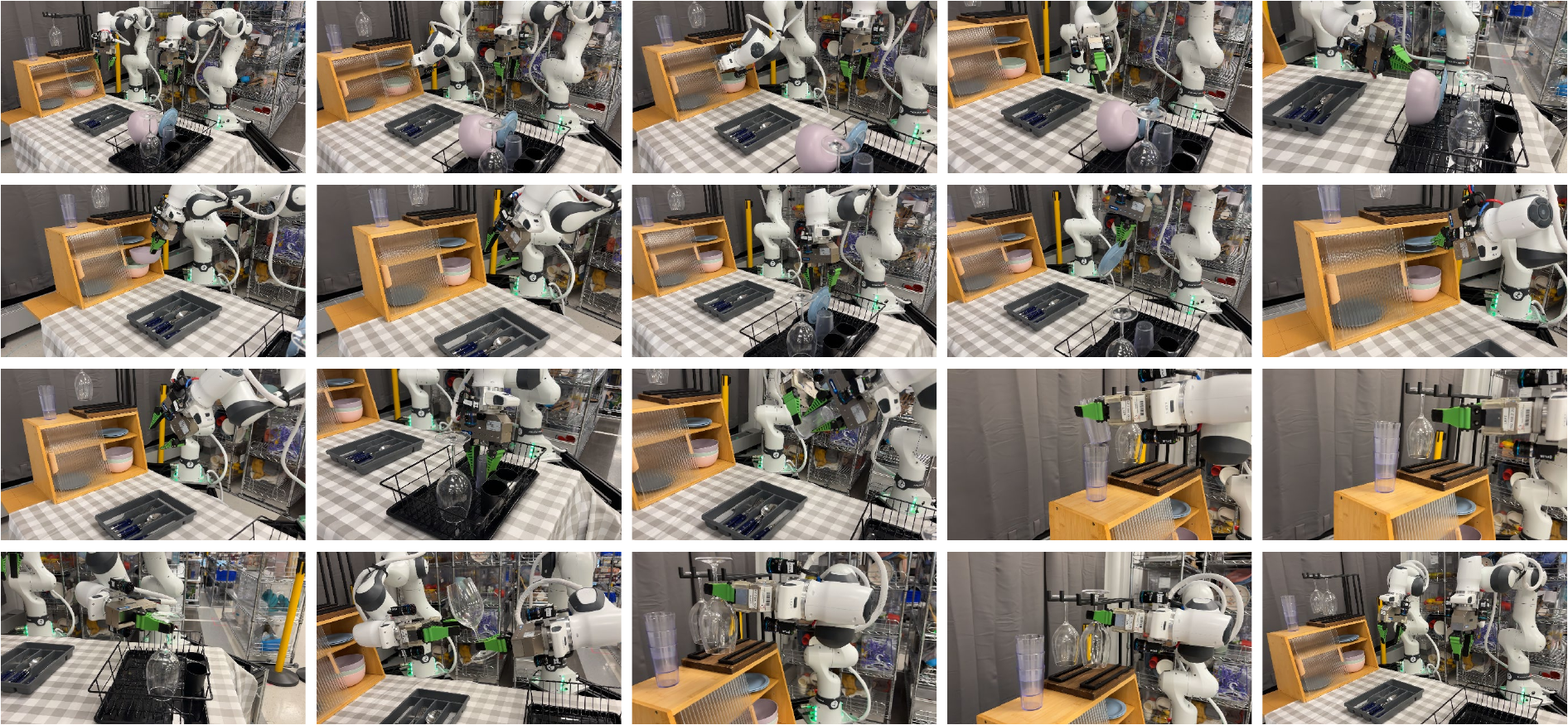}}{\includegraphics[width=\linewidth]{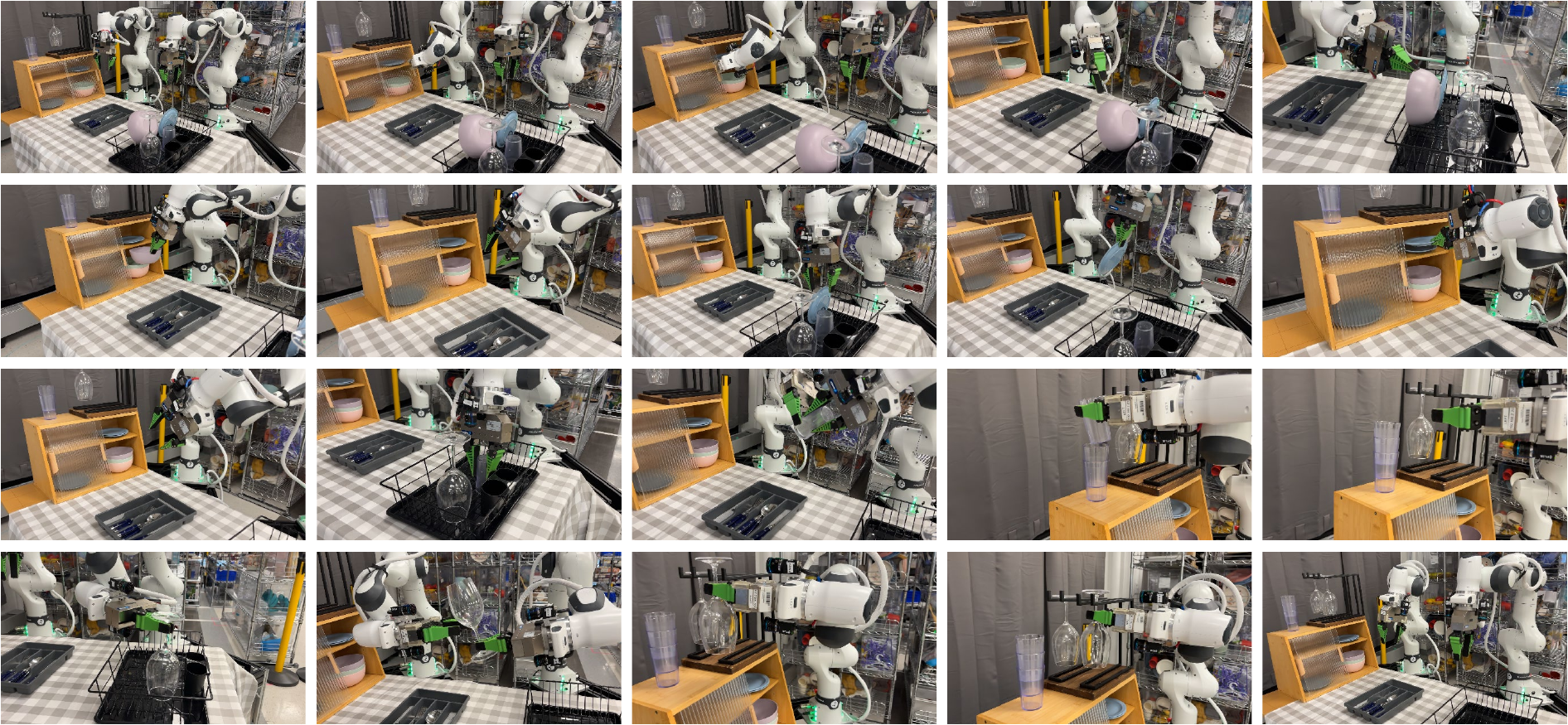}}
    \caption{Video snapshots showing the robot executing the \textit{StoreCleanDishes} task. The goal is to store all the items (bowl, plate, cup, and wine glass) from the drying rack into the dish cabinet. The robot must place the bowl onto the bowl stack, the plate onto the plate stack, stack the cup on top of the other cups, and insert the wine glass into the glass rack.}
    \label{fig:video_store_clean_dishes}
\end{figure*}

\begin{figure*}[t]
    \centering
    \includegraphics[width=\linewidth]{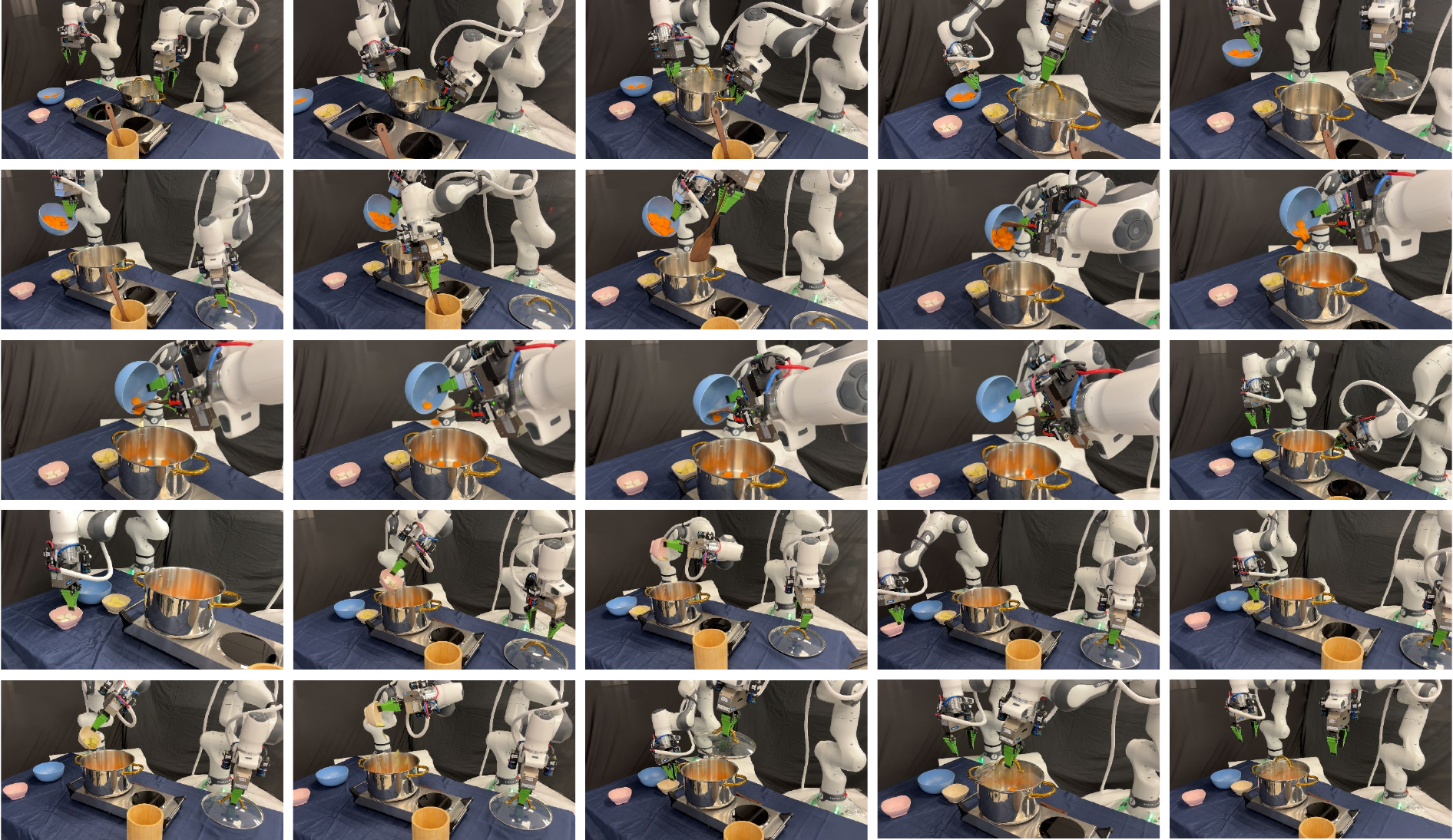}
    \caption{Video snapshots showing the robot executing the \textit{PourIngredientsIntoSoup} task. The goal is to pour all the ingredients (chopped carrots, mushrooms, and cucumbers) into the soup pot. Prior to pouring, the robot must put the pot onto the stove and remove its lid. The robot must scoop out all of the carrot slices using a spatula.}
    \label{fig:video_pour_ingredients_soup}
\end{figure*}

We visualize the sample initial conditions for the three tasks in Fig.~\ref{fig:dexterous_ics}. A collage of video frames showing the task execution is displayed in Fig.~\ref{fig:video_pack_items_bag},~\ref{fig:video_store_clean_dishes},~\ref{fig:video_pour_ingredients_soup}. The task completion milestones for each task are defined as follows:

\textbf{PackItemsIntoStringBag}
\begin{itemize}
    \item The left arm grasps the rim of the drawstring bag.
    \item The right arm grasps the opposite rim and opens the bag.
    \item The robot picks up the bottle.
    \item The robot picks up the bottle cap.
    \item The robot securely caps the bottle.
    \item The robot places the bottle inside the bag.
    \item The robot picks up the ball.
    \item The robot places the ball inside the bag.
    \item The robot picks up the sunglasses.
    \item The robot folds the first temple of the sunglasses.
    \item The robot folds the second temple of the sunglasses.
    \item The robot places the folded sunglasses inside the bag.
    \item The robot grasps the two drawstrings.
    \item The robot closes the bag by pulling the drawstrings.
\end{itemize}

\textbf{PourIngredientsIntoSoup}
\begin{itemize}
    \item The robot grasps both handles of the soup pot.
    \item The robot places the pot onto the left burner of the stove.
    \item The right arm picks up the lid.
    \item The right arm places the lid on the table.
    \item The left arm picks up the carrot bowl and moves it on top of the pot.
    \item The right arm grasps the spatula.
    \item The robot pours the carrots into the pot using the spatula.
    \item The right arm places the spatula on the table.
    \item The left arm places the empty bowl on the table.
    \item The left arm picks up the mushroom bowl and moves it on top of the pot.
    \item The left arm pours the mushrooms into the pot and places the empty bowl on the table.
    \item The left arm picks up the cucumber bowl and moves it on top of the pot.
    \item The left arm pours the cucumbers into the pot and places the empty bowl on the table.
    \item The right arm picks up the lid.
    \item The right arm places the lid onto the pot.
\end{itemize}

\textbf{StoreCleanDishes}
\begin{itemize}
    \item The robot opens the cabinet door.
    \item The robot picks up the bowl.
    \item The robot stacks the bowl onto the bowl inside the cabinet.
    \item The robot picks up the plate.
    \item The robot stacks the plate onto the plate inside the cabinet.
    \item The robot picks up the cup.
    \item The right arm passes the cup to the left arm.
    \item The robot stacks the cup into the cup on the cabinet.
    \item The robot picks up the glass.
    \item The right arm passes the glass to the left arm.
    \item The robot inserts the glass upside-down into the rack.
\end{itemize}

\subsection{Experimental Procedure}
\label{sub:real_exp_procedure}

All real-world evaluations follow a standardized and controlled experimental procedure to ensure fair comparison across policies. For each evaluation setting, we first generate a list of policy checkpoints to be evaluated. Checkpoints are selected sequentially from this list. For a given evaluation session, an initial condition (IC), including the set of objects and their poses, is sampled according to the task specification. The table and surrounding workspace are then manually configured to match this IC as closely as possible.

The selected policy checkpoint is executed once on the physical robot, after which the experimenter records task outcomes by following the rubrics. The evaluator then proceeds to the next checkpoint in the list, using the same initial condition. This process is repeated until all checkpoints have been evaluated on that IC, after which a new IC is sampled and the process restarts.

By evaluating all checkpoints under the same initial condition within a short time window, this protocol ensures that differences in performance are attributable to policy behavior rather than variations in hardware, lighting, or environmental setup. In effect, all checkpoints are evaluated under as nearly identical physical conditions as possible. Moreover, any slow and imperceptible changes in the conditions would equally affect all the checkpoints.

\subsection{Rubric QA}
Similar to~\cite{barreiros2025careful}, we conduct a quality assurance (QA) round to estimate the frequency of potential discrepancies due to human error or bias in the rubric evaluation of real-world rollouts. We sample 895 evaluation rollouts (31.6\% of the total) and ask reviewers drawn from a separate pool than the original evaluators to review the rollout videos and their rubrics. To mitigate bias, the reviewers do not know which checkpoint is evaluated in each video. The QA overall task completion discrepancy, calculated as the average over all episodes and milestones, is 2.01\%.

\begin{table*}[t]
\centering
\caption{VLM backbone evaluation results across vision-language benchmarks.}
\label{tab:vlm_benchmark}
\scriptsize
\setlength{\tabcolsep}{4pt}
\begin{tabular}{p{8cm}cccccccc}
\toprule
\textbf{Model} & \textbf{RealWorldQA} & \textbf{GQA} & \textbf{SpatialEval} & \textbf{MMBench} & \textbf{SEED} & \textbf{MME-P} & \textbf{MME-R} & \textbf{LEGO} \\
\midrule
PaliGemma2-PT (VLA backbone) & 0.24 & 31.68 & 0.26 & 0.01 & 0.24 & 701.56 & 242.50 & 0.03 \\
PaliGemma2-Mix & 0.55 & 59.81 & 0.34 & 0.68 & 0.70 & 1449.43 & 299.29 & 0.27 \\
Baseline (no co-training) & 0.02 & 0.06 & 0.02 & 0.00 & 0.03 & 193.75 & 44.29 & 0.01 \\
Standard Vision Language Data Co-training (+Vision-Language-Data) & 0.39 & 57.69 & 0.28 & 0.26 & 0.49 & 88.24 & 54.64 & 0.27 \\
Scripted Language Annotation Co-training & 0.05 & 0.00 & 0.07 & 0.00 & 0.06 & 80.06 & 32.14 & 0.02 \\
VLM-based Language Annotation Co-training & 0.02 & 0.06 & 0.03 & 0.00 & 0.04 & 92.00 & 31.07 & 0.01 \\
Cross-embodiment Robot Data Co-training & 0.01 & 0.01 & 0.01 & 0.00 & 0.01 & 114.02 & 33.21 & 0.00 \\
VLM-generated Language Annotation for Human Video Co-training & 0.03 & 0.02 & 0.01 & 0.07 & 0.09 & 105.43 & 65.36 & 0.03 \\
+Robot-Annotation-Data & 0.49 & 57.58 & 0.32 & 0.60 & 0.65 & 977.79 & 177.14 & 0.27 \\
+Human-Annotation-Data & 0.48 & 55.78 & 0.39 & 0.33 & 0.51 & 1397.43 & 253.57 & 0.27 \\
+Cross-Embodiment-Robot-Data (Final Model) & 0.47 & 58.02 & 0.33 & 0.56 & 0.63 & 1319.75 & 267.86 & 0.27 \\
\bottomrule
\end{tabular}
\end{table*}

\end{document}